\documentclass[]{damo}

\usepackage{wrapfig}
\usepackage{tabularx}
\usepackage{textcomp}
\usepackage{stfloats}
\usepackage{url}
\usepackage{verbatim}
\usepackage{graphicx}
\usepackage{titlesec}
\usepackage{tocloft}
\usepackage{adjustbox}
\usepackage{multirow}
\usepackage{pifont}
\usepackage{tikz}
\usepackage{comment}
\usepackage{amsmath,amssymb} 
\usepackage{colortbl}  
\usepackage{color}
\usepackage{booktabs} 

\usepackage{hyperref}
\usepackage{graphicx}    
\usepackage{subcaption} 
\usepackage{booktabs}
\usepackage{amsmath} 
\usepackage{multirow} 
\usepackage{booktabs} 
\usepackage{subcaption} 
\usepackage{longtable}
\RequirePackage{xspace}
\makeatletter
\DeclareRobustCommand\onedot{\futurelet\@let@token\@onedot}
\def\@onedot{\ifx\@let@token.\else.\null\fi\xspace}

\def\eg{\emph{e.g}\onedot}

\makeatother

\definecolor{adptorange}{RGB}{248, 205, 172}
\definecolor{cmpblue}{RGB}{189, 215, 238}
\definecolor{cmpblue}{RGB}{189, 215, 238}

\definecolor{our_red}{RGB}{232,157,160}
\definecolor{our_blue}{RGB}{136,206,230}
\definecolor{our_orange}{RGB}{246,200,168}
\definecolor{our_green}{RGB}{178,211,164}

\definecolor{attn_code0}{RGB}{247,215,200}
\definecolor{attn_code1}{RGB}{238,169,139}
\definecolor{mlp_code0}{RGB}{204,201,221}
\definecolor{mlp_code1}{RGB}{102,95,153}

\definecolor{token_blue}{RGB}{84, 120, 140}

\usepackage{pifont}       
\usepackage{bbding}       
\usepackage{fontawesome}
\usepackage{xspace}

\usepackage{float}
\usepackage{array} 
\usepackage{ragged2e} 
\usepackage{longtable} 

\newlength\savewidth
\newcommand{\tablestyle}[2]{\setlength{\tabcolsep}{#1}\renewcommand{\arraystretch}{#2}\centering\footnotesize}

\newcolumntype{x}[1]{>{\centering\arraybackslash}p{#1pt}}
\newcolumntype{y}[1]{>{\raggedright\arraybackslash}p{#1pt}}
\newcolumntype{z}[1]{>{\raggedleft\arraybackslash}p{#1pt}}

\renewcommand{\paragraph}[1]{\vspace{1mm}\noindent\textbf{#1}}
\usepackage{colortbl}
\usepackage{xcolor}
\usepackage{wrapfig}

\newcommand\hshline{\noalign{\global\savewidth\arrayrulewidth
  \global\arrayrulewidth 0.5pt}\hline\noalign{\global\arrayrulewidth\savewidth}}

\renewcommand{\paragraph}[1]{\vspace{1.25mm}\noindent\textbf{#1}}

\usepackage{algorithm}
\usepackage{listings}

\definecolor{codeblue}{rgb}{0.25, 0.5, 0.5}
\definecolor{codekw}{rgb}{0.35, 0.35, 0.75}
\lstdefinestyle{Pytorch}{
    language = Python,
    backgroundcolor = \color{white},
    basicstyle = \fontsize{9pt}{8pt}\selectfont\ttfamily\bfseries,
    columns = fullflexible,
    aboveskip=1pt,
    belowskip=1pt,
    breaklines = true,
    captionpos = b,
    commentstyle = \color{codeblue},
    keywordstyle = \color{codekw},
}

\definecolor{green}{HTML}{009000}
\definecolor{red}{HTML}{ea4335}

\title{DyDiT++: Diffusion Transformers with Timestep and Spatial Dynamics for Efficient Visual Generation}

\author[* 1, 2]{Wangbo Zhao}
\author[* 2]{Yizeng Han}
\author[\dagger 2, 3]{Jiasheng Tang}
\author[1]{Kai Wang}
\author[2, 3]{Hao Luo}
\author[2, 3]{Yibing Song}
\author[4]{Gao Huang}
\author[2]{Fan Wang}
\author[\dagger 1]{Yang You}

\affiliation[1]{National University of Singapore}
\affiliation[2]{DAMO Academy, Alibaba Group \\}
\affiliation[3]{Hupan Lab}
\affiliation[4]{Tsinghua University}
\contribution[*]{Equal contribution}
\contribution[\dagger]{Corresponding author}

\abstract{
Diffusion Transformer (DiT), an emerging diffusion model for visual generation, has demonstrated superior performance but suffers from substantial computational costs. Our investigations reveal that these costs primarily stem from the \emph{static} inference paradigm, which inevitably introduces redundant computation in certain \emph{diffusion timesteps} and \emph{spatial regions}. To overcome this inefficiency, we propose \textbf{Dy}namic \textbf{Di}ffusion \textbf{T}ransformer (DyDiT), an architecture that \emph{dynamically} adjusts its computation along both \emph{timestep} and \emph{spatial} dimensions. Specifically, we introduce a \emph{Timestep-wise Dynamic Width} (TDW) approach that adapts model width conditioned on the generation timesteps. In addition, we design a \emph{Spatial-wise Dynamic Token} (SDT) strategy to avoid redundant computation at unnecessary spatial locations. TDW and SDT can be seamlessly integrated into DiT and significantly accelerate the generation process. Building on these designs, we present an extended version, \textbf{DyDiT++}, with improvements in three key aspects. 
First, it extends the generation mechanism of DyDiT beyond diffusion to flow matching, demonstrating that our method can also accelerate flow-matching-based generation, enhancing its versatility. Furthermore, we enhance DyDiT to tackle more complex visual generation tasks, including video generation and text-to-image generation, thereby broadening its real-world applications.  Finally, to address the high cost of full fine-tuning and  democratize technology access, we investigate the feasibility of training DyDiT in a parameter-efficient manner and introduce timestep-based dynamic LoRA (TD-LoRA).
Extensive experiments on diverse visual generation models, including DiT, SiT, Latte, and FLUX, demonstrate the effectiveness of DyDiT++. Remarkably, with $<$3\% additional fine-tuning iterations, our approach reduces the FLOPs of DiT-XL by 51\%, yielding 1.73$\times$ realistic speedup on hardware, and achieves a competitive FID score of 2.07 on ImageNet.  The code is available at \textbf{\url{https://github.com/alibaba-damo-academy/DyDiT}}. \\

}

\date{\today}

\begin{document}
\thispagestyle{firstheader}
\maketitle
\pagestyle{empty}

\renewcommand{\footnoterule}{%
    \vspace{1em} 
    \hrule width 0.9\textwidth 
    \vspace{0.5em} 
}
\renewcommand{\thefootnote}{$\bullet$}
\footnotetext[1]{\textbf{This paper was accepted to the IEEE Transactions on Pattern Analysis and Machine Intelligence (TPAMI) on January 9, 2026.}}

\section{Introduction} \label{sec:introduction}
Diffusion models \citep{ho2020denoising, dhariwal2021diffusion, rombach2022high, blattmann2023stable} have demonstrated significant superiority in visual generation tasks.  Recently, the remarkable scalability of Transformers \citep{vaswani2017attention, dosovitskiy2020image} has led to the growing prominence of Diffusion Transformer (DiT) \citep{peebles2023scalable}. DiT and its variants have shown strong potential in a wide range of applications, including image generation~\citep{chen2023pixart, chen2024pixart, esser2024scaling, flux2024} and video generation~\citep{videoworldsimulators2024, ma2024latte, polyak2024movie, wan2.1}. Like Transformers in other vision and language domains~\citep{dosovitskiy2020image, brown2020language}, DiT experiences notable generation inefficiency.
However, unlike ViT~\citep{dosovitskiy2020image} or LLMs~\citep{brown2020language}, the multi-timestep paradigm in DiT inherently introduces additional computational complexity. Moreover, generative tasks often exhibit unbalanced difficulty across spatial regions, further amplifying the inefficiency issue.

To this end, we propose to perform \textbf{dynamic computation} for efficient inference of DiT. In orthogonal to other lines of work such as efficient samplers \citep{song2020denoising, song2023consistency, salimans2022progressive, meng2023distillation, luo2023latent} and global acceleration techniques \citep{ma2023deepcache, pan2024t}, this work focuses on reducing computational redundancy within DiT, from the model perspective. A representative solution in this line is network compression, \eg structural pruning \citep{fang2024structural, molchanov2016pruning, he2017channel}. However, pruning methods typically retain a \emph{static} architecture across both \emph{timestep} and \emph{spatial} dimensions throughout the diffusion process. As shown in Figure~\ref{fig:figure1(2)}(a), both the original and the pruned DiT employ a fixed model width across all diffusion timesteps and allocate the same computational cost to every image patch. This \emph{static} inference paradigm overlooks the varying complexities associated with different timesteps and spatial regions, leading to significant inefficiency. To explore this redundancy in more detail, we analyze the training process of DiT, during which it is optimized for a noise prediction task. Our analysis yields two key insights:

\begin{figure}[t]
    \centering
    \includegraphics[width=0.9\textwidth]{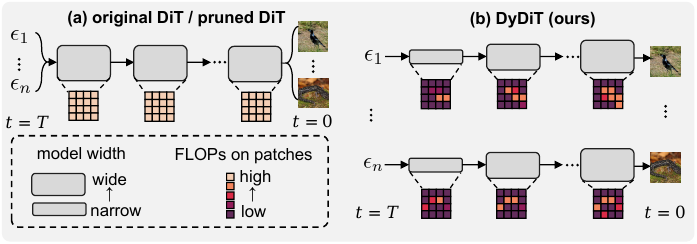}
\caption{The core idea of DyDiT.
 } 
\label{fig:figure1(2)}
\end{figure}

\begin{figure*}[t]
    \centering
    \includegraphics[width=\textwidth]{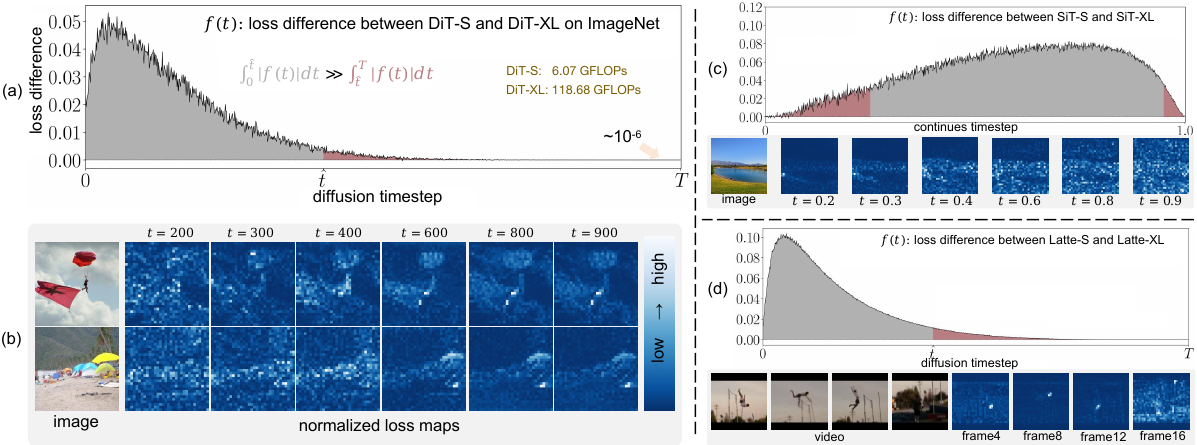}
\caption{
(a) The loss difference between DiT-S and DiT-XL is slight at most timesteps. (b) The Loss maps (normalized within [0, 1]) show that the noise in different spatial locations has varying difficulty levels to predict. 
(c) The loss paradigm of the flow matching-based method, SiT~\citep{ma2024sit}. (d) The loss paradigm of Latte~\citep{ma2024latte} with 16 frames sampled from $t=600$. }
\label{fig:figure1(1)}
\end{figure*}

\textbf{a) \emph{Timestep perspective}}: We plot the loss value differences between a pre-trained small model (DiT-S) and a larger model (DiT-XL) in Figure~\ref{fig:figure1(1)}(a). The results show that the loss differences diminish substantially for $t > \hat{t}$, and even approach negligible levels as $t$ nears the noise distribution ($t \to T$). This indicates that the prediction task becomes \emph{progressively easier at later timesteps} and could be managed effectively even by \emph{a smaller model}. However, DiT applies the same architecture across all timesteps, leading to excessive computational costs at \textit{timesteps where the task complexity is low}.

\textbf{b) \emph{Spatial perspective}}: We visualize the loss maps in Figure~\ref{fig:figure1(1)}(b) and observe a noticeable imbalance in loss values in different spatial regions of the image. Losses are higher in patches corresponding to the main object, while the background regions exhibit relatively lower loss. This suggests that the difficulty of noise prediction varies across spatial regions. Consequently, \textit{uniform computational treatment of all patches introduces redundancy and is likely suboptimal.}

Based on the above insights, 
we propose \textbf{Dy}namic \textbf{Di}ffusion \textbf{T}ransformer (\textbf{DyDiT}), which adaptively allocates computational resources during the generation process, as illustrated in Figure~\ref{fig:figure1(2)}(b). Specifically, from the timestep perspective, we introduce a \emph{Timestep-wise Dynamic Width} (\textbf{TDW}) mechanism, where the model learns to adjust the width of the attention and MLP blocks based on the current \emph{timestep}. From a spatial perspective, we develop a \emph{Spatial-wise Dynamic Token} (\textbf{SDT}) strategy, which identifies image patches where noise prediction is relatively ``easy'', allowing them to bypass computationally intensive blocks, thus reducing unnecessary computation.

Building on the aforementioned explorations, we propose \textbf{DyDiT++}, improving DyDiT in three  key aspects:

\textbf{a) \emph{Flow-Matching Generation Acceleration}}: Flow matching~\citep{ma2024sit, esser2024scaling, lipman2022flow, liu2022flow, flux2024} has spearheaded progress in visual synthesis. 
However, computational redundancy in the flow matching's iterative process has been less explored. Our analysis identifies distinct redundancy dynamics unique to flow matching and demonstrates that our method seamlessly accelerates flow-matching generation.

\textbf{b) \emph{Cross-Task Generalization}}: 
While DiT variants extend generative scope to video and multi-modal tasks~\citep{ma2024latte, esser2024scaling, flux2024}, their architecture-specific complexities persist.
We demonstrate DyDiT\xspace's adaptability through both \emph{architecture} refinements and \emph{training} scheme optimizations.

\textbf{c) \emph{Training Cost Reduction}}: Beyond inference efficiency, training efficiency remains critical for large-scale models under computational constraints (\eg limited GPU memory). We observed that the standard LoRA~\citep{hu2021lora} would degrade DyDiT's generative performance. To address this, we propose \emph{timestep-dependent LoRA (TD-LoRA)}, enabling an improved parameter-efficient finetuning scheme for DyDiT.

One of the most appealing advantages of our method
may be its \textbf{generalizability}: Both TDW and SDT are plug-and-play modules that can be seamlessly implemented on DiT-based architectures~\citep{peebles2023scalable,ma2024sit,ma2024latte,flux2024}. Moreover, DyDiT contributes to significant speedup due to its \textbf{hardware-friendly} design: 1) TDW allows the model architecture at each timestep to be pre-determined offline, eliminating overhead for width adjustments at runtime; 2) token skipping in SDT is straightforward to implement, incurring minimal overhead. Such hardware efficiency distinguishes DyDiT from traditional dynamic networks~\citep{herrmann2020channel,meng2022adavit,han2023latency}, which adjust their inference graphs for each sample and struggle to improve practical efficiency in batched inference.

Extensive experiments are conducted across multiple visual generation models to validate the effectiveness of the proposed method. Notably, compared to the static counterpart DiT-XL, DyDiT-XL reduces FLOPs by \textbf{51\%} (\textbf{1.73$\times$} realistic speedup on hardware), with \emph{$<$3\%} fine-tuning iterations, while maintaining a competitive FID score of 2.07 on ImageNet (256$\!\times\!$256)~\citep{deng2009imagenet}. Building upon this success, DyDiT++ further expands its advantages to encompass a wider range of visual generation tasks and scenarios. Specifically, it generalizes to reduce the computational cost of the flow-matching-based model SiT~\citep{ma2024sit} by more than \textbf{50\%}. Furthermore, by integrating our dynamic architecture into the video generation model Latte~\citep{ma2024latte} and the text-to-image generation model FLUX~\citep{flux2024}, we achieve a speedup of \textbf{1.62}$\times$ and \textbf{1.59}$\times$, respectively, while maintaining competitive or superior performance. Finally, in resource-constrained scenarios, our TD-LoRA approach introduces only \textbf{1.4\%}  trainable parameters, reduces GPU memory usage by \textbf{26\%}, and achieves an impressive FID score of 2.23, highlighting its training efficiency.

\begin{figure}[t]
    \centering
    \includegraphics[width=0.9\textwidth]{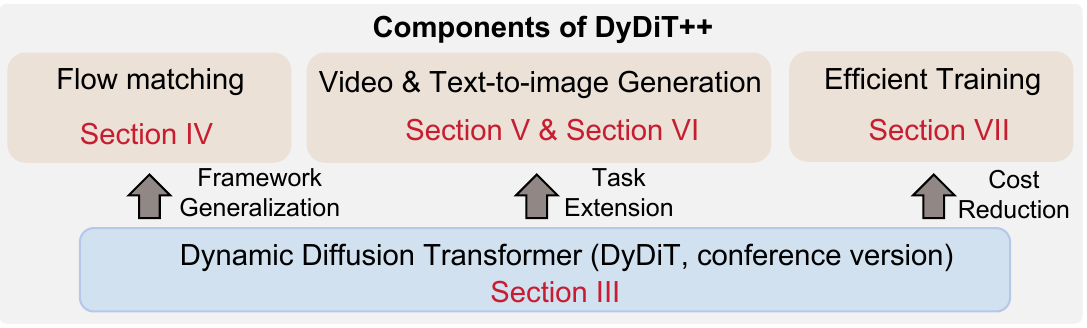}
\caption{The components of DyDiT++, which is a significantly enhanced version of its conference predecessor, DyDiT~\citep{zhao2024dydit}}
\label{fig:compare_dydit}
\end{figure}

This study, DyDiT++, extends its conference version, DyDiT~\citep{zhao2024dydit} with key improvements shown in Figure~\ref{fig:compare_dydit}:

\begin{itemize}

    \item We investigate the computational redundancy problem in \textbf{flow matching}~\citep{ma2024sit, esser2024scaling, lipman2022flow, liu2022flow, flux2024} (Figure~\ref{fig:figure1(1)}(c) and Section~\ref{sec:flow}) and demonstrate that our method can be seamlessly extended to SiT~\citep{ma2024sit}, which replaces the diffusion process in DiT with flow matching, further validating the generalizability of our approach (Table~\ref{tab:sit}).

    \item We identify the computational redundancy across both timestep spatial-temporal dimensions during video generation (Figure~\ref{fig:figure1(1)}(d)). To address this, we propose \textbf{DyLatte} by adapting DyDiT to a representative architecture, Latte~\citep{ma2024latte} (Section~\ref{sec:videodiffusion}). Experiments on diverse datasets validate the effectiveness (Figures~\ref{fig:flops_fvd_latte}\ref{fig:video_visuzalization}, Table~\ref{tab:video_speed}).

    \item We propose \textbf{DyFLUX} to perform dynamic text-to-image generation (Section~\ref{sec:mmdit}). It adapts DyDiT to FLUX~\citep{flux2024}, a representative multi-modal structure. \emph{(i)} From the \emph{architecture} perspective, we modify the design of DyDiT to reduce the redundant computation in both types of blocks in FLUX (Figure~\ref{fig:t2i_flops}). \emph{(ii)} For training, we develop a distillation technique to align both output and intermediate representations between DyFLUX and the static FLUX teacher (Equation~\ref{eq:distill}). This extension significantly reduces the cost of generating high-resolution photorealistic images (1024$\times$1024) while maintaining quality (see Table~\ref{tab:flux_geneval} and Figure~\ref{fig:userstudy}), thereby enhancing the practicality of DyDiT for real-world applications.

    \item We investigate the feasibility of training DyDiT in a \textbf{parameter-efficient} manner, \emph{i.e.} LoRA~\citep{hu2021lora} (Section~\ref{sec:dtlora}). Our findings demonstrate that a static diffusion transformer can be transformed into a dynamic architecture by fine-tuning only a minimal number of parameters. We further propose an \textbf{improved PEFT approach tailored for DyDiT}: Timestep-based Dynamic LoRA (\textbf{TD-LoRA}). By modifying the B-matrix in LoRA as an MoE~\citep{cai2024survey} structure and dynamically mixing the weights based on the diffusion timestep, TD-LoRA uses trainable parameters more effectively than LoRA, improving the generation quality (Tables~\ref{tab:compare_LoRA}, \ref{tab:design_TD-LORA}, \ref{tab:lora_rank}).

\end{itemize}

\vspace{-2mm}
\section{Related Works}
\vspace{-2mm}
\paragraph{Efficient Diffusion Models.}
The generation speed of diffusion models \citep{ho2020denoising, rombach2022high} has always hindered their further applications primarily due to long sampling steps and high computational costs. Existing attempts to make diffusion models efficient can be roughly categorized into sampler-based, model-based, and global acceleration methods. The sampler-based methods \citep{song2020denoising, song2023consistency, salimans2022progressive, meng2023distillation, luo2023latent} aim to reduce the sampling steps. Model-based approaches \citep{fang2024structural, so2024temporal, shang2023post, yang2023diffusion} attempt to compress the size of diffusion models via pruning \citep{fang2024structural, shang2023post} or quantization \citep{li2023q, shang2023post}. Global acceleration methods like Deepcache \citep{ma2023deepcache} tend to reuse or share some features across different timesteps.

DyDiT and DyDiT++ primarily relate to model-based approaches, orthogonal to other works. Unlike pruning methods yielding  \textit{static} architectures, our methods enable \textit{dynamic} computation across diffusion timesteps and input tokens.

\paragraph{Transformer-based Diffusion Models.}
Diffusion Transformer (DiT)~\citep{peebles2023scalable} is an early attempt to extend the scalability of transformers~\citep{vaswani2017attention} to diffusion models. U-ViT~\citep{bao2023all}, developed concurrently, combines the strengths of both U-Net~\citep{ronneberger2015u} and transformers. 
To address the absence of text input support in DiT, PixArt~\citep{chen2023pixart, chen2024pixart} integrates multi-head cross-attention for text input. Building on this, SD3~\citep{esser2024scaling} and FLUX~\citep{flux2024} introduce MM-DiT, enhancing text-image interactions via joint processing of text and image tokens with self-attention. Recognizing DiT's potential, Latte~\citep{ma2024latte} extends it to video generation with temporal attention. Similarly, the video foundation model Sora~\citep{videoworldsimulators2024} is also built upon the DiT architecture.

This proposed DyDiT is primarily designed to accelerate DiT, while the enhanced version, DyDiT++, further demonstrates its generalizability across various architectures.

\paragraph{Dynamic Neural Networks.}
Compared to static models, dynamic neural networks \citep{han2021dynamic, wang2025emulating} adapt their computational graph based on inputs, enabling a superior trade-off between performance and efficiency.  They typically adjust network depth~\citep{teerapittayanon2016branchynet, bolukbasi2017adaptive, yang2020resolution, han2022learning, han2023dynamic} or width~\citep{herrmann2020channel, li2021dynamic, han2023latency} during inference. Some works also explore the spatial redundancy in visual \emph{perception} \citep{wang2021not, song2021dynamic, rao2021dynamicvit, liang2022not, meng2022adavit, zhao2025stitch}. Despite their theoretical efficiency, existing methods usually struggle in achieving \emph{practical efficiency} during batched inference \citep{han2023latency, zhao2025rapid} due to the per-sample inference graph. Moreover, the potential of dynamic architectures in diffusion models, where a \emph{timestep} dimension is introduced, remains unexplored.

This work extends the research of dynamic networks to the image \emph{generation} field. More importantly, our TDW adjusts network structure only conditioned on \emph{timesteps}, avoiding the sample-conditioned weight shapes in batched inference. Together with the efficient token selection in SDT, DyDiT and DyDiT++ show preferable realistic efficiency.

\paragraph{Parameter Efficient Fine-tuning} (PEFT) aims to fine-tune pre-trained models by updating few parameters. Representatively, LoRA~\citep{hu2021lora} reduces the number of tunable parameters through two low-rank matrices, and has been widely adopted due to its simplicity and efficiency. To handle knowledge from different domains, subsequent works \citep{tian2025hydralora,dou2023loramoe,liu2023moelora} modify the LoRA parameters based on input features.

In DyDiT++, we first allow training DyDiT with the standard LoRA. Furthermore, we propose Timestep-based Dynamic LoRA (TD-LoRA), a method tailored to adapt parameters across different timesteps, enhancing the parameter efficiency of DyDiT while maintaining competitive performance.

\section{Dynamic Diffusion Transformer (DyDiT)}
\noindent
We first provide an overview of diffusion models and DiT \citep{peebles2023scalable} in Section~\ref{sec:predliminary}. DyDiT's timestep-wise dynamic width (TDW) and spatial-wise dynamic token (SDT) approaches are then introduced in Sections~\ref{sec:dynamic_channel} and \ref{sec:dynamic_token}. Finally, Section~\ref{sec:training} details the training process of DyDiT.

\vspace{-5mm}
\subsection{Preliminary} \label{sec:predliminary}

\vspace{-1mm}
\paragraph{Diffusion Models} \citep{ho2020denoising,  nichol2021improved, rombach2022high} generate images from random noise through a series of diffusion steps. These models typically consist of a forward diffusion process and a reverse denoising process. In the forward process, given an image $\mathbf{x}_0 \!\sim\!q(\mathbf{x})$ sampled from the data distribution, Gaussian noise $\epsilon\!\sim\!\mathcal{N}(0, I)$ is progressively added over $T$ steps. This process is defined as $q\left(\mathbf{x}_t\!\mid\!\mathbf{x}_{t-1}\right)\!=\!\mathcal{N}\left(\mathbf{x}_t ; \sqrt{1-\beta_t} \mathbf{x}_{t-1}, \beta_t \mathbf{I}\right)$, where $t$ and $\beta_t$ denote the timestep and noise schedule, respectively. In the reverse process, the model removes the noise and reconstructs $\mathbf{x}_0$ from $\mathbf{x}_T\!\sim\!\mathcal{N}(0, I)$ using $p_\theta\left(\mathbf{x}_{t-1}\!\mid\!\mathbf{x}_t\right)\!=\!\mathcal{N}\left(\mathbf{x}_{t-1} ; \mu_\theta\left(\mathbf{x}_t, t\right), \Sigma_\theta\left(\mathbf{x}_t, t\right)\right)$, where $\mu_\theta\left(\mathbf{x}_t, t\right)$ and $\Sigma_\theta(\mathbf{x}_t, t)$ represent the mean and variance of the Gaussian distribution.

\vspace{-1mm}
\paragraph{Diffusion Transformer} (DiT) \citep{peebles2023scalable} exhibits the promising scalability and performance of Transformer~\citep{dosovitskiy2020image,videoworldsimulators2024}. It consists of layers composed of a multi-head self-attention (MHSA) and a multi-layer perceptron (MLP), described as 
\vskip -0.08in
\begin{equation}\label{eq:dit}
\begin{aligned}
    \mathbf{X} &\leftarrow \mathbf{X} + \alpha \text{MHSA}(\gamma \mathbf{X} + \beta), \\
    \mathbf{X} &\leftarrow  \mathbf{X}  + \alpha^{\prime} \text{MLP}(\gamma^{\prime} \mathbf{X} + \beta^\prime),
\end{aligned}
\end{equation}
\vskip -0.08in
\noindent where $\mathbf{X}\!\in\!\mathbb{R}^{N \times C}$ denotes image tokens. Here, $N$ is the token number, and $C$ is the channel dimension. The parameters $\{\alpha, \gamma,\beta, \alpha^{\prime}, \gamma^{\prime}, \beta^{\prime} \}$ are produced by an Adaptive Layer Norm (AdaLN) block \citep{perez2018film}, which takes the class condition embedding $\mathbf{E}_{cls}$ and timestep embedding $\mathbf{E}_t$ as inputs.

\begin{figure*}[t]
    \centering
    \includegraphics[width=\textwidth]{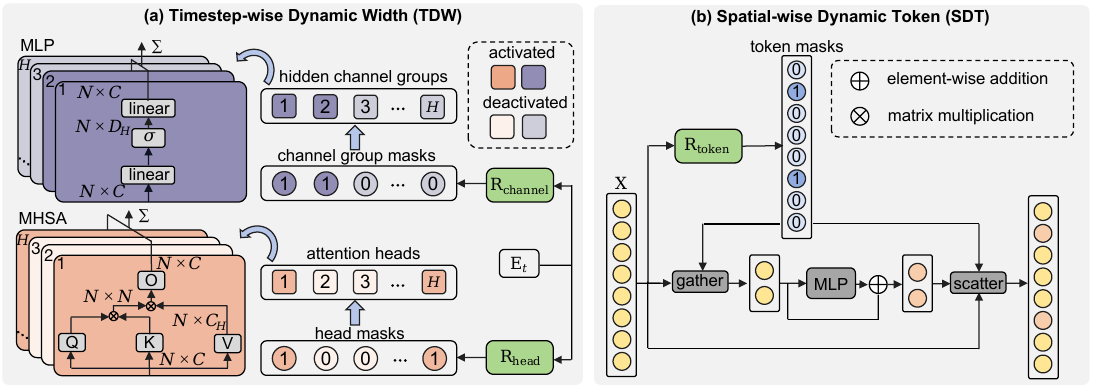}
\caption{\textbf{Overview of the proposed dynamic diffusion transformer (DyDiT)}. }
\label{fig:main}
\vspace{-5mm}
\end{figure*} 

\vspace{-2mm}
\subsection{Timestep-wise Dynamic Width (TDW)} \label{sec:dynamic_channel}
\noindent
As aforementioned, DiT spends equal computation for all timesteps, although different steps share disparate generation difficulty (Figure~\ref{fig:figure1(1)} (a)).  Therefore, the \emph{static} computation paradigm introduces significant redundancy in those ``easy'' timesteps. Inspired by structural pruning methods~\citep{he2017channel, hou2020dynabert, fang2024structural}, we propose a \emph{timestep-wise dynamic width} (\textbf{TDW}) mechanism, which adjusts the width of MHSA and MLP blocks in different timesteps. Note that TDW is \emph{not a pruning method} that permanently removes certain model components, but rather retains the full capacity of DiT and \emph{dynamically} activates different heads/channel groups at each timestep.

\vspace{-1mm}
\paragraph{Heads and channel groups.} Given input $\mathbf{X}\!\in\!\mathbb{R}^{N \times C}$, an MHSA block employs three linear layers with weights $\mathbf{W}_{\text{Q}}, \mathbf{W}_{\text{K}}, \mathbf{W}_{\text{V}}\!\in\!\mathbb{R}^{C \times (H \times C_H)}$ to project it into Q, K, and V representations, respectively. Here, $H$ denotes the head number and $C=\!\!H\!\times\!C_H$. An output linear projection is performed with $\mathbf{W}_{\text{O}}\!\in\!\mathbb{R}^{(H \times C_H) \times C}$ after the attention operation:
\begin{equation}
\begin{aligned}
&\text{MHSA}(\mathbf{X})=\sum
\nolimits_{h=1}^{H} \mathbf{X}_{\text{attn}}^{h} \mathbf{W}_{\text{O}}^{h,:, :}= \\ 
&\sum_{h=1}^{H} (\operatorname{Softmax}((\mathbf{X}\mathbf{W}_{\text{Q}}^{:,h,:}) (\mathbf{X}\mathbf{W}_{\text{K}}^{:,h,:})^{\top}) \mathbf{X} \mathbf{W}_{\text{V}}^{:, h, :}) \mathbf{W}_{\text{O}}^{h,:, :}.
\end{aligned}
\end{equation}

An MLP block contains two linear layers with weights $\mathbf{W}_{1}\!\in\!\mathbb{R}^{C\!\times\!D}$ and $\mathbf{W}_{2}\!\in\!\mathbb{R}^{D\!\times\!C}$, where $D$ is set as $4C$ by default.
To dynamically control the MLP width, we divide $D$ hidden channels into $H$ groups, reformulating the weights into $\mathbf{W}_{1}\!\in\!\mathbb{R}^{C \times (H \times D_{H})}$ and $\mathbf{W}_{2}\!\in\!\mathbb{R}^{(H \times D_{H}) \times C}$. The MLP operation can be formulated as:
\begin{equation}
\begin{aligned}
\text{MLP}(\mathbf{X}) &= \sum\nolimits_{h=1}^{H}\sigma(\mathbf{X}_{\text{hidden}}^h)\mathbf{W}_{2}^{h,:,:} \\
&= \sum\nolimits_{h=1}^{H}\sigma(\mathbf{X} \mathbf{W}_{1}^{:, h, :})\mathbf{W}_{2}^{h,:,:}, 
\end{aligned}
\end{equation}
where $\sigma$ denotes the activation layer.

\vspace{-1mm}
\paragraph{Timestep-aware dynamic width control.}
To selectively activate the necessary heads and channel groups at each diffusion timestep, we feed the timestep embedding $\mathbf{E}_t \in \mathbb{R}^{C}$ into routers $\operatorname{R}_{\text{head}}$ and $\operatorname{R}_{\text{channel}}$ in each block (Figure~\ref{fig:main}(a)). The router is lightweight, which comprises a linear layer followed by the Sigmoid function, producing the probability of each head and channel group being activated:
\begin{equation}~\label{eq:tdw}
\begin{aligned}
\mathbf{S}_{\text{head}} &= \operatorname{R}_{\text{head}}(\mathbf{E}_t)\in[0,1]^H, \\ 
\mathbf{S}_{\text{channel}} &= \operatorname{R}_{\text{channel}}(\mathbf{E}_t)\in[0,1]^H.
\end{aligned}
\end{equation}
A threshold of 0.5 is then used to convert the continuous-valued $\mathbf{S}_{\text{head}}$ and $\mathbf{S}_{\text{channel}}$ into binary masks $\mathbf{M}_{\text{head}}\!\in\!\{0,1\}^H$ and $\mathbf{M}_{\text{channel}}\!\in\!\{0,1\}^H$, indicating the activation decisions for attention heads and channel groups. The $h$-th head (group) is activated only when $\mathbf{M}_{\text{head}}^{h}\!=\!1$ ($\mathbf{M}_{\text{channel}}^{h}\!=\!1$). 

\vspace{-1mm}
\paragraph{Inference.}
After obtaining the discrete decisions  $\mathbf{M}_{\text{head}}$ and $\mathbf{M}_{\text{channel}}$, each DyDiT block only computes the activated heads and channel groups during generation:
\begin{equation} \label{eq:tdw_inf}
    \begin{aligned}
        \text{MHSA}(\mathbf{X})&=\sum_{h: \mathbf{M}_{\text{head}}^h=1} \mathbf{X}_{\text{attn}}^{h} \mathbf{W}_{\text{O}}^{h,:, :},\\
        \text{MLP}(\mathbf{X})&= \sum_{h: \mathbf{M}_{\text{channel}}^h=1} \sigma(\mathbf{X}_{\text{hidden}}^h)\mathbf{W}_{2}^{h,:,:}.
    \end{aligned}
\end{equation}
Let $\tilde{H}_{\text{head}}=\sum_h \mathbf{M}_{\text{head}}^h$ and $\tilde{H}_{\text{channel}}=\sum_h \mathbf{M}_{\text{channel}}^h$ denote the number of activated heads/groups. TDW reduces the MHSA computation from $\mathcal{O}(H\times(4NCC_{H}+2N^2C_{H}))$ to $\mathcal{O}(\tilde{H}_{\text{head}}\times(4NCC_{H}+2N^2C_{H}))$ and MLP blocks from $\mathcal{O}(H\times2NCD_{H})$ to $\mathcal{O}(\tilde{H}_{\text{channel}}\times2NCD_{H})$.
It is worth noting that as the activation choices depend solely on the timestep $\mathbf{E}_t$, we can pre-compute the masks offline once the training is completed, and \emph{pre-define} the activated network architecture before deployment. This avoids the sample-dependent inference graph in traditional dynamic architectures~\citep{meng2022adavit,han2023latency} and facilitates the realistic speedup in batched inference.

\vspace{-3mm}
\subsection{Spatial-wise Dynamic Token (SDT)} \label{sec:dynamic_token}
\noindent
In addition to the timestep dimension, redundancy widely exists in the spatial dimension due to the varying complexity of different patches (Figure~\ref{fig:figure1(1)}(b)). To this end, we propose a spatial-wise dynamic token (SDT) method to reduce computation for the patches where noise estimation is ``easy''.

\paragraph{Token skipping in the MLP block.}
As shown in Figure~\ref{fig:main}(b), SDT adaptively identifies the tokens associated with image regions that present lower noise prediction difficulty. These tokens are then allowed to bypass the computationally intensive MLP blocks. Theoretically, this token-skipping operation can be applied to both MHSA and MLP. However, we find MHSA crucial for establishing token interactions, which is essential for the generation quality. More critically, varying token numbers across images in MHSA might result in incomplete tensor shapes in a batch, reducing the overall throughput. Therefore, SDT is applied only to each MLP block.

Concretely, before an MLP block, the input $\mathbf{X}\!\in\!\mathbb{R}^{N \times C}$ is fed into a token router $\operatorname{R}_{\text{token}}$. This router predicts $\mathbf{S}_{\text{token}} \in \mathbb{R}^{N}$, representing the probability of each token being processed:
\begin{equation}~\label{eq:sdt}
\mathbf{S}_{\text{token}} = \operatorname{R}_{\text{token}}(\mathbf{X})\in[0,1]^N.
\end{equation}
We then convert it into a binary mask $\mathbf{M}_{\text{token}}$ using a threshold of 0.5. Each element $\mathbf{M}^{i}_{\text{token}}\in\{0,1\}$ in the mask indicates whether the $i$-th token should be processed by the block  (if $\mathbf{M}_{\text{token}}^{i} =1$) or directly bypassed (if $\mathbf{M}_{\text{token}}^{i} =0$).

\paragraph{Inference.}
During inference (Figure~\ref{fig:main}(b)), we gather the selected tokens based on the mask $\mathbf{M}_{\text{token}}$ and feed them to the MLP, thereby avoiding unnecessary computational costs for other tokens. Then, we adopt a scatter operation to reposition the processed tokens. This further reduces the computational cost of the MLP block from $\mathcal{O}(\tilde{H}_{\text{channel}}N\times2CD_{H})$ to $\mathcal{O}(\tilde{H}_{\text{channel}}\tilde{N}\times 2CD_{H})$, where $\tilde{N}\!=\!\sum_i \mathbf{M}_{\text{token}}^i$ denotes number of selected tokens to be processed. Since there is no token interaction within the MLP, the SDT operation supports batched inference, improving the practical generation efficiency.

\vspace{-3mm}
\subsection{FLOPs-aware end-to-end Training} \label{sec:training}
\noindent
\paragraph{End-to-end training.}
In TDW, we multiply $\mathbf{M}_{\text{head}}$ and $\mathbf{M}_{\text{channel}}$ with their corresponding features ($\mathbf{X}_{\text{attn}}$ and $\mathbf{X}_{\text{hidden}}$) to zero out the deactivated heads and channel groups, respectively. Similarly, in SDT, we multiply $\mathbf{M}_{\text{token}}$ with $\text{MLP}(\mathbf{X})$ to deactivate the tokens that should not be processed. Following the common practice of dynamic networks~\citep{wang2018skipnet,he2017channel}, Straight-through-estimator \citep{bengio2013estimating} and Gumbel-Sigmoid~\citep{jang2016categorical} are employed to enable the end-to-end training of routers.

\vspace{-1mm}
\paragraph{Training with FLOPs-constrained loss.}
We design a FLOPs-constrained loss to control the computational cost during the generation process. We find it impractical to obtain the entire computation graph during $T$ timesteps since the total timestep $T$ is large \eg $T=1000$. Fortunately, the timesteps in a batch are sampled from $t \sim \text{Uniform}(0, T)$ during training, which approximately covers the entire computation graph. Let $B$ denote the batch size, with $t_b$ as the timestep for the $b$-th sample, we compute the total FLOPs at the sampled timestep, $F^{t_b}_\text{dynamic}$, using masks $\{\mathbf{M}_{\text{head}}^{t_b}, {\mathbf{M}}_{\text{channel}}^{t_b}, \mathbf{M}_{\text{token}}^{t_b} \}$ from each layer, as detailed in Sections~\ref{sec:dynamic_channel} and~\ref{sec:dynamic_token}.  Let $F_{\text{static}}$ denote the total FLOPs of MHSA and MLP blocks in the static DiT. We formulate the FLOPs-constrained loss as: 
\begin{equation} \label{eq:flops}
    \mathcal{L}_\text{FLOPs} = (\frac{1}{B}\sum_{t_b:b\in[1, B]} \frac{F^{t_b}_\text{dynamic}}{F_{\text{static}}} - \lambda)^2,
\end{equation}
where $0<\lambda<1$ is the target FLOPs ratio, and $t_b$ is uniformly sampled from the interval $[0, T]$. The overall training objective combines this FLOPs-constrained loss with the original DiT training loss, expressed as
\begin{equation}\label{eq:loss_overall}
    \mathcal{L} = \mathcal{L}_\text{DiT} + w\mathcal{L}_\text{FLOPs},
\end{equation}
where $w$ is fixed as 1.0 for DyDiT.

\paragraph{Fine-tuning stabilization.}
In practice, we find that directly fine-tuning DyDiT with $\mathcal{L}$ might occasionally lead to unstable training. To address this, we employ two stabilization techniques. First, for a warm-up phase we maintain a complete DiT model supervised by the same diffusion target, introducing an additional item, $\mathcal{L}_{\text{DiT}}^{\text{complete}}$ along with $\mathcal{L}$. After this phase, we remove this item and continue training solely with $\mathcal{L}$. Additionally, prior to fine-tuning, we rank the heads and hidden channels in MHSA and MLP blocks based on a magnitude criterion~\citep{he2017channel}. We consistently select the most important head and channel group in TDW. This ensures that at least one head and channel group is activated in each MHSA and MLP block across all timesteps, thereby alleviating the instability.

\section{Generalization to Flow matching}~\label{sec:flow}
\vspace{-4mm}

\noindent  
Building on DyDiT, DyDiT++ successfully demonstrates its generalization to accelerate flow-matching-based generation. As a similar approach to diffusion models, flow matching-based methods \citep{ma2024sit, esser2024scaling, lipman2022flow, liu2022flow, flux2024} consider learning a continuous interpolant process between the data distribution $x_0 \sim q(x)$ and the noise distribution $x_1 \sim \mathcal{N}(0, I)$. This is typically defined by a time-dependent interpolation path formulated as $x_t = \alpha_t x_0 + \sigma_t x_1$ over $t \in [0, 1]$, where the scaling coefficients $\alpha_t$ and $\sigma_t$ satisfy $\alpha_t + \sigma_t = 1$ with boundary conditions $\alpha_0 = 1$ and $\sigma_1 = 1$. $\alpha_t$ monotonically decreases while $\sigma_t$ increases, ensuring a trajectory from the data manifold at $t=0$ to the noise distribution at $t=1$. The core objective of flow matching-based methods is to learn the velocity field $v_\theta(x_t, t)$ that governs this path through a simulation-free regression loss $\mathcal{L}_\text{velocity} = \mathbb{E}_{t,x_0,x_1} \left\| v_\theta(x_t, t) - \frac{dx_t}{dt} \right\|^2$.

Despite their effectiveness, it remains unclear whether the computational redundancy problem exists in these methods.  To investigate this, we conduct the same experiments from Section~\ref{sec:introduction} on SiT~\citep{ma2024sit}, a representative flow matching-based model that shares the same architecture as DiT~\citep{peebles2023scalable}.

\paragraph{Computation redundancy in flow-matching.}
In Figure~\ref{fig:figure1(1)}(c), we present the loss difference between a smaller model, SiT-S, and a larger model, SiT-XL.  We can observe that although the loss difference pattern in SiT differs from that of DiT, the gap between the small and large models is also uneven across timesteps. This suggests that the difficulty of velocity estimation is not uniform over time, and treating all timesteps equally leads to redundancy.

Furthermore, as shown in Figure~\ref{fig:figure1(1)}(c), we visualize the loss maps from SiT-XL and observe that the difficulty of velocity prediction varies across different spatial regions. This finding indicates the presence of computational redundancy not only along the timestep dimension but also across the spatial dimension.

\vspace{-1mm}
\paragraph{Enable dynamic architecture for SiT.}
Since SiT adopts the same architecture as DiT, both TDW and STD can be seamlessly integrated into SiT. We also adopt Equation~\ref{eq:flops}, replacing $\mathcal{L}_\text{DiT}$ with $\mathcal{L}_\text{velocity}$, as the training objective, to control the FLOPs of SiT.

\vspace{-2mm}
\section{Extension to Video Generation}~\label{sec:videodiffusion}
\vspace{-4mm}

\noindent Beyond class-to-image generation, DiT~\citep{peebles2023scalable} can be extended to video generation~\citep{ma2024latte, zheng2024open, lin2024open, polyak2024movie}.  Since these methods also rely on the diffusion process, they inherit the same efficiency issues from DiT. However, their computational redundancy pattern remains underexplored, making it challenging to directly apply our dynamic architecture.

To this end, in DyDiT++, we first investigate the computational redundancy patterns in Latte~\citep{ma2024latte}, a representative video diffusion transformer architecture, and then demonstrate how our dynamic design can be adapted for video generation.

\vspace{-1mm}
\paragraph{Architecture.}
The input video tokens in Latte can be represented as $\mathbf{X} \in \mathbb{R}^{L \times N \times C}$, where $L$, $N$, and $C$ correspond to the temporal, spatial, and channel dimensions of the video in the latent space, respectively. To jointly capture the spatial and temporal information, Latte iteratively stacks spatial transformer layers and temporal transformer layers. Although all layers share the same formulation as described in Equation~\ref{eq:dit}, the MHSA blocks are applied along the spatial dimension and the temporal dimension for two types of layers, respectively. Additional details are provided in the Supplementary Material.

\vspace{-1mm}
\paragraph{Computation redundancy in video generation.}
To analyze computational redundancy in video generation, we first plot the loss difference between a large model, Latte-XL, and a small model, Latte-S, in Figure~\ref{fig:figure1(1)}(d). This plot reveals a pattern similar to DiT in Figure~\ref{fig:figure1(1)}(a), confirming the presence of timestep-based redundancy in video generation.

Furthermore, as shown in Figure~\ref{fig:figure1(1)}(d), we visualize the loss maps for a video during training. These visualizations illustrate that loss values vary not only across different regions within the same frame but also across corresponding regions between frames, suggesting the unbalance of noise prediction difficulty. Consequently, treating all spatial-temporal regions equally introduces unnecessary computational redundancy.

\vspace{-1mm}
\paragraph{Enable dynamic architecture during video generation.} To reduce timestep-level redundancy, we use routers to dynamically activate heads and channel groups in MHSA and MLP blocks, as described in Equation~\ref{eq:tdw}, for both spatial and temporal transformer layers. Additionally, to address spatial-temporal redundancy in token processing, we introduce routers in Equation~\ref{eq:sdt} to dynamically skip token computations in MLP blocks for both types of layers. Finally, Equation~\ref{eq:flops} is employed to regulate the model's target FLOPs.
This dynamic architecture is referred to as DyLatte. 
Please refer to the Supplementary Material to find additional details.

\section{Extension to T2I Generation}~\label{sec:mmdit}
\vspace{-2mm}

\noindent To enable the text-to-image generation capability in DiT, several variants have been proposed. For instance, PixArt-$\alpha$~\citep{chen2023pixart} incorporates cross-attention blocks into the original DiT architecture to inject textual information during generation. We validated the effectiveness of DyDiT on PixArt-$\alpha$ in the conference paper (1.32$\times$ speedup with comparable FID, as shown in the supplementary material).
Beyond this, the recently proposed MM-DiT~\citep{esser2024scaling, flux2024} represents an advanced transformer design tailored for text-to-image generation tasks. In this section, we detail how DyDiT++ adapts the proposed dynamic architecture to MM-DiT. For illustration, we focus on FLUX~\citep{flux2024}, a representative model recognized for its high-quality generative performance.

\vspace{-2mm}
\subsection{FLUX preliminaries}
\noindent FLUX first adopts CLIP~\citep{radford2021learning} and T5~\citep{raffel2020exploring} to extract text tokens, denoted as $\mathbf{X}_{\text{text}} \in \mathbb{R}^{N_{\text{text}} \times C}$, which subsequently interact with image tokens, $\mathbf{X}_{\text{image}} \in \mathbb{R}^{N_{\text{image}} \times C}$, to incorporate information from the prompt. Such multi-modal input brings a significant difference with DiT: FLUX employs two types of blocks, referred to as DoubleBlocks and SingleBlocks.

\emph{a) DoubleBlocks} retain a similar architecture as DiT blocks, but introduces two key modifications: \emph{(i)} Before MHSA, image and text tokens are concatenated, resulting in $\mathbf{X} \in \mathbb{R}^{(N_{\text{image}} + N_{\text{text}}) \times C}$, allowing the interaction between two modalities. \emph{(ii)} Each DoubleBlock has two parallel MLPs to process text tokens and image separately, without sharing parameters.

\emph{b) SingleBlocks}. After stacking several DoubleBlocks, FLUX concatenates the image and text tokens, and introduces a sequence of SingleBlocks to jointly process the two modalities. The computation in a SingleBlock can be formulated as:
\begin{equation} ~\label{eq:single_layer}
   \mathbf{X} \leftarrow  \mathbf{X} + \text{FC}(\text{MHSA}^{\prime}(\mathbf{X}) || \text{FC}(\mathbf{X})),
\end{equation}
where $\text{MHSA}^{\prime}$ represents the MHSA block without the output projection layer $\text{O}$, $\text{FC}$ denotes a linear layer, and $||$ indicates the concatenation along the channel dimension. Here, we omit the AdaLN blocks for brevity. For more details, please refer to the official implementation~\citep{flux2024}.

\vspace{-2mm}
\subsection{DyFLUX}\label{sec:dyflux}
We adapt both TDW and SDT to DoubleBlocks and SingleBlocks in FLUX. 
We first illustrate the architecture modifications, and then describe the training approach of our DyFLUX.

\paragraph{DoubleBlocks} share a similar architecture with DiT. Therefore, we directly implement TDW and SDT in DoubleBlocks, and make two modifications: \emph{(i)} SDT is only performed for the MLP blocks that process image tokens, since the spatial redundancy mainly exists in the large amounts of image tokens; \emph{(ii)} Before feeding the image tokens to the linear layer in the token router, we perform a modality fusion:
\begin{equation}
\begin{aligned}
    &\mathbf{S}_{\text{token}} = \operatorname{R}_{\text{token}} (\mathbf{X}_\text{image}, \mathbf{X}_\text{text}, \mathbf{E}_t) \\
    &=\text{FC}(\text{AdaLN}(\mathbf{X}_\text{image};\mathbf{E}_t) + \downarrow(\text{AdaLN}(\mathbf{X}_\text{text};\mathbf{E}_t))),
\end{aligned}
\end{equation}
where FC is a linear layer, and $\downarrow$ denotes the pooling operation along the text token dimension.

\vspace{-1mm}
\paragraph{SingleBlocks} have a distinct design which splits the MLP block and interweaves with the MHSA block (Equation~\ref{eq:single_layer}). We modify our method to adapt to this architecture. 

\begin{figure}[t]
    \centering
    \includegraphics[width=0.8\textwidth]{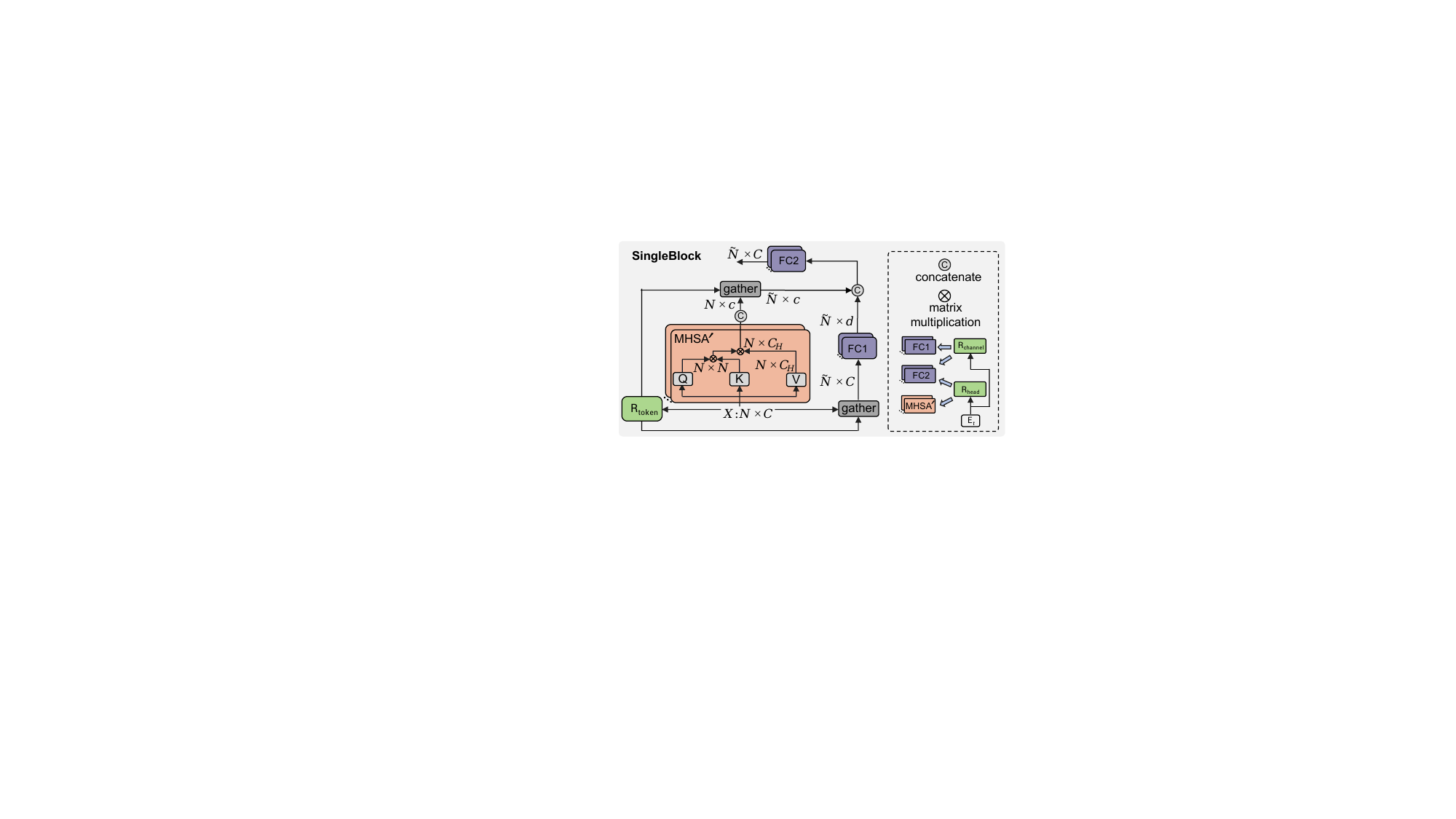}
\caption{
\textbf{Implementation of TDW and SDT in a FLUX SingleBlock.}
The output has a size of $\tilde{N} \times C$. It can then be scattered and added to the input $\mathbf{X} \in \mathbb{R}^{N \times C}$. We omit this scatter-add operation for brevity. Note that the width of $\text{FC2}$ is determined by both $\operatorname{R}_{\text{head}}$ and $\operatorname{R}_{\text{channel}}$.
} 
\label{fig:t2i_flops}
\end{figure}

As shown in Figure~\ref{fig:t2i_flops}, we first perform TDW to MHSA$^\prime$, whose output $\mathbf{X}_{\text{attn}} \in \mathbb{R}^{N \times c}$ has a reduced channel dimension $c=\tilde{H}_{\text{head}} \times C_{H}$ (there is no output projection in MHSA$^\prime$), where $\tilde{H}_{\text{head}}\le H_{\text{head}}$ denotes the number of selected heads.

For the MLP layers, we perform token selection before its first linear layer $\text{FC1}$. $\tilde{N}$ tokens are selected by the token router $\operatorname{R}_{\text{token}}$ and passed into $\text{FC1}$. In $\text{FC1}$, $\tilde{H}_{\text{channel}}$ channel groups are activated, producing an intermediate feature of size $\tilde{N} \times d$, where $d = \tilde{H}_{\text{channel}} \times D_{H}$. This intermediate feature is then concatenated with the corresponding tokens in $\mathbf{X}_{\text{attn}}$ along the channel dimension and passed into the second linear layer, $\text{FC2}$, resulting in the output feature of size $\tilde{N} \times C$. This process reduces the computational cost of the two linear layers by dynamically adapting both width and token usage. Finally, the output from $\text{FC2}$ is scattered and added back to $\mathbf{X}$ to produce the final output of the SingleBlock.

\paragraph{Computational cost.}
We observe that TDW reduces the computational cost of the $\text{MHSA}^{\prime}$ from $\mathcal{O}(H\times(3NCC_{H}+2N^2C_{H}))$ to $\mathcal{O}(\tilde{H}_{\text{head}}\times(3NCC_{H}+2N^2C_{H}))$. Meanwhile,  the computation in $\text{FC1}$ is reduced from $\mathcal{O}(H\times NCD_{H})$ to $\mathcal{O}(\tilde{H}_{\text{channel}}\times \tilde{N}CD_{H})$. 
For the second linear layer, $\text{FC2}$, the computational cost is reduced from $\mathcal{O}(N \times (H_{\text{head}}C_{H} + H_{\text{channel}}D_{H})\times C)$ to $\mathcal{O}(\tilde{N} \times (\tilde{H}_{\text{head}}C_{H} + \tilde{H}_{\text{channel}}D_{H})\times C)$.

\paragraph{Training.} We divide our training into two phases. 

\emph{a) Distillation-based training.} Due to the complexity of the text-to-image task, we develop a distillation technique to train our DyFLUX. Specifically, let $\mathbf{X}^{\ell}$ denote the output tokens of the $\ell$-th block and $\mathbf{Y}$ denote the output of the overall network, we align these representations of our DyFLUX with those of the static FLUX. The loss item can be written as
\begin{equation}\label{eq:distill}        \mathcal{L}_\text{distill}=u\sum\nolimits_{\ell\in\mathbb{L}}\text{MSE}(\mathbf{X}^{\ell}_\text{d}, \mathbf{X}^{\ell}_\text{s}) + v\text{MSE}(\mathbf{Y}_\text{d}, \mathbf{Y}_\text{s}),
\end{equation}
where MSE is the mean-square loss. The subscripts ``d'' and ``s'' denote dynamic and static, and $u, v$ are coefficient hyperparameters, which are fixed as 0.0001 and 0.1, respectively. For efficiency, we select every 4-th block to construct a subset $\mathbb{L}$ of distilled block indices. As shown in Figure~\ref{fig:flux_distill_abl}, this distillation technique significantly improves the generation quality. The overall loss function is the combination of Equation~\ref{eq:loss_overall} ($w$ set to 5.0) and Equation~\ref{eq:distill}.

\emph{b) Classifier-free guidance distillation.} After the first-phase training, we further perform a CFG-distillation to reduce the explicit forward procedure with a negative prompt. Following similar settings in \citet{kong2024hunyuanvideo}, we construct a linear combination of a conditional and an unconditional output with a parameter-frozen DyFLUX (teacher). Then, a guidance embedding layer is used to encode the guidance scale for the student DyFLUX. An MSE loss is adopted to enable the student to directly generate the conditioned output with one single forward pass.

\begin{figure*}[t]
    \centering
    \includegraphics[width=\textwidth]{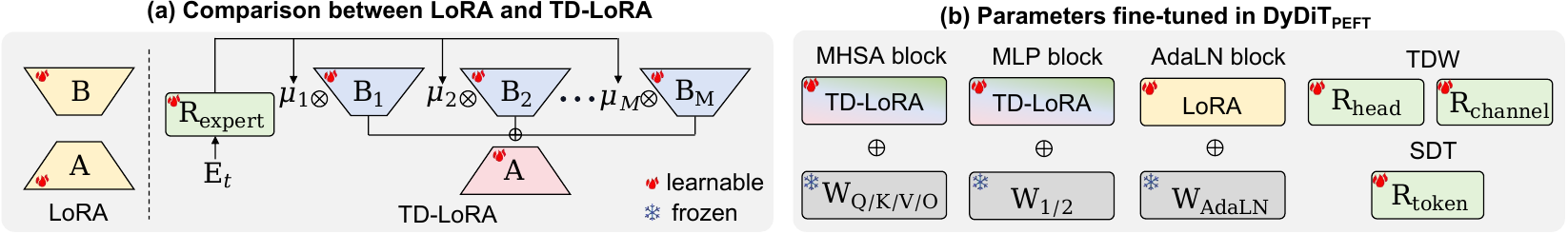}
\caption{(a) \textbf{Comparison between the original LoRA and the proposed TD-LoRA.} We introduce $M$ expert matrices to replace $\mathbf{B}$ in the original LoRA. (b) \textbf{Fine-tuning specific parameters in DyDiT$_{\text{PEFT}}$}. 
} 
\label{fig:dynamic_lora}
\end{figure*}

\section{Improvement for Training Efficiency} \label{sec:dtlora}
\noindent In our conference paper, DyDiT needs to be trained in a full-finetuning manner for reduced inference cost. Nevertheless, training efficiency is also essential, particularly in resource-constrained scenarios.  To address this, DyDiT++ incorporates parameter-efficient fine-tuning (PEFT) techniques, such as LoRA~\citep{hu2021lora}, into DyDiT's static-to-dynamic adaptation process, aiming to achieve both training and inference efficiency.

\paragraph{LoRA Preliminaries.} Let $\mathbf{W} \in \mathbb{R}^{C_{\text{in}} \times C_{\text{out}}}$ denote the weight to be fine-tuned in a layer (\emph{e.g.} Q, K, V projection). Parameter-efficient fine-tuning with LoRA can be expressed as:
\begin{equation}
    \mathbf{X} \mathbf{W}^{\prime}  = \mathbf{X} (\mathbf{W} + \mathbf{A}\mathbf{B})  =  \mathbf{X}\mathbf{W} +  (\mathbf{X} \mathbf{A})\mathbf{B}
\end{equation}
where $\mathbf{A} \in \mathbb{R}^{C_{\text{in}} \times r}$ and  $\mathbf{B} \in \mathbb{R}^{r \times C_{\text{out}}}$ are learnable low-rank matrices, with the rank of $r$. During fine-tuning, the original weight $\mathbf{W}$ remains frozen, and we usually have $r \ll C_{\text{in}}$ and $r \ll C_{\text{out}}$, thereby significantly reducing memory cost.

\paragraph{Trainable Parameters in DyDiT.}
Before applying LoRA, we categorize the parameters that should be fine-tuned during the static-to-dynamic adaptation process into three main groups:

\emph{(i)} Router parameters: core components to perform our dynamic computation. We can fully fine-tune all of them, as they only account for less than 0.5\% of the total parameters, making their computational cost negligible.

\emph{(ii)} AdaLN parameters $\mathbf{W}_{\text{AdaLN}}$, crucial for condition injection during the generation process. We can leverage LoRA to fine-tune them, thereby improving parameter efficiency as this part accounts for approximately 33\% of DiT's total parameters.

\emph{(iii)} Core Transformer parameters, including parameters in MHSA ($\mathbf{W}_{\text{Q}}$, $\mathbf{W}_{\text{K}}$, $\mathbf{W}_{\text{V}}$, $\mathbf{W}_{\text{O}}$) and MLP ($\mathbf{W}_{\text{1}}$, $\mathbf{W}_{\text{2}}$), which form the primary components of the original DiT. For this group of parameters, since heads and channel groups are dynamically selected based on Equation~\ref{eq:tdw_inf}, different parts of these parameters are activated at different timesteps. 
As a result, directly applying the same LoRA across all timesteps overlooks this property, leading to inefficient utilization of trainable parameters and, consequently, suboptimal performance, as demonstrated in Section~\ref{sec:peft_exp}.

\paragraph{Timestep-based dynamic LoRA.}
To address the aforementioned problem, we propose a timestep-based dynamic LoRA (TD-LoRA), inspired by MoE~\citep{cai2024survey}. It adaptively adjusts the low-rank matrices in LoRA based on timesteps, as illustrated in Figure~\ref{fig:dynamic_lora}(a). Specifically, we introduce $M$ expert matrices to replace $\mathbf{B}$ in original LoRA, which can be expressed as: 
\begin{equation}
    \mathbf{X}\mathbf{W}^{\prime}  = \mathbf{X}\mathbf{W} + (\mathbf{X}\mathbf{A}) (\sum_{i=1}^{M}\mu_{i} \mathbf{\hat{B}}_{i}),
\end{equation}
where $\mathbf{A} \in \mathbb{R}^{C_{\text{in}} \times \hat{r}}$
and $\mathbf{\hat{B}}_{i} \in \mathbb{R}^{\hat{r} \times C_{\text{out}}}$.  By employing $M$ expert matrices, we can reduce $\hat{r}$ to be smaller than $r$, ensuring that the total number of parameters remains approximately the same. The weight scores $\mu_{i}$, used to fuse the expert matrices, are determined dynamically based on the diffusion timestep:
\begin{equation}
\begin{aligned}
\mathbf{\mu} &= \operatorname{Softmax}(\operatorname{R}_{\text{expert}}(\mathbf{E}_t)) \in \mathbb{R}^{M},
\end{aligned}
\end{equation}
where $\operatorname{R}_{\text{expert}}$ is a router taking the timestep embedding $\mathbf{E}_t$ as input, achieving timestep-aware LoRA parameters.

It is worth noting that the proposed TD-LoRA is applied exclusively to the parameters in group (iii), while using the original LoRA for the parameters in (ii) and full fine-tuning for the parameters in (i), resulting in DyDiT$_{\text{PEFT}}$. Figure~\ref{fig:dynamic_lora}(b) illustrates how the specific parameters are fine-tuned.

The proposed TD-LoRA also includes two alternative variants: one replaces the matrix $\mathbf{A}$ in the original LoRA, referred to as Inverse TD-LoRA, while the other replaces both $\mathbf{A}$ and $\mathbf{B}$ with multiple experts, referred to as Symmetry TD-LoRA. Empirical results in Section~\ref{sec:peft_exp} demonstrate that our proposed solution outperforms these alternatives.

\paragraph{Inference.}
During inference, we first compute the weighting scores based on the timestep to fuse the expert matrices. Then, the LoRA parameters are fused with the original weight $\mathbf{W}$ for subsequent computations. One potential concern is that the weights in TD-LoRA cannot be pre-fused with the original weights prior to inference, which might introduce additional latency. However, due to the batched inference in our model, the latency introduced by computing the weighting scores and performing weight fusion is negligible compared to the overall computation time for processing a batch of samples. As a result, our method achieves a generation speed comparable to the original DyDiT, as demonstrated in Table~\ref{tab:compare_LoRA}.

\begin{table*}[t]
\centering
\scriptsize
\caption{\textbf{Comparison with diffusion models on ImageNet of 256$\times$256 and 512$\times$512 resolutions}. \textbf{Bold font} and \underline{underline} denote the best and the second-best performance, respectively.
}
\tablestyle{10pt}{1.1}
 \begin{tabular}
{c  c c c c c c  c }
    Model & Params. (M) $\downarrow$   & FLOPs (G) $\downarrow$  & FID $\downarrow$ & sFID $\downarrow$ & IS $\uparrow$ & Precision $\uparrow$ & Recall $\uparrow$ \\
  \midrule[1.2pt]
  \multicolumn{8}{c}{\emph{Static $256 \times 256$}} \\
    ADM & 608 & 1120 & 4.59 & 5.25 & 186.87 & 0.82 & 0.52 \\
    LDM-4 & 400 & 104 & 3.95 & - & 178.22 & 0.81 & 0.55 \\
    U-ViT-L/2 & \textbf{287} & \underline{77} & 3.52 & - & - & - & -  \\
    U-ViT-H/2 & 501 & 113 & 2.29 & - & 247.67 & \textbf{0.87} & 0.48  \\
    DiffuSSM-XL & 673 & 280 & 2.28 & \textbf{4.49} & 269.13 &\underline{0.86} & 0.57 \\
    DiM-L & \underline{380} & 94 & 2.64 & - & - & - & - \\
    DiM-H & 860 & 210 & 2.21 & - & - & - & - \\
    DiffiT  &  561 & 114	& \underline{1.73} & - &276.49 & 0.80 & \underline{0.62} \\ 
    DiMR-XL & 505 & 160 & \textbf{1.70} & -& \textbf{289.00} & 0.79 & \textbf{0.63}\\
    \hshline
    DiT-L & 468 & 81 & 5.02 &- & 167.20 & 0.75 & 0.57 \\
    DiT-XL & 675 & 118 & 2.27 & 4.60 & 277.00 & 0.83 & 0.57 \\ 
\hshline

\multicolumn{8}{c}{\emph{Dynamic $256 \times 256$}} \\
   \rowcolor{gray!15} DyDiT-XL$_{\lambda=0.7}$ &  678 & 84.33 (\textcolor{blue}{$\downarrow$1.40$\times$})  & 2.12 & 4.61 & \underline{284.31} & 0.81 & 0.60  \\
   \rowcolor{gray!15} DyDiT-XL$_{\lambda=0.5}$ &  678 & \textbf{57.88  (\textcolor{blue}{$\downarrow$2.04$\times$})}  & 2.07 & \underline{4.56} & 248.03 & 0.80 & 0.61  \\ 

  \midrule[1.2pt]
  \multicolumn{8}{c}{\emph{Static $512 \times 512$}}\\

ADM-G	& 731	& 2813	& 3.85	& 5.86 & 221.72	& \underline{0.84}	& 0.53 \\
DiffuSSM-XL & 673	& 1066	& 3.41	& -	& \textbf{255.00}	& \textbf{0.85} & 0.49 \\
DiM-Huge	&	860	& 708	&3.78	&-	&-	&-	& - \\ \hshline

DiT-XL & \textbf{675}	& \underline{514}	& 3.04	& \textbf{5.02}	& \underline{240.80}	& \underline{0.84}	& \underline{0.54} \\ \hshline

\multicolumn{8}{c}{\emph{Dynamic $512 \times 512$}}\\
 \rowcolor{gray!15} DyDiT-XL$_{\lambda=0.7}$ &  \underline{678} & \textbf{375.05 (\textcolor{blue}{$\downarrow$1.37$\times$})}	& \textbf{2.88}	& \underline{5.14}	& 228.93	& 0.83	& \textbf{0.56}  \\

\end{tabular}
\label{fig:sota}
\vspace{-2mm}
\end{table*}

\begin{figure*}[htbp]
    \centering
    \includegraphics[width=0.337\textwidth]{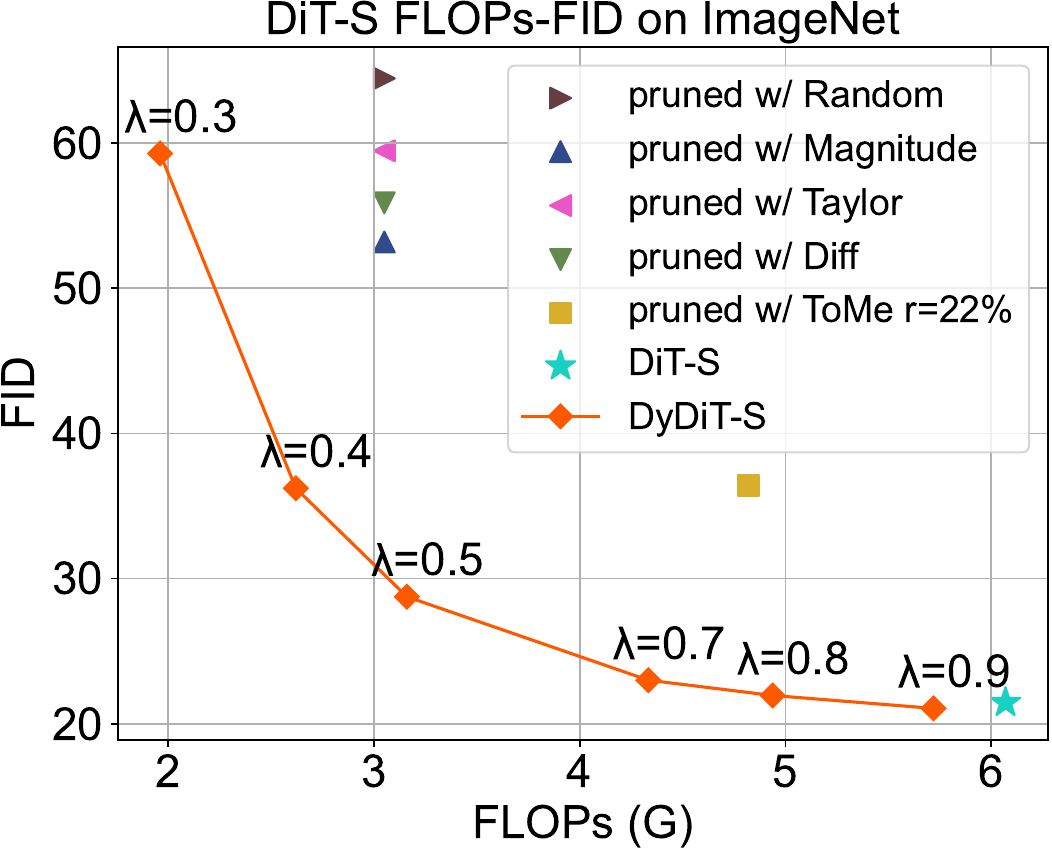}
    \hfill %
    \includegraphics[width=0.321\textwidth]{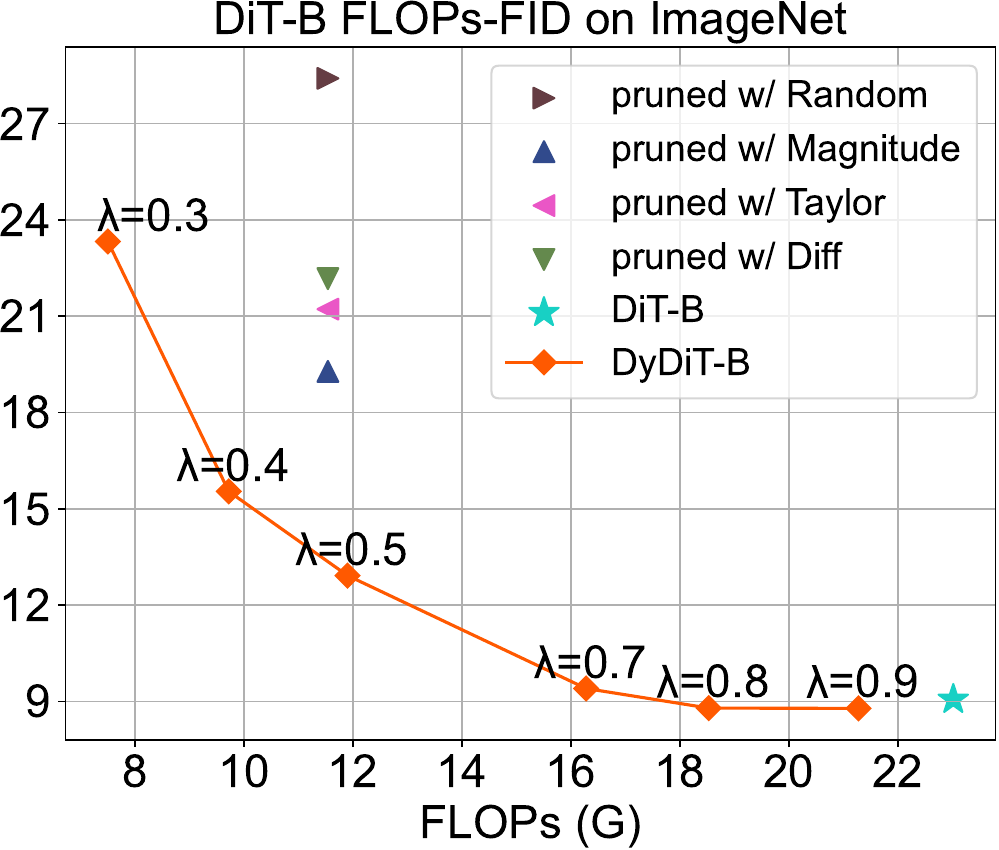}
    \hfill %
    \includegraphics[width=0.325\textwidth]{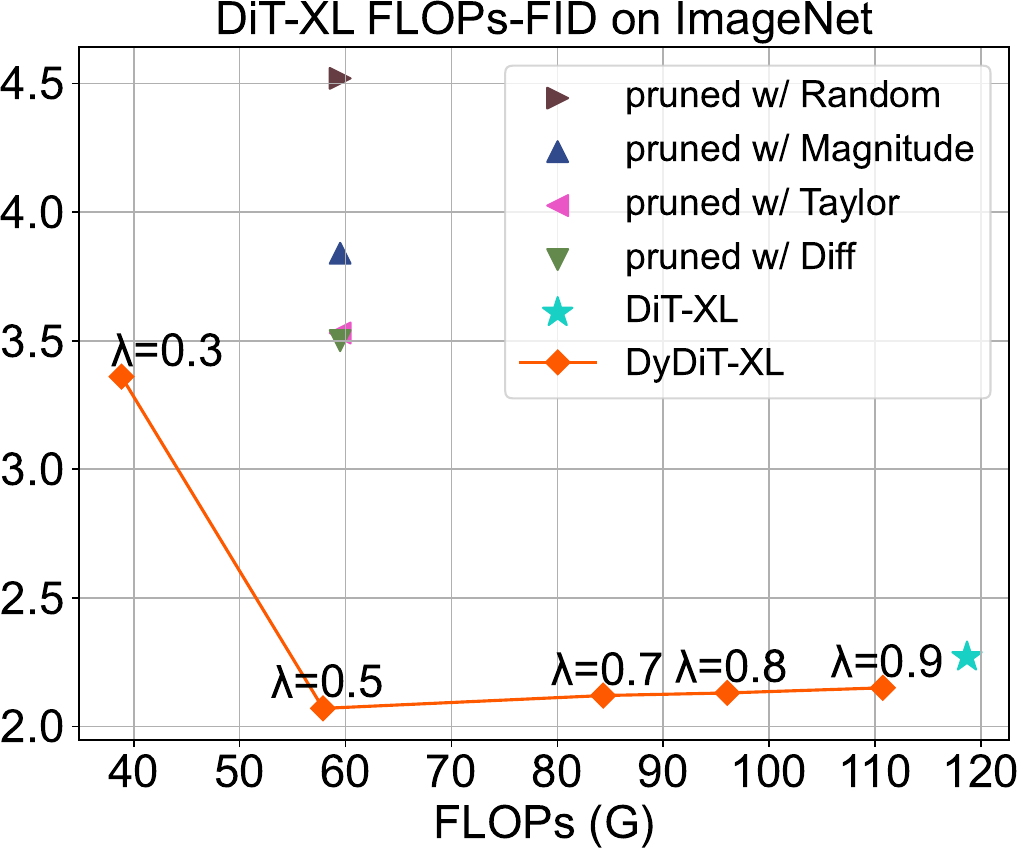}
    \caption{\textbf{FLOPs-FID trade-off on ImageNet.}}
    \label{fig:flops_fid_figure}
\end{figure*}

\begin{figure}[t]
    \centering
    \includegraphics[width=0.45\textwidth]{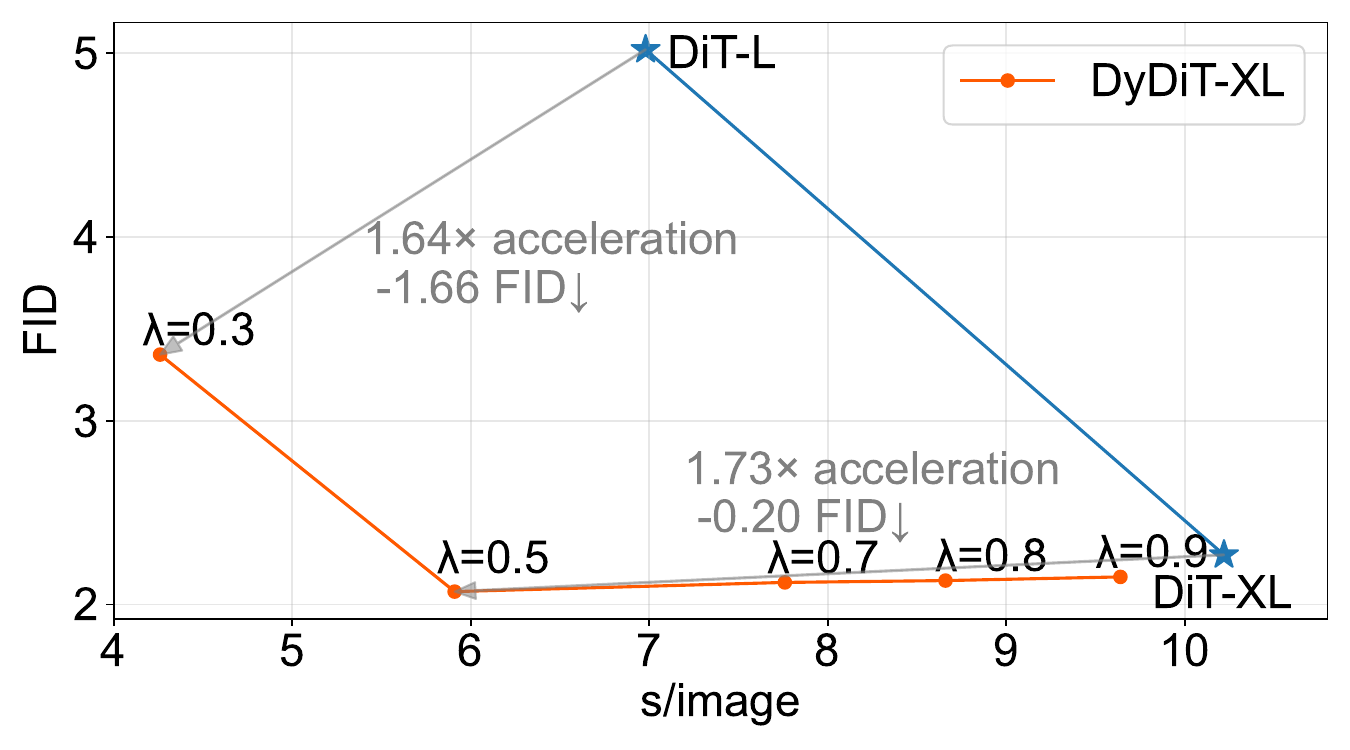}
\caption{
\textbf{Realistic speedup.} DyDiT achieves a better trade-off between speed and FID compared to the static DiT family.
} 
\label{fig:speed}
\end{figure}

\section{Experiments in DyDiT}
\vspace{-1mm}
\paragraph{Implementation details.}
Our DyDiT can be built easily by fine-tuning on pre-trained DiT weights\footnote{We finetune pre-trained models to prevent ``reinventing the wheels''. The effectiveness of DyDiT is not limited to the pretrain-finetune paradigm. The empirical study of training-from-sratch is provided in Supplementary Material.}. We experiment on three different-sized DiT models denoted as DiT-S/B/XL. For DiT-XL, we directly adopt the checkpoint from the official DiT repository~\citep{peebles2023scalable}, while for DiT-S and DiT-B, we use pre-trained models provided in \citet{pan2024t}.
All experiments are conducted on a server with 8$\times$NVIDIA A800 80G GPUs. More details of model configurations and training setup can be found in Supplementary Material.
Following DiT~\citep{peebles2023scalable}, the strength of classifier-free guidance~\citep{ho2022classifier} is set to 1.5 and 4.0 for evaluation and visualization, respectively. Unless otherwise specified, 250 DDPM \citep{ho2020denoising} sampling steps are used. All speed tests are performed on an NVIDIA V100 32G GPU.

\vspace{-0.5mm}
\paragraph{Datasets.} Following DiT \citep{peebles2023scalable}, we mainly conduct experiments on ImageNet~\citep{deng2009imagenet} at 256$\times$256 resolution. We also assess DyDiT on four fine-grained datasets (256$\times$256) used by \citep{xie2023difffit}: Food \citep{bossard2014food}, Artbench \citep{liao2022artbench}, Cars \citep{gebru2017fine} and Birds \citep{wah2011caltech}. We conduct experiments in both in-domain fine-tuning and cross-domain transfer learning manners.

\vspace{-1mm}
\paragraph{Metrics.}
Following \citet{peebles2023scalable, teng2024dim}, we sample 50,000 images to measure the Fréchet Inception Distance (FID) \citep{heusel2017gans} with the ADM’s TensorFlow evaluation \citep{dhariwal2021diffusion}. Inception Score (IS) \citep{salimans2016improved}, sFID \citep{nash2021generating}, and Precision-Recall \citep{kynkaanniemi2019improved} are also reported.

\vspace{-1mm}
\subsection{Comparison with State-of-the-Art Diffusion Models}
\noindent
In Table~\ref{fig:sota}, we compare DyDiT with competitive \emph{static} architectures, including ADM~\citep{dhariwal2021diffusion}, LDM~\citep{rombach2022high}, U-ViT~\citep{bao2023all}, DiffuSSM~\citep{yan2024diffusion}, DiM~\citep{teng2024dim}, DiffiT~\citep{hatamizadeh2025diffit}, DiMR~\citep{liu2024alleviating} and DiT~\citep{peebles2023scalable} on ImageNet of 256$\times$256 and  512$\times$512 resolutions. 
DyDiT is fine-tuned with $<$3\% iterations based on DiT.

On the standard 256$\times$256 setting, DyDiT$_{\lambda=0.5}$ notably achieves a \textbf{2.07} FID score with $<$\textbf{50\%} FLOPs of its counterpart, DiT-XL, and significantly outperforms most models.  This verifies that our method can effectively remove the redundant computation in DiT and maintain the generation performance. It accelerates the generation speed by \textbf{1.73}$\times$ (the detailed speed tests are presented in the Supplementary Material). Increasing the target FLOPs ratio $\lambda$ from 0.5 to 0.7 enables DyDiT$_{\lambda=0.7}$ to achieve competitive performance with DiT-XL across most metrics and obtains the best IS score. This improvement is likely due to DyDiT's dynamic architecture, which offers superior flexibility compared to static architectures, allowing the model to address each timestep and image patch specifically during the generation process. With $\sim$80G FLOPs, DyDiT$_{\lambda=0.7}$  outperforms U-ViT-L/2 and DiT-L, further validating the advantages of our dynamic generation paradigm. For 512$\times$512 resolution, our method achieves performance comparable to the baseline model, DiT-XL, while significantly reducing FLOPs. This demonstrates the effectiveness of DyDiT in high-resolution generation.

\vspace{-1mm}
\subsection{Comparison with Pruning Methods}
\vspace{-1mm}
\paragraph{Benchmarks.} Our DyDiT improves efficiency from the aspects of \emph{architecture} and \emph{token} redundancy. To evaluate the superiority of its \emph{dynamic} paradigm, we compare DyDiT against competitive \textit{static} and token pruning techniques.

\emph{Pruning.} We include Diff pruning \citep{fang2024structural} in the comparison, which is a Taylor-based~\citep{molchanov2016pruning} pruning method specifically optimized for the diffusion process and has demonstrated superiority on diffusion models with U-Net~\citep{ronneberger2015u} architecture \citep{fang2024structural}. Following \citep{fang2024structural}, we also include Random pruning, Magnitude pruning \citep{he2017channel}, and Taylor pruning \citep{molchanov2016pruning} in the comparison. We adopt these four pruning approaches to distinguish important heads and channels in DiT from less significant ones, which can be removed to reduce the model's \emph{runtime} width.

\emph{Token merging.} We also compare DyDiT with ToMe \citep{bolya2022token}, which progressively reduces the token number by merging tokens with high similarities in ViT \citep{dosovitskiy2020image}. Its enhanced version \citep{bolya2023token} can also accelerate Stable Diffusion \citep{rombach2022high}. 

\begin{table*}[t]
\centering
\scriptsize
\caption{\textbf{Results on fine-grained datasets.} The mark $\dag$ corresponds to fine-tuning directly on the target dataset.} 
\tablestyle{10pt}{1.1}
\begin{tabular}{c  c c c c c c c}
    \multirow{2}{*}{Model} & \multirow{2}{*}{s/image $\downarrow$} & \multirow{2}{*}{FLOPs (G) $\downarrow$}  & \multicolumn{5}{c}{FID $\downarrow$}  \\ 
    &  &   & \multicolumn{1}{c}{Food}  & \multicolumn{1}{c}{Artbench}  & \multicolumn{1}{c}{Cars}  & \multicolumn{1}{c}{Birds} & \multicolumn{1}{c}{\#Average} \\
  \midrule[1.2pt]
DiT-S & 0.65  & 6.07  & \underline{14.56} & \textbf{17.54} &  \textbf{9.30} & \textbf{7.69} & \textbf{12.27} \\ 
\hshline
pruned w/ random & 0.38  & 3.05  & 45.66  & 76.75 & 60.26 & 48.60 & 57.81 \\ 
pruned w/ magnitude &  0.38 & 3.05 & 41.93  & 42.04 & 31.49 & 26.45 & 35.44\\ 
pruned w/ taylor &  0.38 & 3.05 & 47.26   & 74.21 & 27.19 & 22.33 & 42.74 \\ 
pruned w/ diff &  0.38 & 3.05  &  36.93 &  68.18 & 26.23 & 23.05 &  38.59 \\ 
pruned w/ ToMe 20\% &  0.61  & 4.82  & 43.87  & 62.96 & 32.16 & 15.20 & 38.54 \\ 
\hshline
\rowcolor{gray!15} DyDiT-S$_{\lambda=0.5}$ & 0.41  & 3.16$_{\textcolor{blue}{\downarrow1.92\times}}$ & 16.74  & 21.35 & \underline{10.01} & \underline{7.85}  &  13.98 \\ 
 DyDiT-S$_{\lambda=0.5} \dag$  & 0.41 & 3.17$_{\textcolor{blue}{\downarrow1.91\times}}$ & \textbf{13.03} & \underline{19.47}  & 12.15  & 8.01 & \underline{13.16} \\
    \label{tab:fine_grained}
    \end{tabular}
\end{table*}

\begin{table*}[t]
\centering
\tablestyle{10pt}{1.1}
\caption{\textbf{Ablation Study} on DyDiT-S$_{\lambda=0.5}$. All models evoke around 3.16 GFLOPs.}
\begin{tabular}{c | c c |c c c c c c}

    \multirow{2}{*}{Model} & \multirow{2}{*}{TDW} & \multirow{2}{*}{SDT}  & \multicolumn{6}{c}{FID $\downarrow$}  \\ 
    &  &  & \multicolumn{1}{c}{ImageNet} & \multicolumn{1}{c}{Food}  & \multicolumn{1}{c}{Artbench}  & \multicolumn{1}{c}{Cars}  & \multicolumn{1}{c}{Birds} & \multicolumn{1}{c}{\#Average} \\
  \midrule[1.2pt]
I &  \ding{51} & & 31.89 & \textbf{15.71}  & 28.19 & 19.67 & 9.23 & 20.93 \\
II &  & \ding{51} & 70.06 & 23.79 & 52.78 & 16.90 & 12.05 & 35.12 \\
\rowcolor{gray!15}  III &  \ding{51} & \ding{51} & \textbf{28.75} & \underline{16.74} & \textbf{21.35} & \textbf{10.01} & \textbf{7.85} &  \textbf{16.94} \\
\hshline
I (random)  & & & 124.38  & 111.88 & 151.99 & 127.53 & 164.29 & 136.01 \\
I (manual)  & & & 34.08  & 23.89 & 40.02  & 22.34 & 20.17 & 28.10 \\
III (layer-skip) &  \ding{51} &   & \underline{30.95} & 17.75 & \underline{23.15} &\underline{10.53} & \underline{9.01} & \underline{18.29} \\
    
    \end{tabular}
    \label{tab:ablation_study}
\end{table*}

\vspace{-1mm}
\paragraph{Results.}
We present the FLOPs-FID curves for S, B, and XL size models in Figure~\ref{fig:flops_fid_figure}. DyDiT significantly outperforms all pruning methods with similar or even lower FLOPs, highlighting the superiority of dynamic architecture over static models.

Interestingly, Magnitude pruning shows slightly better performance among structural pruning techniques on DiT-S and DiT-B, while Diff pruning and Taylor pruning perform better on DiT-XL. This indicates that different-sized DiT prefer distinct pruning criteria. Although ToMe \citep{bolya2023token} successfully accelerates U-Net models with acceptable performance loss, its application to DiT results in performance degradation, as also observed in~\citet{moon2023early}. We conjecture that the errors introduced by token merging become irrecoverable in DiT due to the absence of convolutional layers and long-range skip connections present in U-Net architectures. Therefore, we omit ToMe's performance on DiT-B and DiT-XL in Figure~\ref{fig:flops_fid_figure}.

\vspace{-1mm}
\paragraph{Scalability.} We can observe from Figure~\ref{fig:flops_fid_figure} that the performance gap between DyDiT and DiT diminishes as model size increases. Specifically, DyDiT-S achieves a comparable FID to the original DiT only at $\lambda=0.9$, while DyDiT-B achieves this with a lower FLOPs ratio, \eg $\lambda=0.7$. When scaled to XL, DyDiT-XL attains a slightly better FID even at $\lambda=0.5$. This is due to increased computation redundancy with larger models, allowing our method to reduce redundancy without compromising FID. These results validate the scalability of our approach, which is crucial in the era of large models. We also conduct additional experiments to evaluate scalability in the training-from-scratch setting. The results of these experiments are provided in the Supplementary Material.

\vspace{-1mm}
\subsection{Generation Speed}
\noindent To validate the realistic speedup of DyDiT, we plot the FID-latency (on V100 GPU) curves of DyDiT and the original DiT family~\citep{peebles2023scalable} in Figure~\ref{fig:speed}. The results demonstrate that the theoretical efficiency of DyDiT (Figure~\ref{fig:flops_fid_figure}) successfully translates into realistic speedup on GPU, thanks to our hardware-friendly design. Furthermore, DyDiT, with varying target FLOPs ratios ($\lambda$), achieves a better trade-off between generation speed and FID score compared to the original DiT family, further validating the effectiveness of our approach. More detailed results of speed tests are provided in the Supplementary Material.

In Table~\ref{tab:device_speed}, we extend our analysis to include additional hardware platforms, providing a broader evaluation of speedups. We observe that on devices with limited computational capabilities, such as the CPU and GPUs like the Apple M2 and NVIDIA V100, our method achieves significant acceleration, demonstrating the superiority of DyDiT in resource-constrained environments. When it comes to high-performance devices like the NVIDIA A100, the acceleration ratio tends to become less pronounced, as the proportion of time spent on memory access increases relative to computation. This observation demonstrates potential opportunities for further optimization in future research.

\begin{table}[t]
\centering
\scriptsize
\tablestyle{10pt}{1.2}

\caption{\textbf{Acceleration ratios across  hardware platforms.} }
\begin{tabular}{lccc}
\multirow{2}{*}{Type} & \multirow{2}{*}{Hardware} & \multicolumn{2}{c}{s/image$\downarrow$} \\
{} & {} & {DiT-XL} & {DyDiT-XL$_{\lambda=0.5}$} \\ 
\midrule[1.2pt]
CPU & AMD EPYC 7R13                 & 93.82      & 58.24$_{\textcolor{blue}{\downarrow1.61\times}}$              \\
\midrule
\multirow{3}{*}{GPU} 
    & Apple M2                      &  139.03    & 72.91$_{\textcolor{blue}{\downarrow1.91\times}}$      \\
    & NVIDIA V100                   &  10.22     &  5.91$_{\textcolor{blue}{\downarrow1.73\times}}$      \\
    & NVIDIA A100                   &  1.81      &  1.26$_{\textcolor{blue}{\downarrow1.43\times}}$              \\
\end{tabular}
    \label{tab:device_speed}
\end{table}

\begin{figure}[t]
    \centering
    \includegraphics[width=0.9\textwidth]{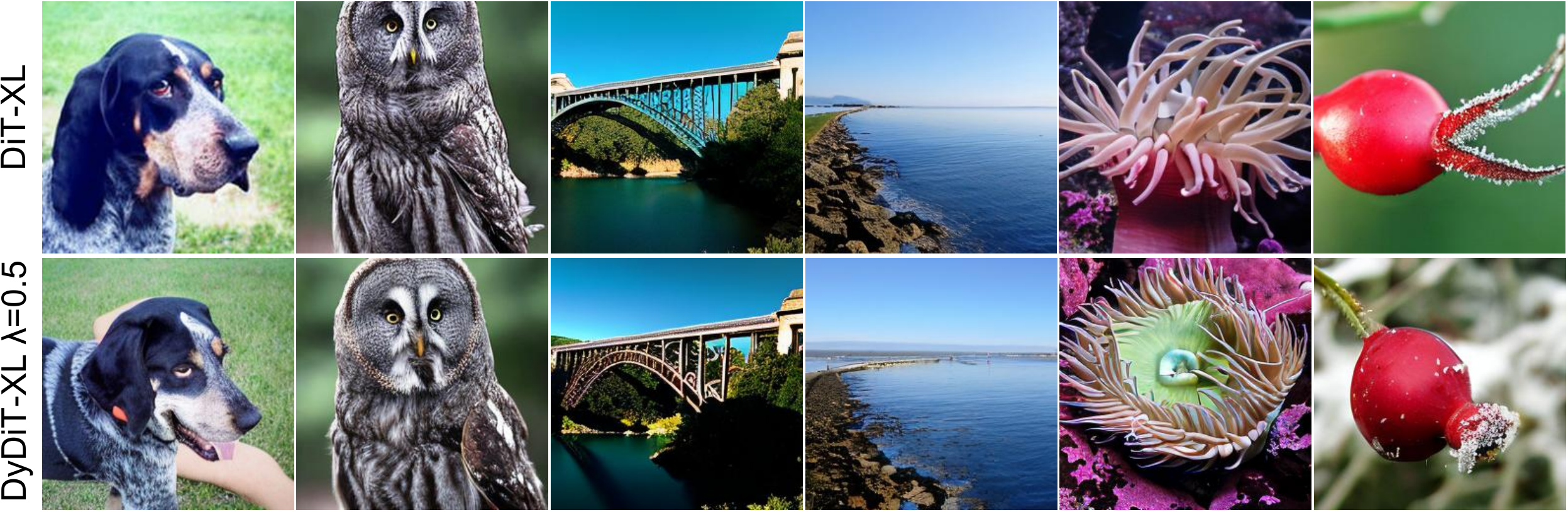}
\caption{\textbf{Visual Comparison between DiT and DyDiT.} Images generated on ImageNet at a 256$\times256$ resolution.}
\label{fig:imagenet_vis_compare}  
\end{figure} 

\subsection{Results on fine-grained datasets}
\vspace{-1mm}
\paragraph{Quantitative results.}
We further compare DyDiT with structural pruning and token pruning approaches on fine-grained datasets under the in-domain fine-tuning setting, where DiT is first pre-trained on the target dataset and subsequently fine-tuned for pruning or dynamic adaptation. 
As presented in Table~\ref{tab:fine_grained}, our method with $\lambda=0.5$ FLOPs ratio significantly reduces computation and improves generation speed while maintaining performance levels comparable to the original DiT. To ensure fair comparisons at similar FLOPs, we set width pruning ratios to 50\%. Magnitude pruning shows relatively better performance among structural pruning techniques, yet DyDiT consistently outperforms it by a substantial margin. With a 20\% merging ratio, ToMe speeds up generation but sacrifices performance. As mentioned, the lack of convolutional layers and skip connections makes ToMe suboptimal on DiT.

\begin{figure}[t]
    \centering
    \includegraphics[width=0.9\textwidth]{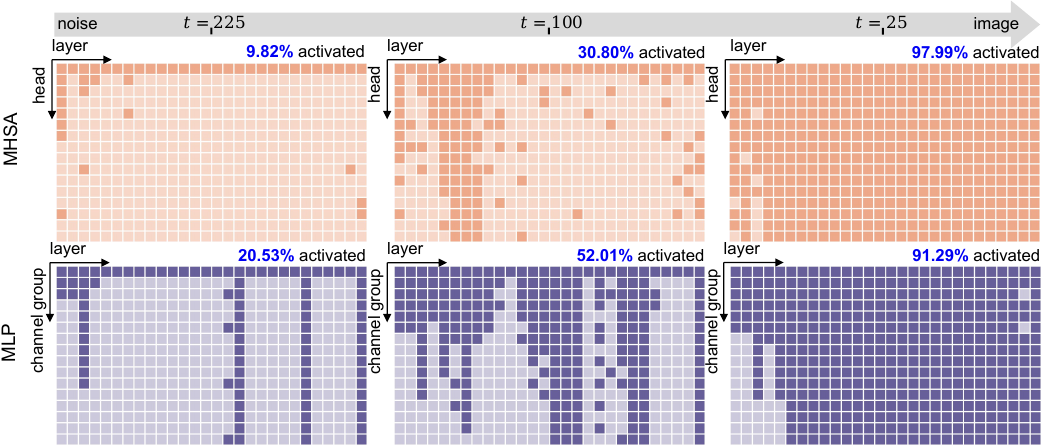}
\caption{\textbf{Visualization of dynamic architecture} in a 250-step DDPM generation. \raisebox{0.25em}{\colorbox{attn_code0!100}{}} and \raisebox{0.25em}{\colorbox{attn_code1!100}{}} indicate the deactivated and activated heads in an MHSA block, while \raisebox{0.25em}{\colorbox{mlp_code0!100}{}} and \raisebox{0.25em}{\colorbox{mlp_code1!100}{}} denote the channel group deactivated or activated in an MLP block.} 
\label{fig:activated_architecture}
\end{figure} 

\begin{figure}[t]
    \centering
    \includegraphics[width=0.9\textwidth]{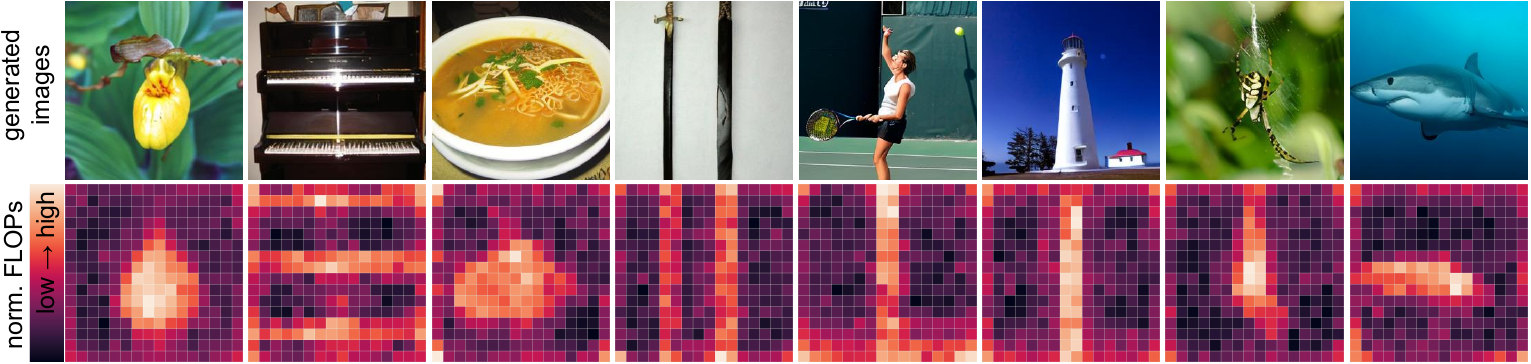}
\caption{\textbf{Computational cost normalized within $[0,1]$ across different image patches.}} 
\label{fig:token_flops}
\end{figure} 

\begin{table*}[t]
\centering
\scriptsize
\caption{\textbf{DyDiT combined with efficient samplers}~\citep{song2020denoising, lu2022dpm}.}
\tablestyle{9.3pt}{1.1}  
\begin{tabular}{c c c  c c c c  c c}
\multirow{2}{*}{Model}    &  \multicolumn{2}{c}{250-DDPM} & \multicolumn{2}{c}{50-DDIM}   & \multicolumn{2}{c}{20-DPM-solver++}  & \multicolumn{2}{c}{10-DPM-solver++} \\

  & \multicolumn{1}{c}{s/image $\downarrow$} & \multicolumn{1}{c}{FID $\downarrow$} & \multicolumn{1}{c}{s/image $\downarrow$} & \multicolumn{1}{c}{FID $\downarrow$} & \multicolumn{1}{c}{s/image $\downarrow$} & \multicolumn{1}{c}{FID $\downarrow$} & \multicolumn{1}{c}{s/image $\downarrow$} & \multicolumn{1}{c}{FID $\downarrow$}  \\
  \midrule[1.2pt]  

  DiT-XL & 10.22 & 2.27 & 2.00 & 2.26 & 0.84 & 4.62   & 0.42 & 11.66 \\ \hshline
 \rowcolor{gray!15} DyDiT-XL$_{\lambda=0.7}$ & 7.76$_{\textcolor{blue}{\downarrow1.32\times}}$ & 2.12 & 1.56$_{\textcolor{blue}{\downarrow1.28\times}}$ & \textbf{2.16} & 0.62$_{\textcolor{blue}{\downarrow1.35\times}}$  & 4.28 & 0.31$_{\textcolor{blue}{\downarrow1.35\times}}$ & \textbf{11.10}   \\
 \rowcolor{gray!15} DyDiT-XL$_{\lambda=0.5}$  & \textbf{5.91}$_{\textcolor{blue}{\downarrow1.73\times}}$ & \textbf{2.07} & \textbf{1.17}$_{\textcolor{blue}{\downarrow1.71\times}}$ & 2.36   & \textbf{0.46}$_{\textcolor{blue}{\downarrow1.83\times}}$ & \textbf{4.22} & \textbf{0.23}$_{\textcolor{blue}{\downarrow1.83\times}}$ & 11.31 \\  

    \end{tabular}
    \label{tab:sampler}
\end{table*}

\begin{table}[t]
\centering
\scriptsize
\tablestyle{12pt}{1.2}
\caption{\textbf{DyDiT combined with DeepCache}\citep{ma2023deepcache}. ``interval'' denotes the interval of cached timestep.}
  \begin{tabular}{c c  c c}
\multirow{1}{*}{Model}    &  \multicolumn{1}{c}{interval} & \multicolumn{1}{c}{s/image $\downarrow$} & \multicolumn{1}{c}{FID $\downarrow$} \\
  \midrule[1.2pt]  

  DiT-XL & 0 & 10.22 & 2.27 \\
  \rowcolor{gray!15} DyDiT-XL$_{\lambda=0.5}$  & 0 & \textbf{5.91}$_{\textcolor{blue}{\downarrow1.73\times}}$ & \textbf{2.07}  \\ \hshline
  DiT-XL & 2 & 5.02 & 2.47  \\ 
  \rowcolor{gray!15} DyDiT-XL$_{\lambda=0.5}$  & 2 & \textbf{2.99}$_{\textcolor{blue}{\downarrow1.68\times}}$ & \textbf{2.43}   \\\hshline
  DiT-XL & 5 & 2.03 & 6.73   \\ 
 \rowcolor{gray!15} DyDiT-XL$_{\lambda=0.5}$  & 3 & \textbf{2.01} & \textbf{3.37}$_{\textcolor{blue}{\downarrow3.36}}$   \\
    \end{tabular}
    \label{tab:cache}
\end{table}

\vspace{-1mm}
\paragraph{Cross-domain transfer learning.}
Transferring to downstream datasets is a common practice to leverage pre-trained generation models. In this experiment, we fine-tune a model pre-trained on ImageNet to perform cross-domain adaptation on the target dataset while concurrently learning the dynamic architecture, yielding DyDiT-S$_{\lambda=0.5}  \dag $  in Table~\ref{tab:fine_grained}.  We can observe that learning the dynamic architecture during the cross-domain transfer learning does not hurt the performance, and even leads to slightly better average FID score than DyDiT-S$_{\lambda=0.5}$. This further broadens the application scope of our method. More details, including the qualitative visualization are presented in the Supplementary Material.

\subsection{Ablation Study}
\vspace{-1mm}
\paragraph{Main components.}
We first conduct experiments to verify the effectiveness of each component in our method. We summarize the results in Table~\ref{tab:ablation_study}. ``I''  and ``II'' denote DiT with only the proposed timstep-wise dynamic width (TDW) and spatial-wise dynamic token (SDT), respectively. We can find that ``I'' performs much better than ``II''. This is attributed to the fact that, with the target FLOPs ratio $\lambda$ set to 0.5, most tokens in ``II'' have to bypass MLP blocks, leaving only MHSA blocks to process tokens, significantly affecting performance \citep{dong2021attention}. ``III'' represents the default model that combines both TDW and SDT, achieving obviously better performance than ``I''  and ``II''. Given a computational budget, the combination of TDW and SDT allows the model to discover computational redundancy from both the time-step and spatial perspectives. %

\vspace{-1mm}
\paragraph{Routers in temporal-wise dynamic width (TDW)} adaptively adjust each block's width for each timestep. ``I (random)'' replaces the learnable router with a random selection, leading to performance collapse. This is due to the random activation of heads/channels, which hinders the model's ability to generate high-quality images. We also implement a manually-designed strategy reducing $\sim$50\% FLOPs, termed  ``I (manual)'', in which we activate $5/6$, $1/2$, $1/3$, $1/3$ of the heads/channels for the intervals [0, $1/T$], [$1/T$, $2/T$], [$2/T$, $3/4T$], and [$3/4T$, $T$] timesteps, respectively. This strategy aligns the observation in Figure~\ref{fig:figure1(1)}(a) and allocates more computation to timesteps approaching 0. Therefore, ``I (manual)'' outperforms ``I (random)'' obviously.  However, it still underperforms ``I'', highlighting the importance of learned routers.

\vspace{-1mm}
\paragraph{Importance of token-level bypassing in SDT.}
We also explore an alternative design to token skipping. Specifically, each MLP block adopts a router to determine whether \emph{all tokens of an image} should bypass the block. This modification causes SDT to become a \emph{layer-skipping} approach~\citep{wang2018skipnet}. We replace SDT in ``III'' with this design, resulting in ``III (layer-skip)'' in Table~\ref{tab:ablation_study}. As outlined in Section~\ref{sec:introduction}, varying regions of an image face distinct challenges in noise prediction. A uniform token processing strategy fails to address this heterogeneity effectively. For example, tokens from complex regions might bypass essential blocks, resulting in suboptimal noise prediction. The results presented in Table~\ref{tab:ablation_study} further confirm the effectiveness of token skipping in our SDT.

\begin{table*}[t]
\centering
\scriptsize
\caption{\textbf{Experiment on SiT~\citep{ma2024sit} (DySiT).} In accordance with the original paper, we perform sampling using both the ODE (second-order Heun integrator) and the SDE (first-order Euler-Maruyama integrator).}
\tablestyle{8.5pt}{1.1} 
 \begin{tabular}
{c  c  c c c c c  c }
    Model & Params. (M) $\downarrow$   & FLOPs (G) $\downarrow$  & FID $\downarrow$ & sFID $\downarrow$ & IS $\uparrow$ & Precision $\uparrow$ & Recall $\uparrow$ \\
  \midrule[1.2pt]
    SiT-XL (ODE) & 675 &  118 & 2.11 & 4.62 & 255.87 & 0.80 & 0.61 \\
   \rowcolor{gray!15} DySiT-XL$_{\lambda=0.7}$ (ODE) & 678  & 85.10$_{\textcolor{blue}{\downarrow1.39\times}}$ & \textbf{1.95}  & \textbf{4.59} & \textbf{268.61}  & \textbf{0.82} & \textbf{0.61}    \\
   \rowcolor{gray!15} DySiT-XL$_{\lambda=0.5}$ (ODE) & 678 & \textbf{58.38}$_{\textcolor{blue}{\downarrow2.02\times}}$ & 2.11 & 4.75 & 268.19 & 0.82 & 0.59  \\ 

    SiT-XL (SDE) & 675 & 118  & \textbf{2.04} & \textbf{4.50} & 269.55 & 0.82 & 0.59 \\
   \rowcolor{gray!15} DySiT-XL$_{\lambda=0.7}$ (SDE) & 678   & 85.31$_{\textcolor{blue}{\downarrow1.38\times}}$ & 2.08  & 4.55 & 281.84   & 0.83 & \textbf{0.59}   \\
   \rowcolor{gray!15} DySiT-XL$_{\lambda=0.5}$ (SDE) & 678  & \textbf{58.27}$_{\textcolor{blue}{\downarrow2.03\times}}$ &  2.27 & 4.68 & \textbf{284.17} & \textbf{0.84} & 0.58  \\
\end{tabular}
\label{tab:sit}
\end{table*}

\begin{figure*}[htbp]
    \centering
    \hspace{-2mm}
    \includegraphics[width=0.25\textwidth]{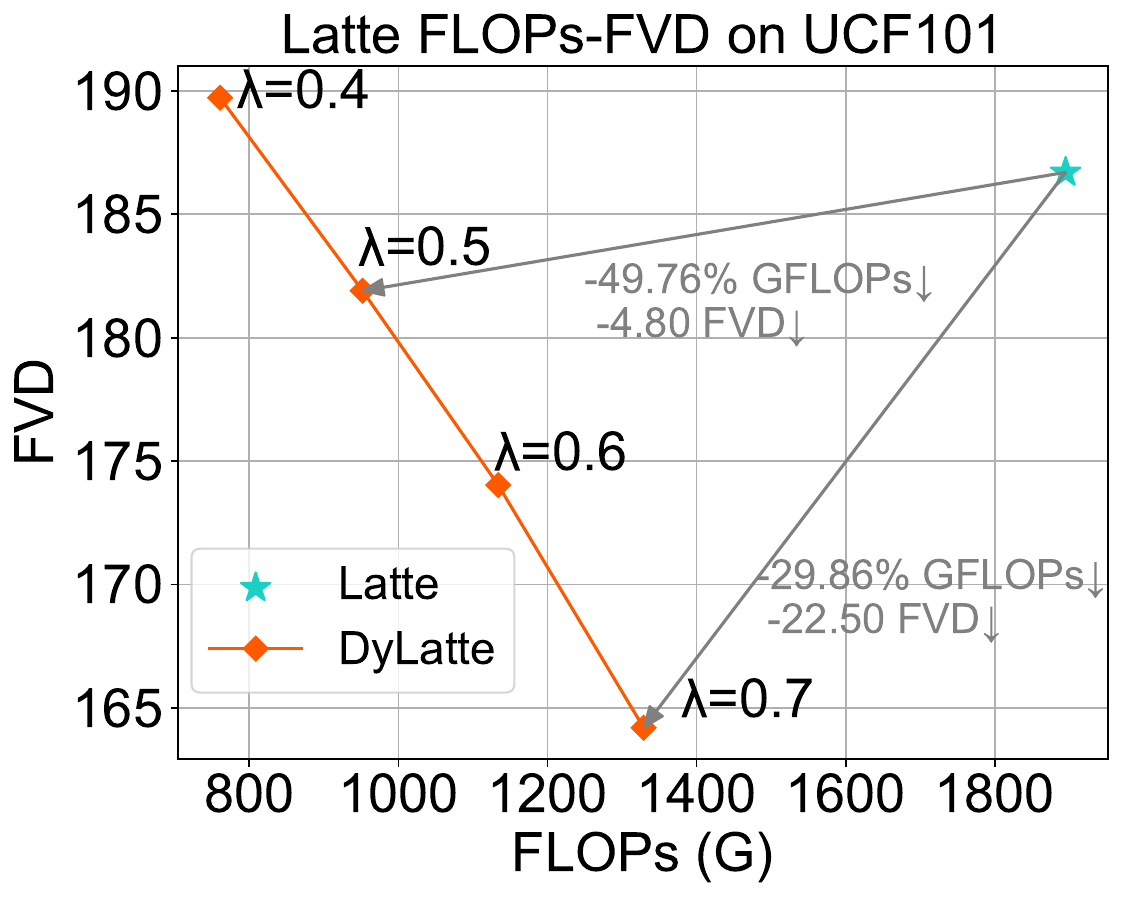}
    \hspace{-1mm}
    \includegraphics[width=0.235\textwidth]{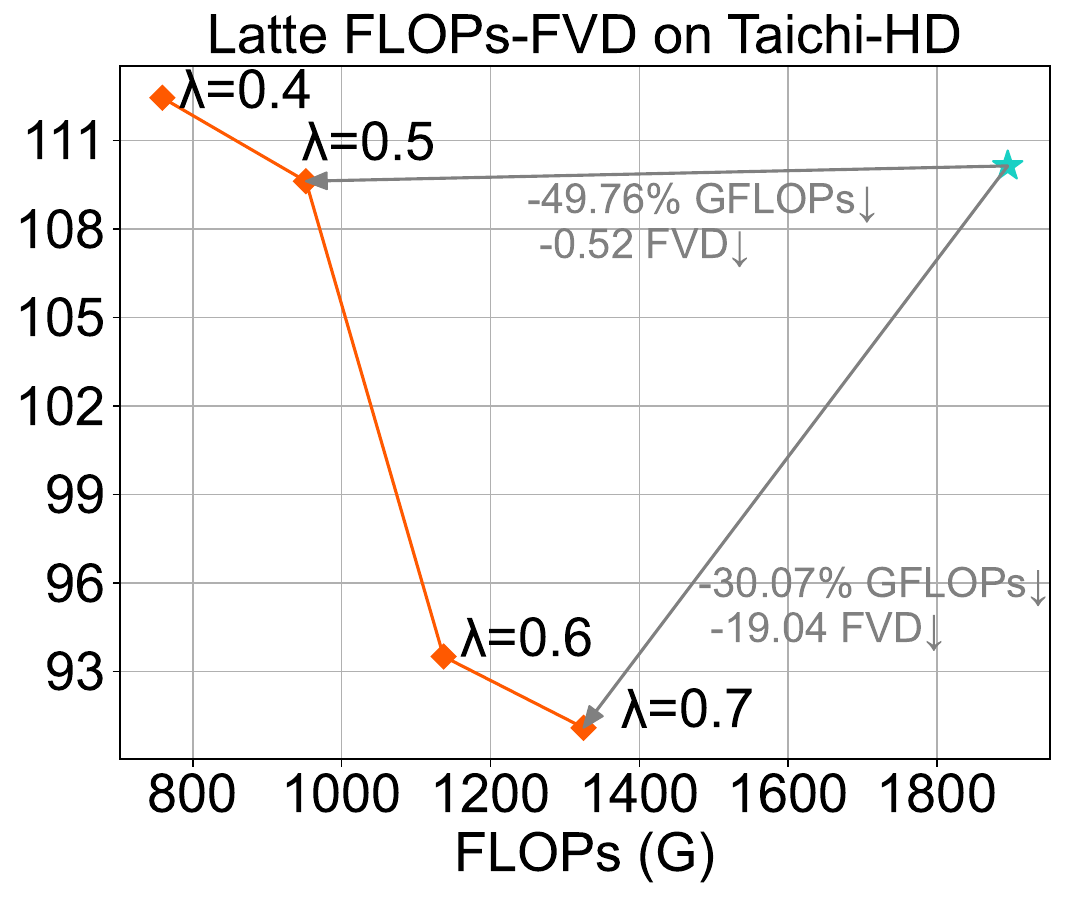}
    \hspace{-1mm}
    \includegraphics[width=0.23\textwidth]{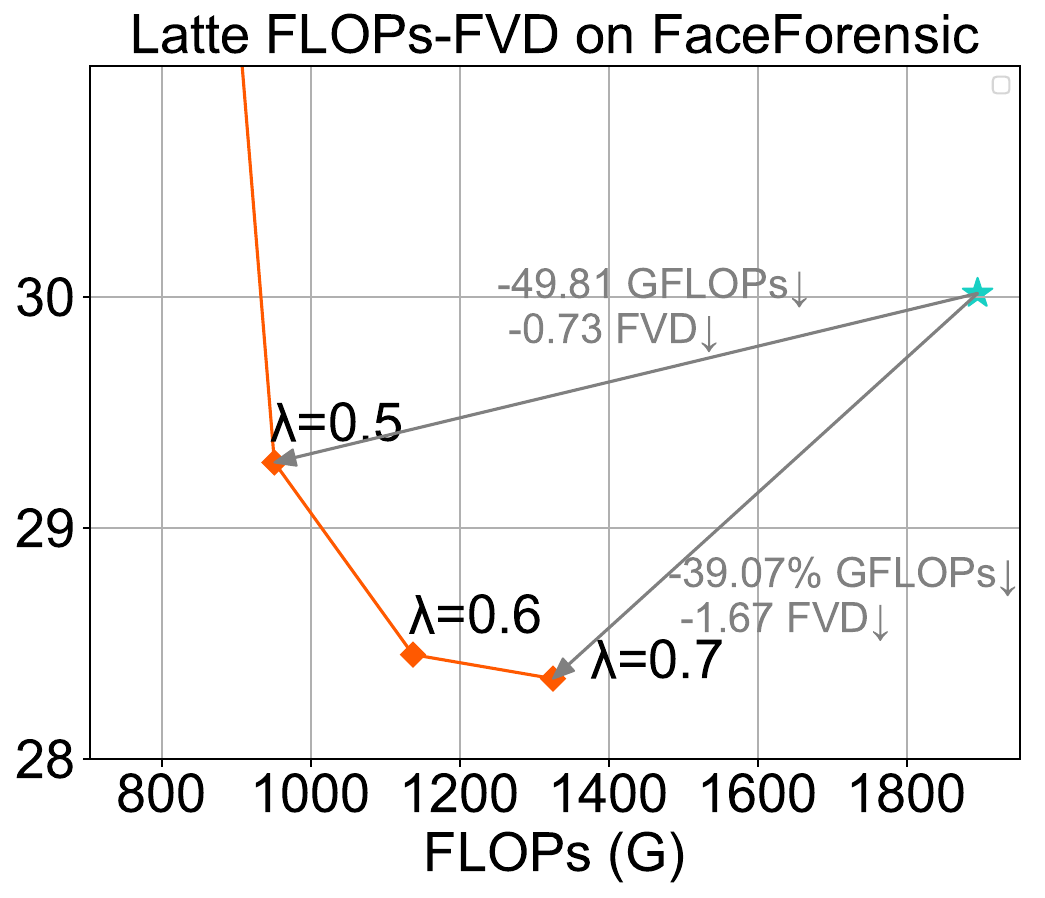}
    \hspace{-1mm}
    \includegraphics[width=0.23\textwidth]{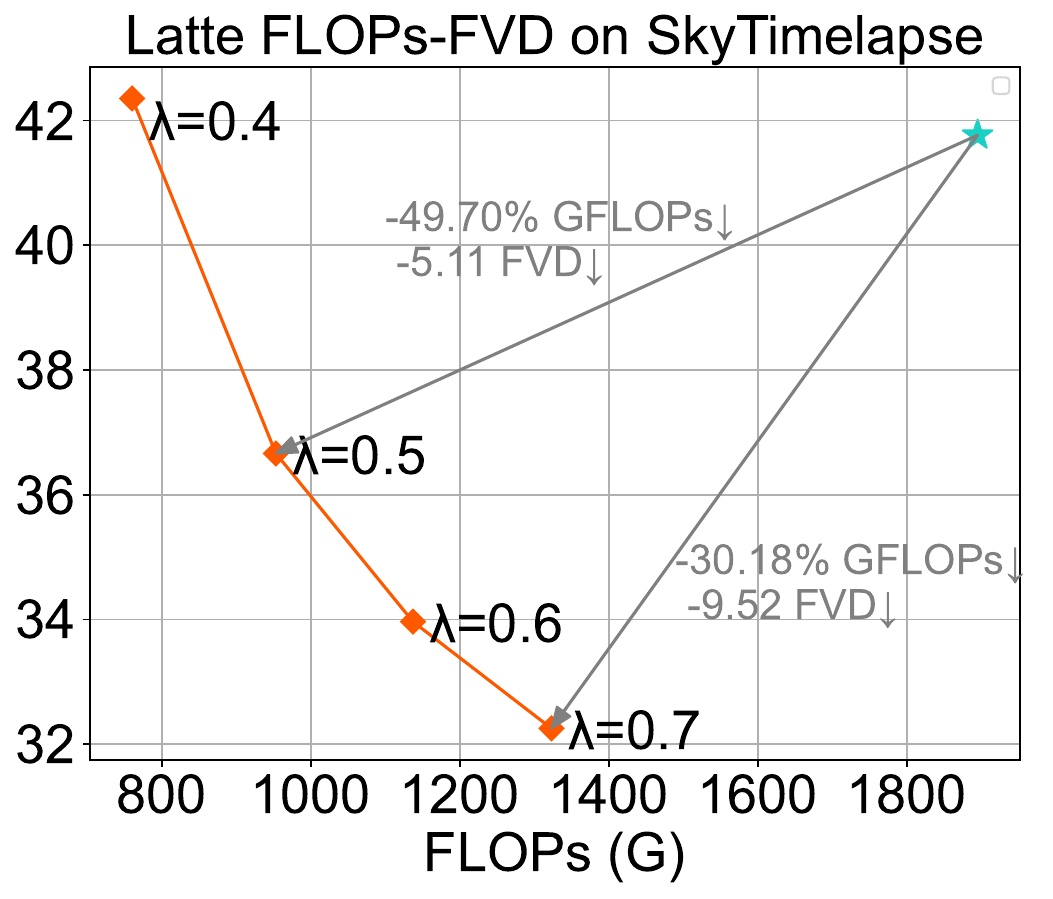}
    \caption{\textbf{FLOPs-FVD trade-off across four video datasets.} All models are of ``XL'' size.}
    \label{fig:flops_fvd_latte}
\end{figure*}

\vspace{-3mm}
\subsection{Visualization}
\vspace{-1mm}
\paragraph{Visual Comparison between DiT and DyDiT.}
In Figure~\ref{fig:imagenet_vis_compare}, we compare the visual quality of images generated by DiT and our approach. It can be observed that DyDiT, with $\lambda=0.5$, achieves similar perceptual fidelity while reducing computational costs by more than 50\%. These results demonstrate that our method effectively maintains both metric scores and generation quality simultaneously. Additional visualizations are provided in the Supplementary Material.

\vspace{-1mm}
\paragraph{Learned timestep-wise dynamic strategy.}
Figure~\ref{fig:activated_architecture} illustrates the activation patterns of heads and channel groups during 250-step DDPM generation. In this process, TDW progressively activates more MHSA heads and MLP channel groups as it transitions from noise to image. As discussed in Section~\ref{sec:introduction}, prediction is more straightforward when generation is closer to noise (larger $t$) and becomes increasingly challenging as it approaches the image (smaller $t$). Our visualization corroborates this observation, demonstrating that the model allocates more computational resources to more complex timesteps. Notably, the activation rate of MLP blocks surpasses that of MHSA blocks at $t\!=\!255$ and $t\!=\!100$. This can be attributed to the token bypass operation in the spatial-wise dynamic token (SDT), which reduces the computational load of MLP blocks, enabling TDW to activate additional channel groups with minimal computational overhead. %

\vspace{-1mm}
\paragraph{Learned spatial-wise computation allocation.} We quantify the computational cost on different image patches during generation, normalize within $[0, 1]$ in Figure~\ref{fig:token_flops}. These results verify that our SDT effectively learns to adjust computational expenditure based on the complexity of image patches. SDT prioritizes challenging patches containing detailed and colorful main objects. Conversely, it allocates less computation to background regions characterized by uniform and continuous colors. This behavior aligns with our findings in Figure~\ref{fig:figure1(1)}(b).

\begin{figure*}[t]
    \centering
    \includegraphics[width=0.95\textwidth]{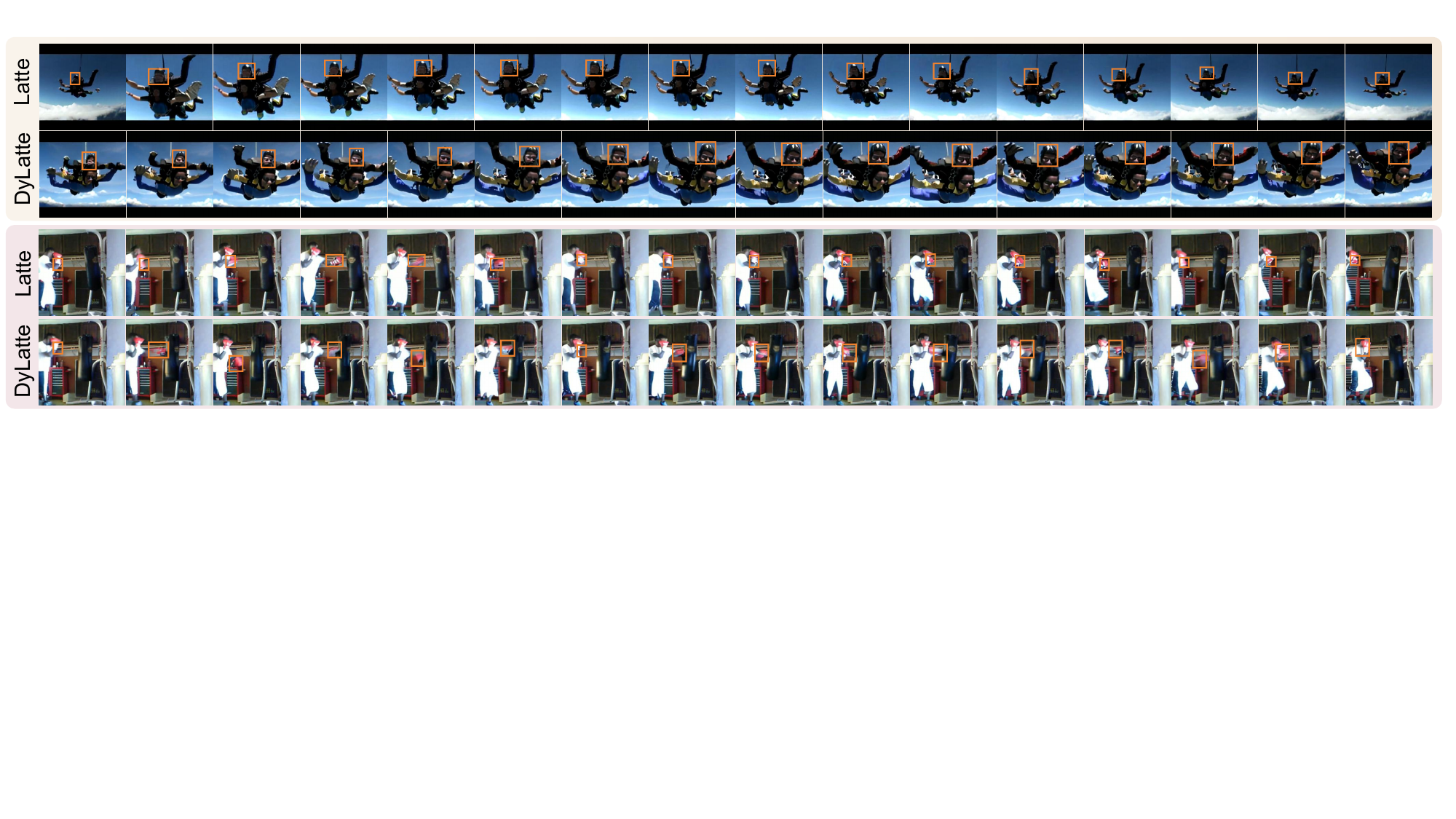}
\caption{\textbf{Qualitative comparison between Latte and DyLatte ($\lambda$=0.5) on UCF101.} We use orange boxes to highlight the key moving elements across frames. Our method achieves competitive video quality with significantly lower computational costs.}
\label{fig:video_visuzalization}  
\end{figure*}

\begin{table}[t]
\centering
\scriptsize
\tablestyle{12pt}{1.1} 
\caption{\textbf{Inference speed of Latte and DyLatte.} }
  \begin{tabular}{c c  c c}
\multirow{1}{*}{Model}    &  \multicolumn{1}{c}{FLOPs (G) $\downarrow$} & \multicolumn{1}{c}{s/video $\downarrow$} & \multicolumn{1}{c}{FVD $\downarrow$} \\
  \midrule[1.2pt]  

  Latte & 1895 & 157 & 186.70 \\
\rowcolor{gray!15} DyLatte $_{\lambda=0.4}$   & \textbf{761}$_{\textcolor{blue}{\downarrow2.49\times}}$ & \textbf{77}$_{\textcolor{blue}{\downarrow2.04\times}}$ & 189.72   \\
 \rowcolor{gray!15} DyLatte $_{\lambda=0.5}$  & 952$_{\textcolor{blue}{\downarrow1.99\times}}$ & 97$_{\textcolor{blue}{\downarrow1.62\times}}$ & 181.90    \\
 \rowcolor{gray!15} DyLatte $_{\lambda=0.6}$  & 1134$_{\textcolor{blue}{\downarrow1.67\times}}$ & 115$_{\textcolor{blue}{\downarrow1.37\times}}$ & 174.03   \\
 \rowcolor{gray!15} DyLatte $_{\lambda=0.7}$  & 1329$_{\textcolor{blue}{\downarrow1.43\times}}$ & 127$_{\textcolor{blue}{\downarrow1.24\times}}$ & \textbf{164.20}  \\
    \end{tabular}
    \label{tab:video_speed}
\end{table}

\subsection{Compatibility with Other Efficient Diffusion Approaches}
\vspace{-1mm}
\paragraph{Combination with efficient samplers.} Our DyDiT is a general architecture which can be seamlessly incorporated with efficient samplers such as DDIM~\citep{song2020denoising} and DPM-solver++ \citep{lu2022dpm}. As presented in Table~\ref{tab:sampler}, when using the 50-step DDIM, both DiT-XL and DyDiT-XL exhibit significantly faster generation, while our method consistently achieves higher efficiency due to its dynamic computation paradigm. When we further reduce the sampling step to 20 and 10 with DPM-solver++, we observe an FID increase on all models, while our method still achieves competitive performance compared to the original DiT. These findings highlight the potential of integrating our approach with efficient samplers, suggesting a promising avenue for future research.

\begin{table*}[t]
\centering
\scriptsize
\caption{\textbf{Performance comparison on GenEval}~\citep{ghosh2023geneval}. The target flops $\lambda$ is set to 0.7 for our DyFLUX-Lite.}
\tablestyle{1.3pt}{1.05}
 \begin{tabular}
{c  c c c c c c c  c c c c}
    Model & Params. (B) $\downarrow$  & s/image & FLOPs (T) $\downarrow$  & Overall $\uparrow$   & Single Obj. & Two Obj. & Counting  & Colors   & Position   & Attr. binding \\
  \midrule[1.2pt]
    FLUX &  12 & 18.85 & 40.8 & 66.48 & 98.75 & 84.85 & \textbf{74.69} & 76.60 & 21.75 & 42.25 \\
    FLUX-Lite & \textbf{8} & 15.23 & 30.0 & 62.06 & 98.44 & 74.24 & 64.38 & 75.53 & 17.00 & 42.75 \\
   \rowcolor{gray!15} DyFLUX-Lite & \textbf{8} & \textbf{11.84}$_{\textcolor{blue}{\downarrow1.59\times}}$ & \textbf{21.2}$_{\textcolor{blue}{\downarrow1.92\times}}$ & \textbf{67.64} & \textbf{99.06} & \textbf{86.36} & 67.19 & \textbf{78.99} & \textbf{22.25} & \textbf{52.00} \\

\end{tabular}
\label{tab:flux_geneval}
\end{table*}

\begin{table*}[t]
\centering
\scriptsize
\caption{\textbf{Evaluation on DPGBench~\citep{hu2024ella} and GenAIBench~\citep{li2024genai}}. All other settings remain the same as in Table~\ref{tab:flux_geneval}. }
\tablestyle{4.0pt}{1.05}
 \begin{tabular}
{c  | c c c c c c | c c c c c c}

\multirow{2}{*}{Model}  & \multicolumn{6}{c|}{DPGBench} & \multicolumn{3}{c}{GenAIBench-VQAScore} & \multicolumn{3}{c}{GenAIBench-CLIPScore} \\
    {} & Global & Entity & Attribute & Relation & Other & \#Average & Basic & Advanced & All & Basic & Advanced & All \\
  \midrule[1.2pt]
FLUX & 85.14 & 85.15 & \textbf{87.09} &  86.30 &  83.00 &  79.39 & 0.85 & 0.64 & 0.73 & 0.28 & 0.26 & 0.27  \\
FLUX-Lite & 83.46 & 86.36 & 84.61  & 84.51 & \textbf{86.40}  & 78.83  & 0.83 & 0.63 & 0.72 & 0.27 & 0.26 & 0.27  \\
\rowcolor{gray!15} DyFLUX-Lite & \textbf{87.32} & \textbf{87.13} & 86.52 & \textbf{86.59}  & 81.56 & \textbf{80.34}   & \textbf{0.88} & \textbf{0.65} & \textbf{0.76} & \textbf{0.29}  & \textbf{0.27} & \textbf{0.28} \\

\end{tabular}
\label{tab:flux_challenging}
\end{table*}

\begin{figure*}[t]
    \centering
    \includegraphics[width=0.9\textwidth]{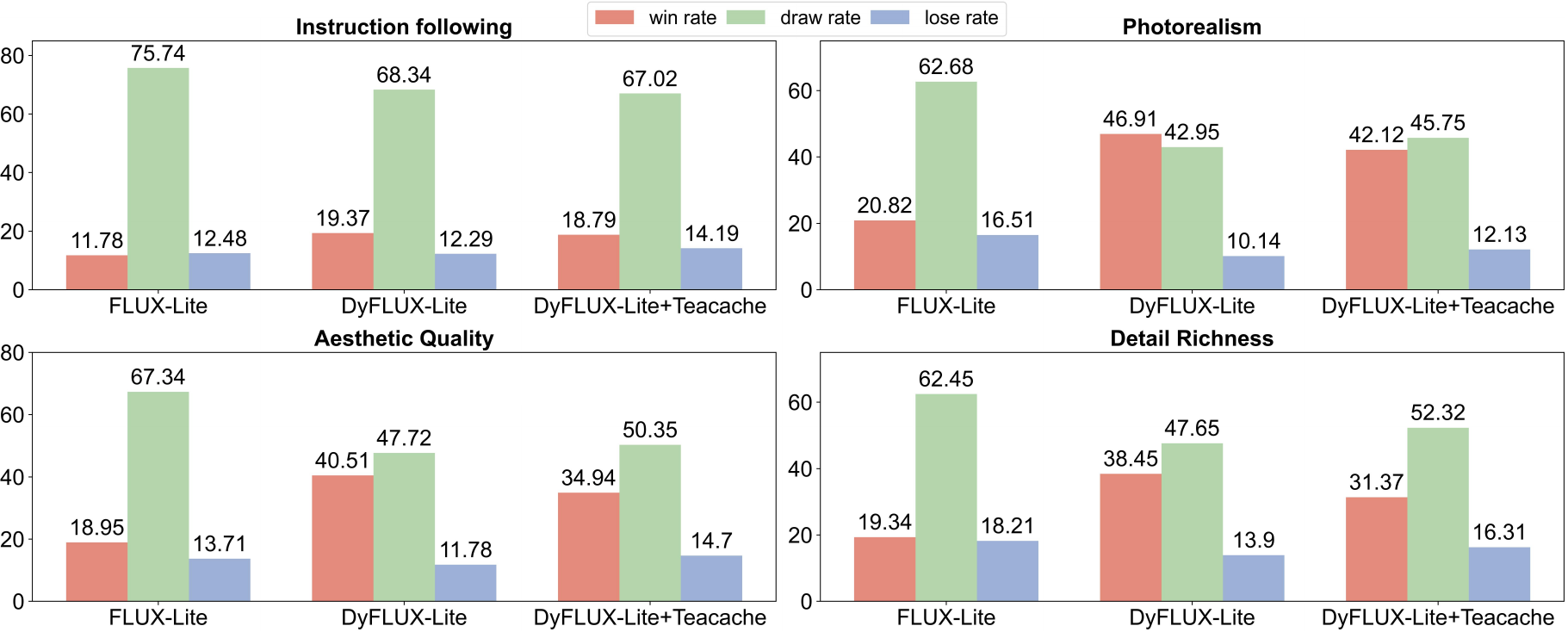}
\caption{\textbf{User study comparing  FLUX-Lite and DyFLUX-Lite with the original FLUX~\citep{flux1-lite}.} For DyFLUX-Lite, we further integrate Teacache~\citep{liu2024timestep}, a training-free global acceleration method, which yields an additional 1.47× speed improvement.
}
\label{fig:userstudy}  
\end{figure*}

\vspace{-1mm}
\paragraph{Combination with Caching.}
DeepCache \citep{ma2023deepcache} is a train-free technique which globally accelerates generation by caching feature maps at specific timesteps and reusing them in subsequent timesteps. As shown in Table~\ref{tab:cache}, with a cache interval of 2, DyDiT achieves further acceleration with only a marginal performance drop. In contrast, DiT with DeepCache requires a longer interval (\eg 5) to achieve comparable speed with ours, resulting in an inferior FID score. 
These results demonstrate the superior compatibility of DyDiT with DeepCache.

\section{DyDiT++ Experiments on Flow Matching}

\paragraph{Setup.} Since SiT~\citep{ma2024sit} shares the same architecture as DiT~\citep{peebles2023scalable}, we use the same hyperparameters to fine-tune SiT, enabling it to learn dynamic mechanisms and resulting in DySiT.

\vspace{-1mm}
\paragraph{Quantitative results.}
\noindent As discussed in Section~\ref{sec:flow}, the flow-based generation model SiT~\citep{ma2024sit} still has redundancy in both temporal and spatial dimensions. Therefore, we perform experiments on SiT to assess the generalization capabilities of the proposed method with flow-based approaches.
Since DiT and SiT share the same architecture, our dynamic architecture can be seamlessly integrated with SiT, resulting in DySiT. 
The results in Table~\ref{tab:sit} demonstrate that DySiT-XL consistently achieves performance comparable to the original SiT under both ODE (second-order Heun integrator) and SED (first-order Euler-Maruyama integrator) sampling methods. Moreover, DySiT-XL maintains comparable performance with $<$50\% computation. This finding directs a promising avenue for enhancing the inference efficiency of flow-based models.

\begin{figure}[t]
    \centering
    \includegraphics[width=0.9\textwidth]{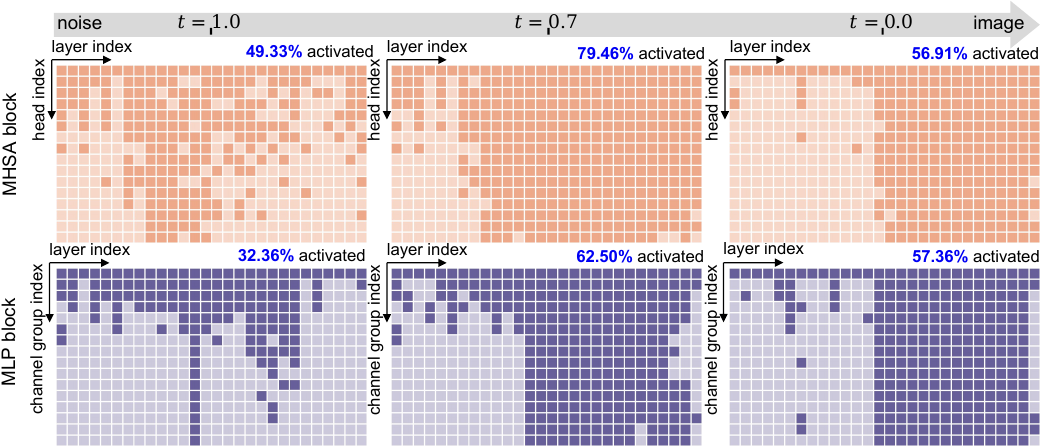}
\caption{\textbf{Visualization of dynamic architecture in DySiT}. \raisebox{0.25em}{\colorbox{attn_code0!100}{}} and \raisebox{0.25em}{\colorbox{attn_code1!100}{}} indicate the deactivated and activated heads in an MHSA block, while \raisebox{0.25em}{\colorbox{mlp_code0!100}{}} and \raisebox{0.25em}{\colorbox{mlp_code1!100}{}} denote the channel group deactivated or activated in an MLP block, respectively.} 
\label{fig:sit_activated_architecture}
\end{figure}

\vspace{-1mm}
\paragraph{Learned timestep-wise dynamic strategy in DySiT.}
From Figure~\ref{fig:sit_activated_architecture}, we observe that DySiT activates more components in the middle of the generation process (\eg, at $t=0.7$), while activating fewer components at the start ($t=1.0$) and end ($t=0.0$) of the generation, a behavior that differs from DyDiT as shown in Figure~\ref{fig:activated_architecture}. These results  align with the difficulty distribution shown in Figure~\ref{fig:figure1(1)} (c), which is relatively higher in the middle and lower at both the start and end.

\begin{figure}[t]
    \centering
    \includegraphics[width=0.9\textwidth]{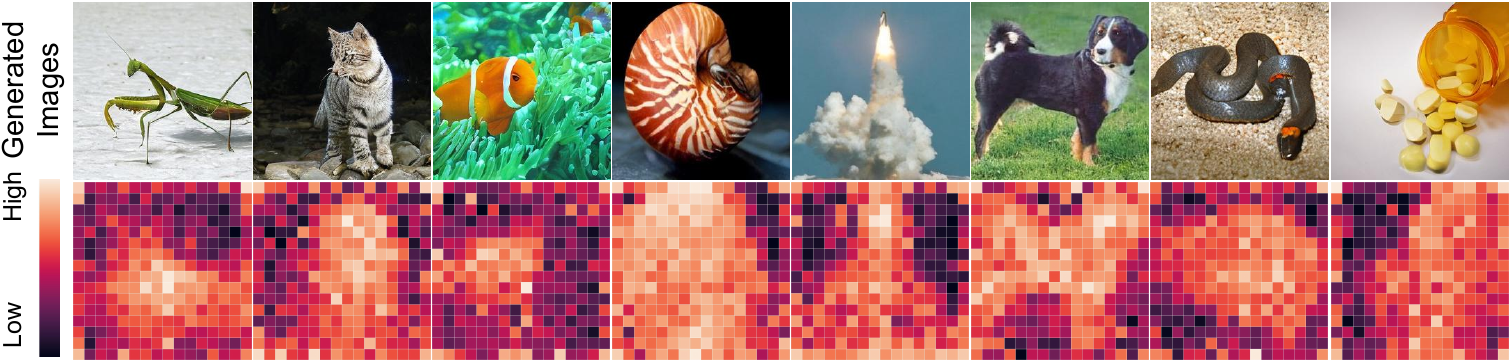}
\caption{\textbf{Computational cost normalized within $[0,1]$ across different image patches from DySiT.}} 
\label{fig:sit_token_flops}
\end{figure} 

\begin{figure*}[t]
    \centering
    \includegraphics[width=0.90\textwidth]{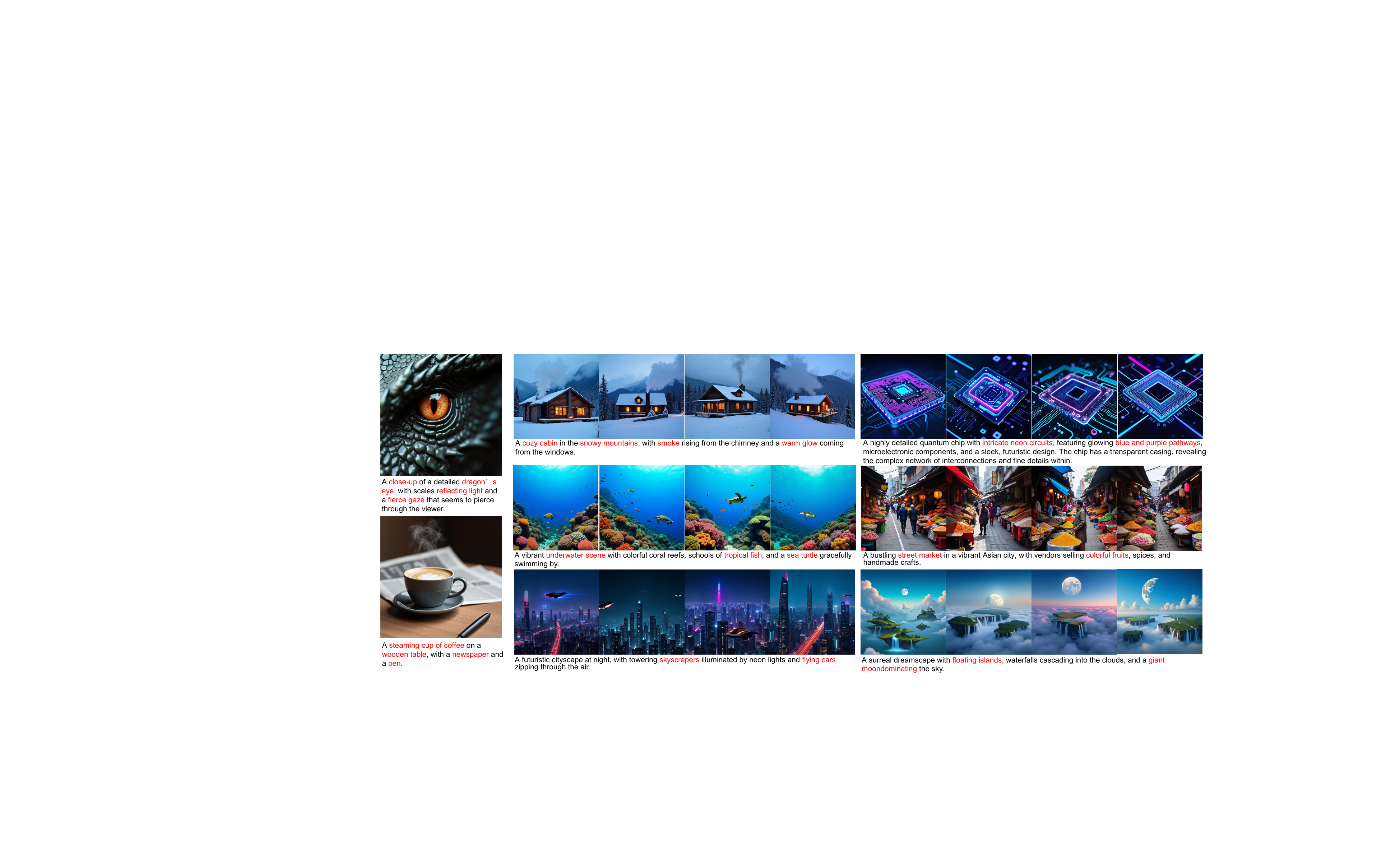}
\caption{\textbf{Images generated by our DyFLUX-Lite.}  The target flops $\lambda$ is set to 0.7. Please zoom in for a clear view.} 
\label{fig:flux_vis}  
\end{figure*}

\vspace{-1mm}
\paragraph{Learned spatial-wise computation allocation in DySiT.}
From Figure~\ref{fig:sit_token_flops}, we can observe that DySiT has learned to dynamically allocate computation across different image patches, similar to the behavior observed in DyDiT, and aligns with the loss pattern shown in Figure~\ref{fig:figure1(1)} (c).

\section{DyDiT++ Experiments on Video Generation}
\vspace{-1mm}
\paragraph{Setup.}
Following Latte~\citep{ma2024latte}, we conduct experiments on four video datasets: UCF101~\citep{soomro2012dataset}, Taichi-HD~\citep{siarohin2019first}, FaceForensics~\citep{rossler2018faceforensics}, and SkyTimelapse~\citep{xiong2018learning}, We generate 2,048 video clips, each consisting of 16 frames at a resolution of 256×256, and evaluate the 
generation quality using the Fréchet Video Distance (FVD)~\citep{unterthiner2018towards} metric. Unless otherwise specified, we experiment with the largest Latte-XL. We initialize DyLatte using the official pre-trained checkpoints and fine-tune it for 150,000 iterations to adapt to the dynamic architecture.

\vspace{-1mm}
\paragraph{Quantitative results.} We first illustrate the FLOPs-FVD curves across four datasets in Figure~\ref{fig:flops_fvd_latte}. Notably, DyLatte ($\lambda=0.5$) achieves comparable performance with the original Latte while requiring significantly fewer FLOPs. Furthermore, by increasing the FLOPs to 70\% ($\lambda\!=\!0.7$), the FVD score can be further improved. We further present the efficiency-performsance trade-off on UCF-101 in Table~\ref{tab:video_speed}.
The speed tests are performed on an NVIDIA V100 32G GPU. These results highlight that our method not only reduces computational cost, but also achieves tangible acceleration.
Furthermore, these results demonstrate the generalizability of the proposed dynamic architecture for video generation.

\vspace{-1mm}
\paragraph{Qualitative results.}
Figure~\ref{fig:video_visuzalization} visualizes the sampled videos by Latte and our DyLatte on UCF101~\citep{soomro2012dataset}. The results demonstrate that DyLatte generates videos with quality comparable to Latte while significantly reducing computational cost. 
This verifies that our method not only maintains the FVD score but also ensures high visualization quality.

\section{DyDiT++ Experiments on FLUX}

\vspace{-1mm}
\paragraph{Setup.}
We apply our proposed method to FLUX-Lite \citep{flux1-lite}, a lightweight version distilled from the original FLUX \citep{flux2024}. We set the target FLOPs ratio $\lambda$ to 0.7 by default and fine-tune the model for 500k iterations on our internal dataset (Details are provided in the Supplementary Material) to adapt the dynamic architecture. Performance evaluation includes both the GenEval \citep{ghosh2023geneval} benchmark and a comprehensive user study. Images are generated at a resolution of 1024$\times$1024.

\vspace{-1mm}
\paragraph{Quantitative results on benchmarks}
We first evaluate our method on GenEval~\citep{ghosh2023geneval}, with the results presented in Table~\ref{tab:flux_geneval}. DyFLUX-Lite achieves the highest overall score despite requiring less computational cost and offering faster generation speed than competing methods. Notably, DyFLUX-Lite scales effectively to approximately 8B parameters—11 times larger than DiT-XL—while maintaining superior performance, demonstrating both the scalability of our approach.

Additionally, we conduct experiments on two more challenging benchmarks, DPGBench~\citep{hu2024ella} and GenAIBench~\citep{li2024genai}, which feature prompts with more complex spatial and semantic relationships. The results show that our method effectively handles these scenarios while maintaining significantly lower computational costs compared to its counterparts, further validating the effectiveness of our approach.

\begin{figure}[t]
    \centering
    \includegraphics[width=0.9\textwidth]{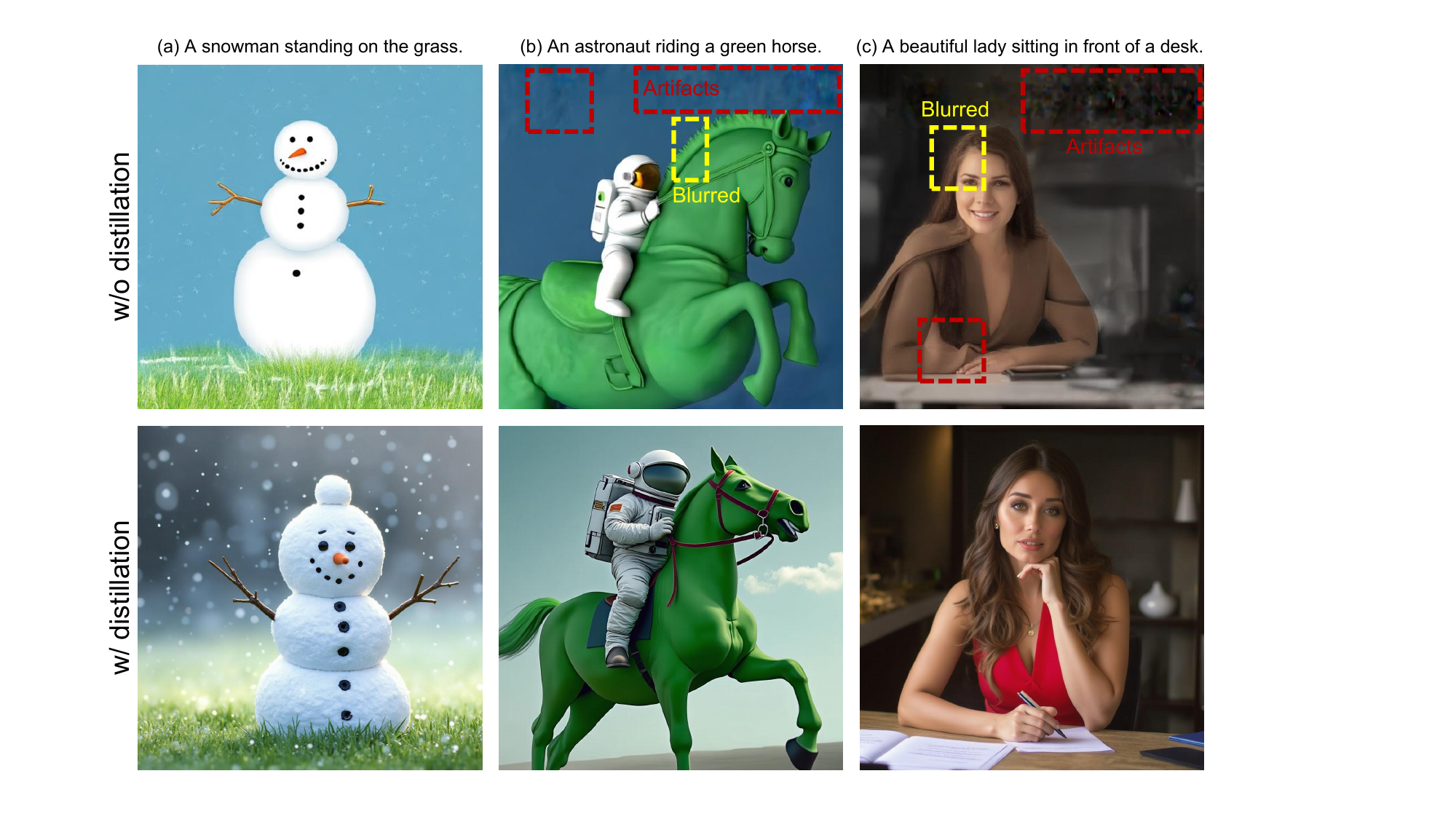}
\caption{\textbf{Necessity of distillation} when training DyFLUX.}
\label{fig:flux_distill_abl}  
\end{figure} 

\vspace{-1mm}
\paragraph{User study.} We conduct user studies to compare FLUX-Lite and DyFLUX-Lite with the original FLUX. Twelve participants are organized into six groups, with each group evaluating images generated from 3,108 prompt-image pairs. Participants rate each image on a scale from 1 to 5, and we report the win, draw, and loss rates relative to the original FLUX. The user study assesses the generation quality from instruction following, photorealism, aesthetic quality, and detail richness.

In Figure~\ref{fig:userstudy},  we observe that our method DyFLUX-Lite consistently achieves better win rates than FLUX-Lite across all metrics, demonstrating that even with lower computational costs, images generated by our method still satisfy human preferences. Furthermore, we integrate TeaCache~\citep{liu2024timestep}, a training-free global acceleration method, with DyFLUX-Lite, yielding an additional 1.47× speed improvement. Results show that DyFLUX-Lite+TeaCache maintains competitive performance, further verifying the compatibility of our approach.

\vspace{-1mm}
\paragraph{Visualization.} 
In Figure~\ref{fig:flux_vis}, we visualize some images generated by DyFLUX-Lite. These examples demonstrate our method's capability to produce visually rich details.

\vspace{-1mm}
\paragraph{Necessity of distillation.} As mentioned in Section~\ref{sec:dyflux}, we use a distillation technique to enhance DyFLUX training. Figure~\ref{fig:flux_distill_abl} is a visualized ablation study, illustrating that our distillation significantly improves the generation quality. For example, Figure~\ref{fig:flux_distill_abl}(a) shows that the image generated from the model trained with distillation is more photorealistic.  Figures~\ref{fig:flux_distill_abl}(b) and (c) demonstrate that distillation effectively reduces unwanted artifacts and mitigates blurred regions.  Its effectiveness for other dynamic models, such as DyDiT and DyLatte, is provided in the supplementary materials.

\begin{table}[t]
\centering
\normalsize
\tablestyle{10pt}{1.2}
    \caption{\textbf{Comparison with full fine-tuning and original LoRA.} L. Params. (M) refers to the number of learnable parameters. The target FLOPs ratio $\lambda$ is set to 0.5. }
    \begin{tabular}{c c  c c c}
    \multirow{1}{*}{Model}    &  \multicolumn{1}{c}{L. Params. (M) $\downarrow$} & \multicolumn{1}{c}{s/image $\downarrow$} & \multicolumn{1}{c}{FID $\downarrow$} \\
    \midrule[1.2pt]  
    DyDiT-XL &  678 & 5.91 & 2.07   \\
    \hshline
    DyDiT-XL$_{\text{LoRA}}$ & 9.29$_{\textcolor{blue}{\downarrow98.6\%}}$ & 5.91 & 2.41  \\
   \rowcolor{gray!15} DyDiT-XL$_{\text{PEFT}}$ & 9.94$_{\textcolor{blue}{\downarrow98.5\%}}$ & 5.96  & 2.23  \\
    \end{tabular}
    \label{tab:compare_LoRA}
\end{table}

\begin{table}[t]
\centering
\normalsize
\tablestyle{12pt}{1.2}
    \caption{\textbf{Different designs of TD-LoRA.} The ranks in each model are adjusted to ensure a comparable number of learnable parameters. The target FLOPs ratio $\lambda$ is set to 0.5.}
    \vskip -0.05in
    \begin{tabular}{c c  c c}
    \multirow{1}{*}{Design}    &  \multicolumn{1}{c}{L. Params. (M) $\downarrow$} & \multicolumn{1}{c}{FID $\downarrow$} \\
    \midrule[1.2pt]  
  \rowcolor{gray!15}  TD-LoRA & 9.94 &  \textbf{2.23}  \\
    Inverse TD-LoRA & \textbf{9.42} & 2.39   \\
    Symmetry TD-LoRA & 10.22 & 2.66   \\
    TD-LoRA w/o $\mathbf{E}_{t}$ & 9.91 & 3.16  \\
    \end{tabular}
    \label{tab:design_TD-LORA}
    \vspace{-4mm}
\end{table}

\begin{table}[t]
\centering
\normalsize
\tablestyle{12pt}{1.1}
    \caption{\textbf{Performance of TD-LORA with different ranks.} The rank $r=4$ yields a preferable performance.}
    \begin{tabular}{c c c  c c}
    \multirow{1}{*}{Model} & \multirow{1}{*}{Rank}    &  \multicolumn{1}{c}{L. Params. (M) $\downarrow$} & \multicolumn{1}{c}{FID $\downarrow$} \\
    \midrule[1.2pt]  
  DyDiT-XL  & - & 678 & 2.07  \\
  \hshline
  DyDiT-XL$_{\text{PEFT}}$  & 1 & 5.30 & 2.94  \\
  DyDiT-XL$_{\text{PEFT}}$  & 2 & 6.84 & 3.09  \\
 \rowcolor{gray!15} DyDiT-XL$_{\text{PEFT}}$  & 4 & 9.94 & 2.23  \\
  DyDiT-XL$_{\text{PEFT}}$  & 8 & 16.13 &  2.48  \\
    \end{tabular}
    \label{tab:lora_rank}
\end{table}

\begin{table*}[t] %
\centering
\setlength{\tabcolsep}{2.8pt}
\renewcommand{\arraystretch}{1.05}
\caption{\textbf{TD-LoRA on DySiT, DyLatte, and DyFLUX.} Evaluations are conducted on ImageNet, UCF101, and GenEval.}
\begin{minipage}[t]{0.32\textwidth} %
\centering
\scriptsize %
\begin{tabular}{c c c} %
Model & L. Params. (M) & FID $\downarrow$ \\ %
\midrule[1.2pt] %
DySiT-XL & 678 & 2.11 \\ 
DySiT-XL$_{\text{PEFT}}$ & 9.94$_{\textcolor{blue}{\downarrow98.5\%}}$ &  2.45 \\ %
\end{tabular}
\end{minipage}%
\begin{minipage}[t]{0.32\textwidth} %
\centering
\scriptsize %
\begin{tabular}{c c c} %
Model & L. Params. (M) & FVD $\downarrow$ \\ %
\midrule[1.2pt] %
DyLatte & 678  & 181.90 \\ 
DyLatte$_{\text{PEFT}}$ & 9.94$_{\textcolor{blue}{\downarrow98.5\%}}$  & 185.22  \\ %
\end{tabular}
\end{minipage}%
\begin{minipage}[t]{0.34\textwidth} %
\centering
\scriptsize %
\begin{tabular}{c c c} %
Model & L. Params. (B) & Overall  $\uparrow$ \\ %
\midrule[1.2pt] %
DyFLUX-Lite & 8 & 67.64 \\ 
DyFLUX-Lite$_{\text{PEFT}}$ & 2$_{\textcolor{blue}{\downarrow75.0\%}}$ & 58.99   \\ %
\end{tabular}
\end{minipage}
\label{tab:tdlora_other}
\end{table*}

\vspace{3mm}
\section{DyDiT++ Experiments under PEFT Setting} \label{sec:peft_exp}
\vspace{-2mm}

\paragraph{Setup.}
To train DyDiT in a parameter-efficient manner (denoted as DyDiT$_{\text{PEFT}}$), we adopt TD-LoRA proposed in Section~\ref{sec:dtlora} to fine-tune the parameters from both MHSA and MLP blocks, while employing LoRA~\citep{hu2021lora} to fine-tune the AdaLN blocks and fully fine-tuning the routers in both TDW and SDT.  In TD-LoRA, the expert number and the rank are set to 3 and 4 by default, whereas the standard LoRA rank is set to 8. We follow DyDiT for the other training strategies.  We adopt \colorbox{gray!15}{gray} to indicate this default setting.

\vspace{-1mm}
\paragraph{Comparison with full fine-tuning and original LoRA.}
In Table~\ref{tab:compare_LoRA}, we compare the proposed DyDiT-XL$_{\text{PEFT}}$, to the fully fine-tuned model (DyDiT-XL) and the original LoRA-based model (DyDiT-XL$_{\text{LoRA}}$). DyDiT-XL$_{\text{PEFT}}$ achieves an FID score comparable to the fully fine-tuned model while significantly reducing the number of learnable parameters.  Notably, fine-tuning with our method reduces memory consumption by 26\%, from 34GB to 25GB, compared to full fine-tuning, thereby lowering the  burden of adapting the dynamic architecture. Furthermore, compared to LoRA, our method achieves a superior FID score, highlighting the importance of dynamically adjusting parameters across different timesteps.

\vspace{-1mm}
\paragraph{Different designs of TD-LoRA.}
TD-LoRA, the key component in our DyDiT$_{\text{PEFT}}$, also has several alternative designs. Specifically, Inverse TD-LoRA replaces the matrix $\mathbf{A}$ with multiple experts from the original LoRA, while Symmetry TD-LoRA does the same for both $\mathbf{A}$ and $\mathbf{B}$. Additionally, we include TD-LoRA w/o $\mathbf{E}_{t}$ in the comparison. This variant removes the router that uses $\mathbf{E}_{t}$ as input in TD-LoRA and instead fuses the expert matrices through averaging, thereby maintaining the same parameters across all timesteps.

The comparison results are presented in Table~\ref{tab:design_TD-LORA}. The ranks in each model are adjusted to ensure a comparable number of learnable parameters. Compared to Inverse TD-LoRA and Symmetry TD-LoRA, our design performs the best, indicating that modifying the matrix $\mathbf{B}$ through weighted fusion across different timesteps is more effective than modifying $\mathbf{A}$ or both $\mathbf{A}$ and $\mathbf{B}$ simultaneously. Furthermore, the comparison between TD-LoRA and TD-LoRA without $\mathbf{E}_{t}$ shows that expert matrices should be adaptively fused across different timesteps rather than relying on simple averaging.

\vspace{-1mm}
\paragraph{Performance of TD-LORA with different ranks.}
In Table~\ref{tab:lora_rank}, we adjust the ranks in TD-LoRA to explore the trade-off between performance and the number of learnable parameters. The results demonstrate that with a rank of 4, introducing only 9.94M learnable parameters in DyDiT-XL$_{\text{PEFT}}$, our model achieves performance comparable to DyDiT-XL with full fine-tuning, resulting in a better trade-off between FID and learnable parameters. Notably, we are surprised to find that even with an extremely small number of parameters in TD-LoRA (\eg when the rank is set to 1), DyDiT-XL$_{\text{PEFT}}$ still achieves a competitive FID score, highlighting the superior parameter efficiency of our method.

\vspace{-1mm}
\paragraph{Performance of TD-LoRA on other models.} We further evaluate the effectiveness of TD-LoRA on other proposed models. In these experiments, DySiT and DyLatte adopt the same settings as DyDiT, with the number of experts and the rank set to 3 and 4, respectively. The target FLOPs ratio $\lambda$ is set to 0.5. For DyFLUX experiments, DyFLUX-Lite$_{\text{PEFT}}$ is initially trained at a resolution of 512$\times$512 and subsequently fine-tuned at a resolution of 1024$\times$1024 over just 10,000 iterations due to computational limitations, leaving it notably undertrained and with potential for further performance improvement through extended training. The target FLOPs ratio $\lambda$ is set to 0.7. From Table~\ref{tab:tdlora_other}, we observe that models trained with TD-LoRA significantly reduce the number of trainable parameters while maintaining competitive performance. These results demonstrate that the proposed TD-LoRA is highly effective for diverse models, further highlighting its adaptability.

\vspace{-2mm}
\section{Discussion and Conclusion}
\noindent
In this study, we investigate the training process of the Diffusion Transformer (DiT) and identify significant computational redundancy associated with specific diffusion \emph{timesteps} and \emph{spatial locations}. To this end, we propose Dynamic Diffusion Transformer (DyDiT), an architecture that can adaptively adjust the computation allocation across different timesteps and spatial regions. Building on this design, the enhanced version, DyDiT++, expands it to accelerate flow-matching-based generation and extends its applicability to more complex visual generation tasks, such as video generation and text-to-image generation. Additionally, it introduces Timestep-based Dynamic LoRA (TD-LoRA) to effectively reduce training costs. Comprehensive experiments on various models, including DiT, SiT, Latte, and FLUX, demonstrate the effectiveness of our method. We anticipate that the proposed method will advance the development of visual generation models. 

\vspace{-1mm}
\paragraph{Future Work.} This work primarily focuses on foundation models for visual generation, with potential extensions to various downstream applications, including image editing~\citep{huang2025diffusion, shuai2024survey} and controllable image generation~\citep{zhang2023adding, mou2024t2i}.

\vspace{-1mm}
\paragraph{Acknowledgment.}
This work was supported by Damo Academy through Damo Academy Research Intern
Program. Yang You’s research group is being
sponsored by NUS startup grant (Presidential Young Professorship), Singapore MOE Tier-1 grant, ByteDance
grant, ARCTIC grant, SMI grant (WBS number: A8001104-00-00), Alibaba grant, and Google grant for TPU
usage.

\vspace{-2mm}

\bibliographystyle{assets/plainnat}
\bibliography{paper}

\newpage
\beginappendix
\appendix

We organize our appendix as follows.

\paragraph{Experimental settings:}
\begin{itemize}
    \item Section~\ref{app_sec:our_imagenet}: Training details of DyDiT on ImageNet.
    \item Section~\ref{app_sec:model_details}: Model configurations of both DiT and DyDiT.
    \item Section~\ref{app_sec:prun_imagenet}: Implement details of pruning methods on ImageNet.
    \item Section~\ref{app_sec:fine_grained}: Details of in-domain fine-tuning on fine-grained datasets.
    \item Section~\ref{app_sec:exp_cross}: Details of cross-domain fine-tuning.
\end{itemize} %

\paragraph{Additional results:}
\begin{itemize}

    \item Section~\ref{supp:scratch}:  The results of training DyDiT from scratch.

    \item Section~\ref{app_sec:inference}: The inference speed of DyDiT and its acceleration over DiT across models of varying sizes and specified FLOP budgets.
    
    \item Section~\ref{app_sec:uvit}: The generalization capability of our method on the U-ViT~\citep{bao2023all} architecture.
    \item Section~\ref{app_sec:further_finetune}: Further fine-tuning the original DiT to show that the competitive performance of our method is not due to the additional fine-tuning.
    \item Section~\ref{app_sec:high_res}: The effectiveness of DyDiT on 512$\times$512 resolution image generation.
    \item Section~\ref{app_sec:t2i_gen}: The effectiveness of DyDiT in text-to-image generation, based on PixArt~\citep{chen2023pixart}. 
    \item Section~\ref{app_sec:lcm}: Integration of DyDiT with a representative distillation-based efficient sampler, the latent consistency model (LCM)~\citep{luo2023latent}.
    \item Section~\ref{app_sec:early}: Comparison between DyDiT with the early exiting diffusion model~\citep{moon2023early}.
    \item Section~\ref{app:training_efficiency}: Fine-tuning efficiency of DyDiT. We fine-tune our model by fewer iterations.
    \item Section~\ref{app:data_efficiency}: Data efficiency of DyDiT. Our model is fine-tuned on only 10\% of the training data.

    \item Section~\ref{app:latte_anay}: Dynamic mechanism learned in DyLatte.

    \item Section~\ref{app:dysis_ablation}: Ablation study on DySiT.

    \item Section~\ref{app:dysit_scratch}: The results of training SiT from scratch.

    \item Section~\ref{app:pred_mode}: Analysis of diffusion models with various prediction modes.

    \item Section~\ref{app:scala_scratch}: Scalability experiments of DyDiT in the training-from-scratch setting.

    \item Section~\ref{app_sec:distill}: Effectiveness of the distillation technique in dynamic models.

\end{itemize} %

\paragraph{Visualization:}
\begin{itemize}
    \item Section~\ref{supp:flux}: Qualitative visualization of images generated by DyFLUX.

    \item Section~\ref{supp:fine_grained}: Qualitative visualization of images generated by DyDiT-S on fine-grained datasets.
    
    \item Section~\ref{app_sec:loss_map}: Additional visualizations of loss maps of DiT-XL. 
    \item Section~\ref{app_sec:flops_map}: Additional visualizations of computational cost across different image patches. 
    \item Section~\ref{app:visualization}: Visualization of images generated by DyDiT-XL$_{\lambda=0.5}$ on the ImageNet dataset at at resolution of $256 \times 256$.
    \item Section~\ref{rebuttal:visualize_lambda}: Visualization of DyDiT with different $\lambda$s.

    \item Section~\ref{rebuttal:visualize_coco}: Visual comparison of images generated by PixArt~\citep{chen2023pixart} and the proposed DyPixArt on the COCO dataset.

\end{itemize} %

\paragraph{Others:} 
\begin{itemize}
    \item Section~\ref{supp:latte}: Detailed architecture  of Latte~\citep{ma2024latte} and the proposed DyLatte.
    \item Section~\ref{supp:internal_dataset}: Details of internal dataset used for DyFLUX.
    \item Section~\ref{supp:loss_disc}: Loss pattern discrepancy of DiT and SiT. 
    \item Section~\ref{rebuttal:freq_ques}: Frequently asked questions. 
\end{itemize} %

\addtocounter{table}{+1} %

\newpage

\section{Experimental Settings.} \label{app_details}
\subsection{Training details of DyDiT on ImageNet} \label{app_sec:our_imagenet}
In Table~\ref{app_tab:training_detail}, we present the training details of our model on ImageNet. For DiT-XL, which is pre-trained over 7,000,000 iterations, only 200,000 additional fine-tuning iterations (around 3\%) are needed to enable the dynamic architecture ($\lambda=0.5$) with our method. For a higher target FLOPs ratio $\lambda=0.7$, the iterations can be further reduced.

\begin{table}[!h]
    \centering
    \tablestyle{5.0pt}{1.1}
\begin{adjustbox}{width=1.0\textwidth}
    \begin{tabular}{lccc}
    
        model & DiT-S & DiT-B  & DiT-XL \\ 
        \midrule[1.2pt]

        optimizer & \multicolumn{3}{c}{AdamW~\citep{loshchilov2017decoupled}, learning rate=1e-4} \\
        global batch size & \multicolumn{3}{c}{256} \\
        target FLOPs ratio $\lambda$ & [0.9, 0.8, 0.7, 0.5, 0.4, 0.3] & [0.9, 0.8, 0.7, 0.5, 0.4, 0.3] & [0.7, 0.6, 0.5, 0.3] \\
        fine-tuning iterations & 50,000 & 100,000  &  150,000 for $\lambda=0.7$ 200,000 for others \\
        warmup iterations & 0 & 0 & 30,000 \\
   augmentation & \multicolumn{3}{c}{random flip} \\
   cropping size & \multicolumn{3}{c}{224$\times$224} \\
    \end{tabular}
      \end{adjustbox}
    \caption{Experimental settings of our adaption framework.}
    \label{app_tab:training_detail}
\end{table}

\subsection{Details of DiT and DyDiT models} \label{app_sec:model_details}
We present the configuration details of the DiT and DyDiT models in Table~\ref{tab:dit_model}. For DiT-XL, we use the checkpoint from the official DiT repository\footnote{\url{https://github.com/facebookresearch/DiT}}~\citep{pan2024t}. For DiT-S and DiT-B, we leverage pre-trained models from a third-party repository\footnote{\url{https://github.com/NVlabs/T-Stitch}} provided by \citet{pan2024t}.

\begin{table*}[!h]
\caption{\textbf{Details of DiT and DyDiT models.}  The router in DyDiT introduce a small number of parameters. $\dag$  denotes that the architecture is dynamically adjusted during generation.} \label{tab:dit_model}
\label{app:model_size}
\centering
\tablestyle{5.0pt}{1.1}
\begin{tabular}{c c  c c c c  c}

    \multirow{1}{*}{model}  & \multirow{1}{*}{params. (M) $\downarrow$}   &  \multicolumn{1}{c}{layers}  & \multicolumn{1}{c}{heads}   & \multicolumn{1}{c}{channel} & \multicolumn{1}{c}{pre-training}   & \multicolumn{1}{c}{source}  \\ \midrule[1.2pt]
    DiT-S  & 33 &  12 & 6 & 384 & 5M iter & \citep{pan2024t}  \\
    DiT-B  & 130 & 12 & 12 & 768 & 1.6M iter  & \citep{pan2024t}  \\
    DiT-XL & 675     & 28 & 16  & 1152 & 7M iter  & \citep{peebles2023scalable}  \\ \hshline
    DyDiT-S & 33  &  12 & 6 $\dag$  & 384 $\dag$  & - & - \\
    DyDiT-B & 131 &  12 & 12 $\dag$ & 768 $\dag$  & - & - \\  
    DyDiT-XL & 678   & 28   & 16 $\dag$ & 1152 $\dag$ & - & - \\     
    \end{tabular}    
\end{table*}

\subsection{Comparison with pruning methods on ImageNet.}  \label{app_sec:prun_imagenet}
We compare our method with structure pruning and token pruning methods on ImageNet dataset.

\begin{itemize}

    \item Random pruning, Magnitude Pruning \citep{he2017channel}, Taylor Pruning \citep{molchanov2016pruning}, and Diff Pruning \citep{fang2024structural}: We adopt the corresponding pruning strategy to rank the importance of heads in multi-head self-attention blocks and channels in MLP blocks. Then, we prune the least important 50\% of heads and channels. The pruned model is then fine-tuned for the same number of iterations as its DyDiT counterparts.

    \item ToMe~\citep{bolya2023token}: Originally designed to accelerate transformer blocks in the U-Net architecture, ToMe operates by merging tokens before the attention block and then unmerging them after the MLP blocks. We set the token merging ratio to 20\% in each block.

\end{itemize}

\subsection{In-domain fine-tuning on fine-trained datasets.}   \label{app_sec:fine_grained}
We first fine-tune a DiT-S model, which is initialized with parameters pre-trained on ImageNet, on a fine-grained dataset. Following the approach in \citep{xie2023difffit}, we set the training iteration to 24,000. Then, we further fine-tune the model on the same dataset by another 24,000 iterations to adapt the pruning or dynamic architecture  to improve the efficiency of the model on the same dataset. We also conduct the generation at a resolution off $224 \times 224$. We search optimal classifier-free guidance weights for these methods. %

\begin{itemize}
    
    \item Random pruning, Magnitude Pruning~\citep{he2017channel}, Taylor Pruning \citep{molchanov2016pruning}, and Diff Pruning~\citep{fang2024structural}:  For each method, we rank the importance of heads in multi-head self-attention blocks and channels in MLP blocks, pruning the least important 50\%.
    \item ToMe~\citep{bolya2023token}: Originally designed to accelerate transformer blocks in the U-Net architecture, ToMe operates by merging tokens before the attention block and then unmerging them after the MLP blocks. We set the token merging ratio to 20\% in each block.

\end{itemize}

\subsection{Cross-domain transfer learning} \label{app_sec:exp_cross}
In contrast to the aforementioned in-domain fine-tuning, which learns the dynamic strategy within the same dataset, this experiment employs cross-domain fine-tuning. We fine-tune a DiT-S model (pre-trained exclusively on ImageNet) to adapt to the target dataset while simultaneously learning the dynamic architecture. The model is fine-tuned over 48,000 iterations with a batch size of 256.

\section{Additional Results}

\subsection{Training DyDiT from scratch}~\label{supp:scratch}

In Table~\ref{app_tab:scratch}, we present the results of training the original DiT~\citep{peebles2023scalable} and our DyDiT from scratch on ImageNet~\citep{deng2009imagenet}. We strictly follow the training settings outlined in the original DiT paper. It can be observed that DyDiT-XL$_{\lambda=0.7}$ does not outperform the original DiT under the same number of training iterations. This can be attributed to that the Gumbel noise~\citep{jang2016categorical}, which is introduced during DyDiT training, slightly slows down the convergence speed. However, when we increase the training iterations to 11,000,000, our approach achieves comparable performance while requiring fewer FLOPs.

Additionally, we report the results of fine-tuning a pre-trained DiT to adapt it to the proposed dynamic architecture, referred to as DyDiT-XL$_{\lambda=0.7}$ (fine-tuning). This approach achieves a better FID score than DiT-XL with only 150,000 fine-tuning iterations. Since pre-trained model checkpoints are often publicly available, we recommend directly fine-tuning on these checkpoints instead of training from scratch, to avoid ``reinventing the wheel''.

\begin{table*}[!h]
\caption{\textbf{Results of training from scratch on ImageNet~\citep{deng2009imagenet}}. DyDiT-XL$_{\lambda=0.7}$ (fine-tuning) is obtained by fine-tuning from a pre-trained checkpoint of DiT.}  
\centering
\tablestyle{5.0pt}{1.1}
\begin{tabular}{c  cc  c c  c c}

    \multirow{1}{*}{model} & \multirow{1}{*}{training iterations} & \multirow{1}{*}{s/image $\downarrow$ }  & \multirow{1}{*}{FLOPs (G) $\downarrow$}  & \multicolumn{1}{c}{FID $\downarrow$}   & \multicolumn{1}{c}{FID $\Delta \downarrow$}  \\ 
          \midrule[1.2pt]
    DiT-XL  & 7,000,000 & 10.22 & 118.69  & 2.27   & +0.00 \\     \hshline
    DyDiT-XL$_{\lambda=0.7}$ & 7,000,000  & 7.75 & 84.31  & 2.37      & +0.10   \\
    DyDiT-XL$_{\lambda=0.7}$ & 11,000,000  & 7.76 & 84.32  & 2.25      & -0.02 \\ \hshline
    DyDiT-XL$_{\lambda=0.7}$ (fine-tuning) & 150,000  & 7.76 & 84.33  & 2.12     & -0.15 \\

    \end{tabular}
    \label{app_tab:scratch}
\end{table*}

\subsection{Inference acceleration} 
\label{app_sec:inference}
In Table~\ref{app_tab:inference}, we present the acceleration ratio of DyDiT  compared to the original DiT across different FLOPs targets $\lambda$.  The results demonstrate that our method effectively enhances batched inference speed, distinguishing our approach from traditional dynamic networks~\citep{herrmann2020channel,meng2022adavit,han2023latency}, which adapt inference graphs on a per-sample basis and struggle to improve practical efficiency in batched inference. %

\begin{table*}[!h]
\centering
\tablestyle{5.0pt}{1.1}
\caption{We conduct batched inference on an NVIDIA V100 32G GPU using the optimal batch size for each model. The actual FLOPs of DyDiT may fluctuate around the target FLOPs ratio.}
\begin{tabular}{c c c c  c c}
\multirow{1}{*}{model}  & \multirow{1}{*}{s/image $\downarrow$} & \multirow{1}{*}{acceleration $\uparrow$}  & \multirow{1}{*}{FLOPs (G) $\downarrow$}  & \multicolumn{1}{c}{FID $\downarrow$}   & \multicolumn{1}{c}{FID $\Delta \downarrow$}  \\ 
      \midrule[1.2pt]

    DiT-S   & 0.65 & 1.00 $\times$ & 6.07   & 21.46 & +0.00 \\
    DyDiT-S$_{\lambda=0.9}$ & 0.63 & 1.03 $\times$&  5.72 & 21.06 & -0.40\\
    DyDiT-S$_{\lambda=0.8}$ & 0.56 & 1.16 $\times$& 4.94 & 21.95 & +0.49 \\
    DyDiT-S$_{\lambda=0.7}$ & 0.51 & 1.27 $\times$& 4.34 & 23.01 & +1.55 \\
    DyDiT-S$_{\lambda=0.5}$ & 0.42 & 1.54 $\times$& 3.16 & 28.75 & +7.29\\
    DyDiT-S$_{\lambda=0.4}$ & 0.38 & 1.71 $\times$& 2.63 & 36.21 & +14.75 \\
    DyDiT-S$_{\lambda=0.3}$ & 0.32 & 2.03 $\times$& 1.96 & 59.28 & +37.83 \\
\hshline
    DiT-B   & 2.09 & 1.00 $\times$  &   23.02  & 9.07 & +0.00 \\
    DyDiT-B$_{\lambda=0.9}$ & 1.97 & 1.05 $\times$ & 21.28 & 8.78 & -0.29 \\
    DyDiT-B$_{\lambda=0.8}$ & 1.76 & 1.18 $\times$ & 18.53 & 8.79 & -0.28 \\
    DyDiT-B$_{\lambda=0.7}$ & 1.57 & 1.32 $\times$ & 16.28 & 9.40 & +0.33 \\
    DyDiT-B$_{\lambda=0.5}$ & 1.22 & 1.70 $\times$ & 11.90 & 12.92 & +3.85\\
    DyDiT-B$_{\lambda=0.4}$ & 1.06 & 1.95 $\times$ & 9.71 & 15.54 & +6.47 \\  
    DyDiT-B$_{\lambda=0.3}$ & 0.89 & 2.33 $\times$ & 7.51 & 23.34 & +14.27 \\  
\hshline
    DiT-XL   & 10.22  & 1.00 $\times$  &  118.69  & 2.27  & +0.00 \\
    DyDiT-XL$_{\lambda=0.9}$ &  9.64 & 1.06 $\times$ & 110.73 & 2.15 & -0.12  \\
    DyDiT-XL$_{\lambda=0.8}$ &  8.66 & 1.18 $\times$ & 96.04  & 2.13 & -0.14 \\
    DyDiT-XL$_{\lambda=0.7}$ & 7.76 & 1.32 $\times$ & 84.33 & 2.12 & -0.15 \\
    DyDiT-XL$_{\lambda=0.5}$ & 5.91 & 1.73 $\times$ & 57.88 & 2.07 & -0.20 \\
    DyDiT-XL$_{\lambda=0.3}$ & 4.26 & 2.40 $\times$ & 38.85 & 3.36 & +1.09 \\

    \end{tabular}
        \label{app_tab:inference}%
\end{table*}

\subsection{Effectiveness on U-ViT.} 
\label{app_sec:uvit}
We evaluate the architecture generalization capability of our method through experiments on U-ViT~\citep{bao2023all}, a transformer-based diffusion model with skip connections similar to U-Net~\citep{ronneberger2015u}. The results, shown in Table~\ref{app_tab:uvit}, indicate that configuring the target FLOPs ratio $\lambda$ to 0.4 and adapting U-ViT-S/2 to our dynamic architecture (denoted as DyUViT-S/2 $_{\lambda=0.4}$) reduces computational cost from 11.34 GFLOPs to 4.73 GFLOPs, while maintaining a comparable FID score. We also compare our method with the structure pruning method Diff Pruning~\citep{fang2024structural} and sparse pruning methods ASP~\citep{pool2021channel, mishra2021accelerating} and SparseDM~\citep{wang2024sparsedm}. The results verify the superiority of our dynamic architecture over static pruning.

In Table~\ref{app_tab:uvit_imagenet}, we apply our method to the largest model, U-ViT-H/2, and conduct experiments on ImageNet. The results demonstrate that our method effectively accelerates U-ViT-H/2 with only a marginal performance drop. These results verify the generalizability of our method in U-ViT.

\begin{table*}[!h]
\caption{\textbf{U-ViT~\citep{bao2023all} performs image generation on the CIFAR-10 dataset~\citep{krizhevsky2009learning}.} Aligning with its default configuration, we generate images using 1,000 diffusion steps with the Euler-Maruyama SDE sampler~\citep{song2020score}.}  
\centering
\tablestyle{5.0pt}{1.1}
\begin{tabular}{c c c c c c}
    \multirow{1}{*}{model} & \multirow{1}{*}{s/image $\downarrow$ } & \multirow{1}{*}{acceleration $\uparrow$} & \multirow{1}{*}{FLOPs (G) $\downarrow$}  & \multicolumn{1}{c|}{FID $\downarrow$}   & \multicolumn{1}{c}{FID $\Delta \downarrow$}  \\ \midrule[1.2pt]
    U-ViT-S/2  &   2.19 & 1.00 $\times$ & 11.34  & 3.12 & 0.00 \\
    DyU-ViT-S/2$_{\lambda=0.4}$ & 1.04 &  2.10 $\times$   & 4.73   & 3.18  &  +0.06 \\
    pruned w/ Diff &- &- & 5.32 & 12.63 & +9.51 \\
    pruned w/ ASP &- &- & 5.76 & 319.87 & +316.75 \\
    pruned w/ SparseDM & - &- & 5.67 & 4.23 & +1.11 \\
    \end{tabular}
    \label{app_tab:uvit}
\end{table*}

\begin{table*}[!h]
\caption{\textbf{U-ViT~\citep{bao2023all} performs image generation on the ImageNet~\citep{deng2009imagenet}.} Aligning with its default configuration, we generate images using 50-step DPM-solver++\citep{lu2022dpm}.}  
\centering
\tablestyle{5.0pt}{1.1}
\begin{tabular}{c c c c c c}
    \multirow{1}{*}{model} & \multirow{1}{*}{s/image $\downarrow$ } & \multirow{1}{*}{acceleration $\uparrow$} & \multirow{1}{*}{FLOPs (G) $\downarrow$}  & \multicolumn{1}{c|}{FID $\downarrow$}   & \multicolumn{1}{c}{FID $\Delta \downarrow$}  \\ \midrule[1.2pt]
    U-ViT-H/2  &  2.22 &  1.00 $\times$ &  113.00 & 2.29 & 0.00 \\
    DyU-ViT-H/2$_{\lambda=0.5}$ & 1.35 & 1.57 $\times$ & 67.09 & 2.42  & +0.13
    \end{tabular}
    \label{app_tab:uvit_imagenet}
\end{table*}

\subsection{Further fine-tune original DiT on ImageNet.} 
\label{app_sec:further_finetune}
Our method is not attributed to additional fine-tuning. In Table~\ref{app_tab:futher_finetune}, we fine-tune the original DiT for 150,000 and 350,000 iterations, observing a slight improvement in the FID score, which fluctuates around 2.16. ``DiT-XL$^\prime$'' denotes that we introduce the same routers in DiT-XL to maintain the same parameters as that of DyDiT. 
Under the same iterations, DyDiT achieves a better FID while significantly reducing FLOPs, verifying that the improvement is due to our design rather than extended training iterations.%

\begin{table*}[!h]
\centering
\tablestyle{5.0pt}{1.1}
\caption{\textbf{Further fine-tuneing original DiT on ImageNet.}}
\begin{tabular}{c c c c  c c}
\multirow{1}{*}{model}  & \multirow{1}{*}{pre-trained iterations} & \multirow{1}{*}{fine-tuning iterations}  & \multirow{1}{*}{FLOPs (G) $\downarrow$}  & \multicolumn{1}{c}{FID $\downarrow$}   & \multicolumn{1}{c}{FID $\Delta \downarrow$}  \\ 
      \midrule[1.2pt]
    DiT-XL & 7,000,000 & - &  118.69 & 2.27 & +0.00  \\
    DiT-XL & 7,000,000 & 150,000 (2.14\%) &  118.69 & 2.16  & -0.11 \\
    DiT-XL & 7,000,000 & 350,000 (5.00\%) &  118.69  & 2.15  & -0.12 \\
    DiT-XL$^\prime$ & 7,000,000 & 150,000 (2.14\%) &  118.69 & 2.15  & -0.12 \\
    DyDiT-XL$_{\lambda=0.7}$ &  7,000,000 &  150,000 (2.14\%) & 84.33 & 2.12 & -0.15 \\

    \end{tabular}
        \label{app_tab:futher_finetune}
\end{table*}

\subsection{Effectiveness in High-resolution Generation.} 
\label{app_sec:high_res}
We conduct experiments to generate images at a resolution of 512$\times$512 to validate the effectiveness of our method for high-resolution generation. We use the official checkpoint of DiT-XL 512$\times$512 as the baseline, which is trained on ImageNet~\citep{deng2009imagenet} for 3,000,000 iterations. We fine-tune it for 150,000 iterations to enable its dynamic architecture, denoted as DyDiT-XL 512$\times$512. The target FLOP ratio is set to 0.7. The experimental results, presented in Table~\ref{app_tab:high}, demonstrate that our method achieves a superior FID score compared to the original DiT-XL, while requiring fewer FLOPs.

\begin{table*}[!h]
\caption{\textbf{Image generation at 512$\times$512 resolution on ImageNet~\citep{deng2009imagenet}}. We sample 50,000 images and leverage FID to measure the generation quality. We adopt 100 and 250 DDPM steps to generate images. ``FLOPs (G)'' denotes the average FLOPs in one timestep.}  
\centering
\tablestyle{5.0pt}{1.1}
\begin{tabular}{c  cc  c c  c c}

    \multirow{1}{*}{model} & \multirow{1}{*}{DDPM steps} & \multirow{1}{*}{s/image $\downarrow$ } & \multirow{1}{*}{acceleration $\uparrow$} & \multirow{1}{*}{FLOPs (G) $\downarrow$}  & \multicolumn{1}{c}{FID $\downarrow$}   & \multicolumn{1}{c}{FID $\Delta \downarrow$}  \\ 
          \midrule[1.2pt]
    DiT-XL 512$\times$512 & 100 & 18.36  & 1.00$\times$ & 514.80   & 3.75 & 0.00 \\
    DyDiT-XL 512$\times$512 $_{\lambda=0.7}$ & 100 & 14.00 & 1.31$\times$ &  375.35     &  3.61 & -0.14  \\ \hshline

   DiT-XL 512$\times$512 & 250 & 45.90  & 1.00$\times$ & 514.80   & 3.04 & 0.00 \\
   DyDiT-XL 512$\times$512 $_{\lambda=0.7}$ & 250 & 35.01 & 1.31$\times$ &  375.05     &  2.88 & -0.16  \\ 
    \end{tabular}
    \label{app_tab:high}
\end{table*}

\subsection{Effectiveness in Text-to-Image Generation.} 
\label{app_sec:t2i_gen}
We further validate the applicability of our method in text-to-image generation, which is more challenging than the class-to-image generation. We adopt PixArt-$\alpha$~\citep{chen2023pixart}, a text-to-image generation model built based on DiT~\citep{peebles2023scalable} as the baseline. PixArt-$\alpha$ is pre-trained on extensive private datasets and exhibits superior text-to-image generation capabilities. 
Our model is initialized using the official PixArt-$\alpha$ checkpoint fine-tuned on the COCO dataset~\citep{lin2014microsoft}. We further fine-tune it with our method to enable dynamic architecture adaptation, resulting in the DyPixArt-$\alpha$ model, as shown in Table~\ref{app_tab:t2i}.  Notably, DyPixArt-$\alpha$ with $\lambda=0.7$ achieves an FID score comparable to the original PixArt-$\alpha$, while significantly accelerating the generation.  %

\begin{table*}[!h]
    \caption{\textbf{Text-to-image generation on COCO~\citep{lin2014microsoft}}. We
    randomly select text prompts from COCO
    and adopt 20-step DPM-solver++~\citep{lu2022dpm} to sample 30,000 images for evaluating the FID score.}    %
    \centering
    \tablestyle{5.0pt}{1.1}
\begin{tabular}{c  c c c  c c}

    \multirow{1}{*}{Model} & \multirow{1}{*}{s/image $\downarrow$ } & \multirow{1}{*}{acceleration $\uparrow$} & \multirow{1}{*}{FLOPs (G) $\downarrow$}  & \multicolumn{1}{c|}{FID $\downarrow$}   & \multicolumn{1}{c}{FID $\Delta \downarrow$}  \\ \midrule[1.2pt]
    PixArt-$\alpha$ & 0.91 & 1.00 $\times$ & 141.09 & 19.88 & +0.00  \\
    DyPixArt-$\alpha$$_{\lambda=0.7}$ & 0.69 & 1.32 $\times$ & 112.44 & 19.75 & -0.13  \\ 
    
    \end{tabular}
    \label{app_tab:t2i}
\end{table*}

\subsection{Exploration of Combining LCM with DyDiT .} 
\label{app_sec:lcm}
Some sampler-based efficient methods~\citep{meng2023distillation, song2023consistency, luo2023latent} adopt distillation techniques to reduce the generation process to several steps. In this section, we combine our DyDiT, a model-based method, with a representative method, the latent consistency model (LCM)~\citep{luo2023latent} to explore their compatibility for superior generation speed. In LCM, the generation process can be reduced to 1-4 steps via consistency distillation and the 4-step generation achieves an satisfactory balance between performance and efficiency. Hence, we conduct experiments in the 4-step setting. Under the target FLOPs ratio $\lambda=0.9$, our method further accelerates generation and achieves comparable performance, demonstrating its potential with LCM. However, further reducing the FLOPs ratio leads to model collapse. This issue may arise because DyDiT's training depends on noise prediction difficulty, which is absent in LCM distillation, causing instability at lower FLOPs ratios. This encourage us to develop dynamic models and training strategies for distillation-based efficient samplers to achieve superior generation efficiency in the future. %

\begin{table*}[!h]
\caption{\textbf{Combining DyDiT with Latent Consistency Model (LCM)~\citep{luo2023latent} .} We conduct experiments under the 4-step LCM setting, as it achieves a satisfactory balance between performance and efficiency. }  
\centering
\tablestyle{5.0pt}{1.1}
\begin{tabular}{c c c c c}
    \multirow{1}{*}{model} & \multirow{1}{*}{s/image $\downarrow$ } & \multirow{1}{*}{FLOPs (G) $\downarrow$}  & \multicolumn{1}{c}{FID $\downarrow$}   & \multicolumn{1}{c}{FID $\Delta \downarrow$}  \\ \midrule[1.2pt]

     DiT-XL+250-step DDPM  &   10.22 & 118.69 & 2.27  & +0.00 \\
     DiT-XL + 4-step LCM  &   0.082 & 118.69 & 6.53  & +4.26 \\
     DyDiT-XL$_{\lambda=0.9}$ + 4-step LCM & 0.076  & 104.43  & 6.52  &  +4.25 \\
    \end{tabular}
    \label{app_tab:lcm}
\end{table*}

\subsection{Comparison with the Early Exiting Method.} \label{app_sec:early}
We compare our approach with the early exiting diffusion model ASE~\citep{moon2023early, moon2024simple}, which implements a strategy to selectively skip layers for certain timesteps. Following their methodology, we evaluate the FID score using 5,000 samples. Results are summarized in Table~\ref{app_tab:early}. Despite similar generation performance, our method achieves a better acceleration ratio, demonstrating the effectiveness of our design. %

\begin{table*}[!h]
\centering
\caption{\textbf{Comparison with the early exiting method \citep{moon2023early, moon2024simple}.} As methods may be evaluated on different devices, we report only the acceleration ratio for speed comparison.} %
\tablestyle{5.0pt}{1.1}
\begin{tabular}{c  c  c c}
    \multirow{1}{*}{model}     & \multirow{1}{*}{acceleration $\uparrow$}  & \multicolumn{1}{c|}{FID $\downarrow$}   & \multicolumn{1}{c}{FID $\Delta \downarrow$ }  \\ \midrule[1.2pt]

    DiT-XL    & 1.00 $\times$   & 9.08 & 0.00 \\
    DyDiT-XL$_{\lambda=0.5}$   &   1.73 $\times$  & 8.95  & -0.13 \\ 
    ASE-D4 DiT-XL &  1.34 $\times$  & 9.09 &  +0.01 \\
    ASE-D7 DiT-XL &  1.39 $\times$  & 9.39 &  +0.31 \\

    \end{tabular}
   \label{app_tab:early}
\end{table*}

\subsection{Training efficiency} \label{app:training_efficiency}
Our approach enhances the inference efficiency of the diffusion transformer while maintaining training efficiency. It requires only a small number of additional fine-tuning iterations to learn the dynamic architecture. In Table~\ref{app_tab:training_efficiency}, we present our model with various fine-tuning iterations and their corresponding FID scores. The original DiT-XL model is pre-trained on the ImageNet dataset over 7,000,000 iterations with a batch size of 256. Remarkably, our method achieves a 2.12 FID score with just 50,000 fine-tuning iterations to adopt the dynamic architecture-approximately 0.7\% of the pre-training schedule. Furthermore, when extended to 100,000 and 150,000 iterations, our method performs comparably to DiT.  We observe that the actual FLOPs during generation converge as the number of fine-tuning iterations increases.

\begin{table*}[!h]
\caption{\textbf{Training efficiency.} The original DiT-XL model is pre-trained on the ImageNet dataset over 7,000,000 iterations with a batch size of 256. }  %
\centering
\tablestyle{5.0pt}{1.1}
\begin{tabular}{c c c c c}

    \multirow{1}{*}{model} & \multirow{1}{*}{fine-tuning iterations} & \multirow{1}{*}{FLOPs (G) $\downarrow$} & \multicolumn{1}{c}{FID $\downarrow$}    & \multicolumn{1}{c}{FID $\Delta \downarrow$}  \\ \midrule[1.2pt]

    DiT-XL   & -        & 118.69 & 2.27   & +0.00 \\
    DyDiT-XL$_{\lambda=0.7}$ &  10,000  (0.14\%) & 103.08  & 45.95 & 43.65   \\   
    DyDiT-XL$_{\lambda=0.7}$ &  25,000  (0.35\%)& 91.97  & 2.97 &  +0.70   \\   
    DyDiT-XL$_{\lambda=0.7}$ &  50,000 (0.71\%) &  85.07 & 2.12 & -0.15   \\      
    DyDiT-XL$_{\lambda=0.7}$ &  100,000 (1.43\%) &  84.30 & 2.17 & -0.10  \\      
    DyDiT-XL$_{\lambda=0.7}$ &  150,000 (2.14\%) &  84.33 & 2.12 & -0.15 \\

    \end{tabular}
    \label{app_tab:training_efficiency}

\end{table*}

\subsection{Data efficiency} \label{app:data_efficiency}
To evaluate the data efficiency of our method, we randomly sampled 10\% of the ImageNet dataset~\citep{deng2009imagenet} for training. DyDiT was fine-tuned on this subset to adapt the dynamic architecture. As shown in Table~\ref{app_tab:data_efficiency}, when fine-tuned on just 10\% of the data, our model DyDiT-XL${_{\lambda=0.7}}$ still achieves performance comparable to the original DiT. When we further reduce the fine-tuning data ratio to  1\%, the FID score increase slightly by 0.06. These results indicate that our method maintains robust performance even with limited fine-tuning data.

\begin{table*}[!h]
\caption{\textbf{Data efficiency.} The slight difference in FLOPs of our models is introduced by the learned TDW and SDT upon fine-tuning convergence. }  %
\centering
\tablestyle{5.0pt}{1.1}
\begin{tabular}{c c c c c}

    \multirow{1}{*}{model} & \multirow{1}{*}{fine-tuning data ratio} & \multirow{1}{*}{FLOPs (G) $\downarrow$} & \multicolumn{1}{c}{FID $\downarrow$}    & \multicolumn{1}{c}{FID $\Delta \downarrow$}  \\ \midrule[1.2pt]

    DiT-XL   & -        & 118.69 & 2.27   & +0.00 \\  
    DyDiT-XL$_{\lambda=0.7}$ &  100\% &  84.33 & 2.12 & -0.15 \\  
    DyDiT-XL$_{\lambda=0.7}$  &   10\% & 84.43 & 2.13 & -0.14 \\  

    DyDiT-XL$_{\lambda=0.7}$ &   1\% & 84.37 & 2.31 & +0.06 \\

    \end{tabular}
    \label{app_tab:data_efficiency}

\end{table*}

\subsection{Dynamic mechanism learned in DyLatte}~\label{app:latte_anay}

To analyze the dynamic allocation mechanism learned for the spatial and temporal layers of Latte~\citep{ma2024latte}, we calculate their average FLOPs ratio in our DyLatte$_{\lambda=0.5}$ (which achieves approximately 50\% FLOPs reduction) relative to the original Latte blocks throughout the generation process. The results are demonstrated in Table~\ref{app_tab:latte_dy}. The results reveal that the computational cost in both spatial and temporal layers is significantly reduced. Moreover, DyLatte learns to reduce slightly more computation in temporal layers, likely due to the higher redundancy present in the temporal dimension of videos.

\begin{table*}[!h]
\caption{\textbf{Dynamic mechanism learned in DyLatte.}  }  
\centering
\tablestyle{5.0pt}{1.1}
\begin{tabular}{ccc}

model & spatial layers FLOPs ratio & temporal layers FLOPs ratio \\ \midrule[1.2pt]
Latte                           & 100\%      & 100\%           \\
DyLatte$_{\lambda=0.5}$                        & 52.82\%      & 49.37\%         
\end{tabular}
\label{app_tab:latte_dy}
\end{table*}

\subsection{Ablation study on DySiT}~\label{app:dysis_ablation}
In Table~\ref{app_tab:dysit_abl}, we conduct an ablation study on DySiT to explore the effectiveness of TDW and SDT. We follow the setup described in Table VII in the main papere and present the results obtained using the ODE sampler. The target FLOPs ratio, $\lambda$, is set to 0.5.

We observe that combining TDW and SDT yields the best performance, emphasizing the importance of both components. Relying solely on SDT forces most tokens to bypass computation in the MLP blocks to meet the FLOPs target, leaving the MHSA blocks as the sole processors of most tokens. This significantly affects training stability. These findings align with observations made in DyDiT.

\begin{table*}[!h]
\caption{\textbf{Ablation study on DySiT.}  }  
\centering
\tablestyle{5.0pt}{1.1}
\begin{tabular}{c |c c | c}

\multirow{1}{*}{Model}  & \multirow{1}{*}{TDW}  & \multirow{1}{*}{SDT} & \multirow{1}{*}{FID  $\downarrow$ } \\
\midrule[1.2pt]

DySiT         &  \ding{51}  &  \ding{51} & \textbf{2.11} \\

DySiT w/o SDT &  \ding{51}  &            & 2.73 \\

DySiT w/o TDW &             & \ding{51}  & unstable training \\

\end{tabular}
\label{app_tab:dysit_abl}
\end{table*}

\subsection{Training DySiT from scratch}~\label{app:dysit_scratch}

To facilitate the training process, we adopt the shorter training schedule used in SiT~\citep{ma2024sit}, training the model for 400,000 iterations from scratch. 
Results are obtained from the SDE sampler. This result in Table~\ref{app_tab:dysit_scratch} demonstrates that our DySiT achieves competitive performance when trained from scratch. 
Since learning a dynamic architecture is more challenging than a static one, its convergence speed is slightly slower, as also observed in DyDiT (Section~\ref{supp:scratch}). Therefore, we recommend fine-tuning publicly available pre-trained model checkpoints rather than training from scratch to avoid `reinventing the wheel'.

\begin{table*}[!h]
\caption{\textbf{Training DySiT from scratch.}  }  
\centering
\tablestyle{5.0pt}{1.1}
\begin{tabular}{c  cc   c c}

    \multirow{1}{*}{Model} & \multirow{1}{*}{training iterations}   & \multicolumn{1}{c}{FID $\downarrow$}   & \multicolumn{1}{c}{FID $\Delta \downarrow$}  \\ 
\midrule[1.2pt]
    SiT-XL & 400,000 & 17.20 & +0.00 \\
    DySiT-XL $_{\lambda=0.5}$ & 400,000 & 19.43  & +2.20 \\
    DySiT-XL $_{\lambda=0.5}$ & 600,000 & 17.21  & +0.01 \\

\end{tabular}
\label{app_tab:dysit_scratch}
\end{table*}

\subsection{Analysis on diffusion models with different prediction modes.}~\label{app:pred_mode}
\begin{figure*}[!h]
  \centering
     \includegraphics[width=0.8\textwidth]{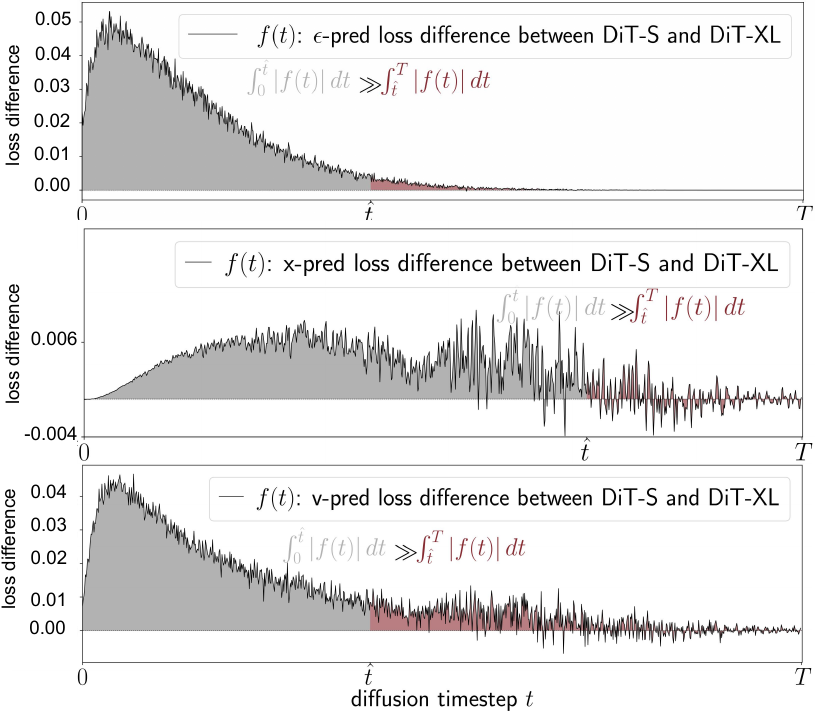}
\caption{\textbf{Analysis on diffusion models with different prediction modes.}} 
\label{app_rebuttal:pred_mode}
\end{figure*}

Given the forward process of a diffusion model, defined as $x_t = \sqrt{\alpha_t} x_0 + \sqrt{1-\alpha_t} \epsilon$ ($\alpha_t$ decreases as $t$ increases), the loss formulations for different prediction modes can be expressed as follows: $\epsilon$-prediction: $\left||\epsilon-\epsilon_\theta\left(x_t, t\right)\right||^2$; $x$-prediction: $\left||x_0 - x_\theta\left(x_t, t\right)\right||^2$; $v$-prediction: $\left||v - v_\theta\left(x_t, t\right)\right||^2$, where $v = \sqrt{\alpha_t}\epsilon - \sqrt{1-\alpha_t}x_0$.  The $\epsilon$-pred is the default setting in both original DiT~\citep{peebles2023scalable} and DyDiT. To better analyze the potential of our method in different prediction modes, we plot the loss differences between the small model (DiT-S) and the large model (DiT-XL) for the $\epsilon$-pred, $x$-pred, and $v$-pred modes in Figure~\ref{app_rebuttal:pred_mode}.

We can find that, all prediction modes exhibit unbalanced loss differences across timesteps, despite having distinct targets. Additionally, the loss gap tends to decrease for timesteps with larger values of $t$. These phenomena suggest potential opportunities to reduce computational redundancy. Since $\epsilon$-prediction is the most widely used prediction mode in diffusion models~\citep{ho2020denoising, nichol2021improved, rombach2022high} and DiT~\citep{peebles2023scalable}, our proposed method, DyDiT, primarily focuses on this setting. Exploring dynamic architectures under other prediction modes is left as a direction for future work.

\subsection{Scalability experiments of DyDiT in the training-from-scratch setting}~\label{app:scala_scratch}

To evaluate the scalability of our method beyond the fine-tuning setting, we conduct experiments by training both DiT and DyDiT from scratch. The results of these experiments are presented in Table~\ref{app_tab:scala_scratch}. Following the short training schedule proposed in \citet{peebles2023scalable} for DiT, we train it for 400,000 iterations. In contrast, DyDiT is trained for 500,000 iterations due to its relatively slower convergence, as discussed in Section\ref{supp:scratch}.

We can observe that, for the smaller models like DyDiT-S and DyDiT-B, larger values of $\lambda$ are needed to achieve competitive performance. However, for the larger model, DyDiT-XL, performance remains comparable to the original DiT-XL even with $\lambda = 0.5$. These results are consistent with the observations made under the fine-tuning setting.

\begin{table*}[!h]
\caption{\textbf{Training DySiT from scratch.}  }  
\centering
\tablestyle{5.0pt}{1.1}
\begin{tabular}{c  cc   c c}
\multirow{1}{*}{model}  & \multirow{1}{*}{FLOPs (G) $\downarrow$}  & \multicolumn{1}{c}{FID $\downarrow$}   & \multicolumn{1}{c}{FID $\Delta \downarrow$}  \\ 
\midrule[1.2pt]
DiT-S & {6.07} & 42.87 & {+0.00} \\ 
DyDiT-S$_{\lambda=0.9}$  & {5.68} & 42.68 & {-0.19} \\
DyDiT-S$_{\lambda=0.7}$  & {4.27} & 45.39 & {+2.52} \\
DyDiT-S$_{\lambda=0.5}$  & {3.07} & 48.77 & {+5.90} \\ \hline

DiT-B & {23.02} & 20.23 & {+0.00} \\
DyDiT-B$_{\lambda=0.9}$  & {21.82} & 19.91 & {-0.32}  \\
DyDiT-B$_{\lambda=0.7}$  & {16.15} & 20.07 & {-0.16}  \\ 
DyDiT-B$_{\lambda=0.5}$  & {11.68} & 22.69 & {+2.46}  \\  \hline

DiT-XL  & {118.69} & 7.21 & {+0.00} \\
DyDiT-XL$_{\lambda=0.9}$ & {107.82} & 6.89 & {-0.32}  \\
DyDiT-XL$_{\lambda=0.7}$ & {83.58} & 7.15  & {-0.06}  \\ 
DyDiT-XL$_{\lambda=0.5}$ & {58.65} & 7.22  & {+0.01}  \\  
\end{tabular}
\label{app_tab:scala_scratch}
\end{table*}

\subsection{Effectiveness of distillation technique in other dynamic models}~\label{app_sec:distill}

Table~\ref{app_tab:distill} provides an evaluation of the effectiveness of the distillation technique developed for DyFLUX, applied to both DyDiT and DyLatte. Both experiments are conducted with $\lambda = 0.5$. We observe that while DyDiT-XL$^{+}$ does not improve the FID score, which is already sufficiently low, it significantly enhances the IS score, indicating an improvement in visual quality. For video generation, DyLatte$^{+}$ also demonstrates improved FVD score.

Below, we present a comparison between the original models, DiT-XL and Latte, and their enhanced counterparts, DyDiT-XL$^{+}$ and DyLatte$^{+}$. Each pair of images or videos is generated from the same initial noise. The results demonstrate that DyDiT-XL$^{+}$ effectively preserves structure, while DyLatte$^{+}$ maintains motion consistency.
\begin{center}
\includegraphics[width=0.8\textwidth]{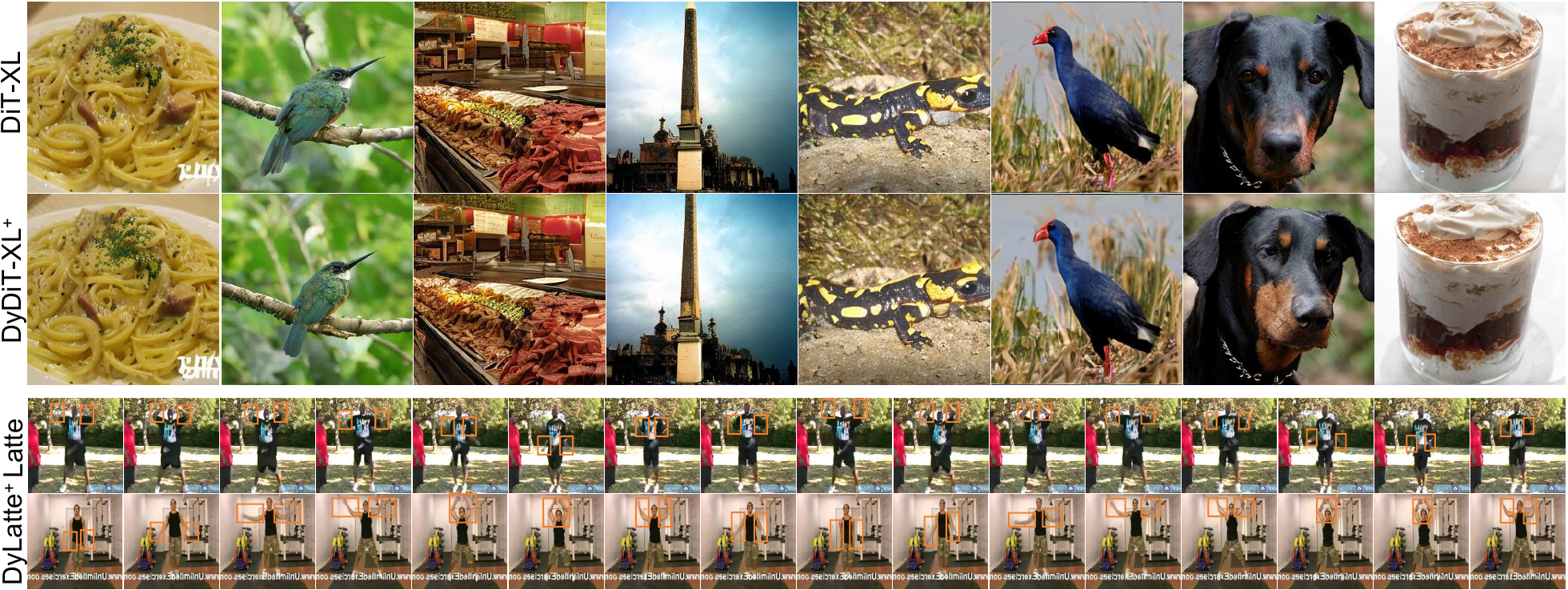}
\end{center}

However, our experiments reveal that it reduces training speed by approximately 20\%, as it requires a full inference process of the original static model. We therefore recommend adopting this technique to further enhance performance where the computational budget allows.

\begin{table}[h] %
\centering
\setlength{\tabcolsep}{12pt}
\renewcommand{\arraystretch}{1.05}
\caption{\textbf{Dynamic models with distillation.}. Model$^{+}$ refers to the model trained using the distillation technique.}
\begin{minipage}[t]{0.3\textwidth} %
\centering
\scriptsize %
\begin{tabular}{c c c} %
Model & FID $\downarrow$ & IS $\uparrow$ \\ %
\midrule[1.2pt] %
DyDiT-XL &  2.07 & 248.03 \\ 
DyDiT-XL$^{+}$ & 2.08 & 265.71 \\  %
\end{tabular}
\end{minipage}%
\begin{minipage}[t]{0.3\textwidth} %
\centering
\scriptsize %
\begin{tabular}{c c } %
Model &  FVD $\downarrow$\\ %
\midrule[1.2pt] %
DyLatte & 181.90  \\ 
DyLatte$^{+}$ & 176.43   \\  %
\end{tabular}
\end{minipage}
\label{app_tab:distill}
\end{table}

\section{Visualization}

\subsection{Qualitative visualization of images generated by DyFLUX} \label{supp:flux}
\begin{figure*}[t]
    \centering
    \includegraphics[width=\textwidth]{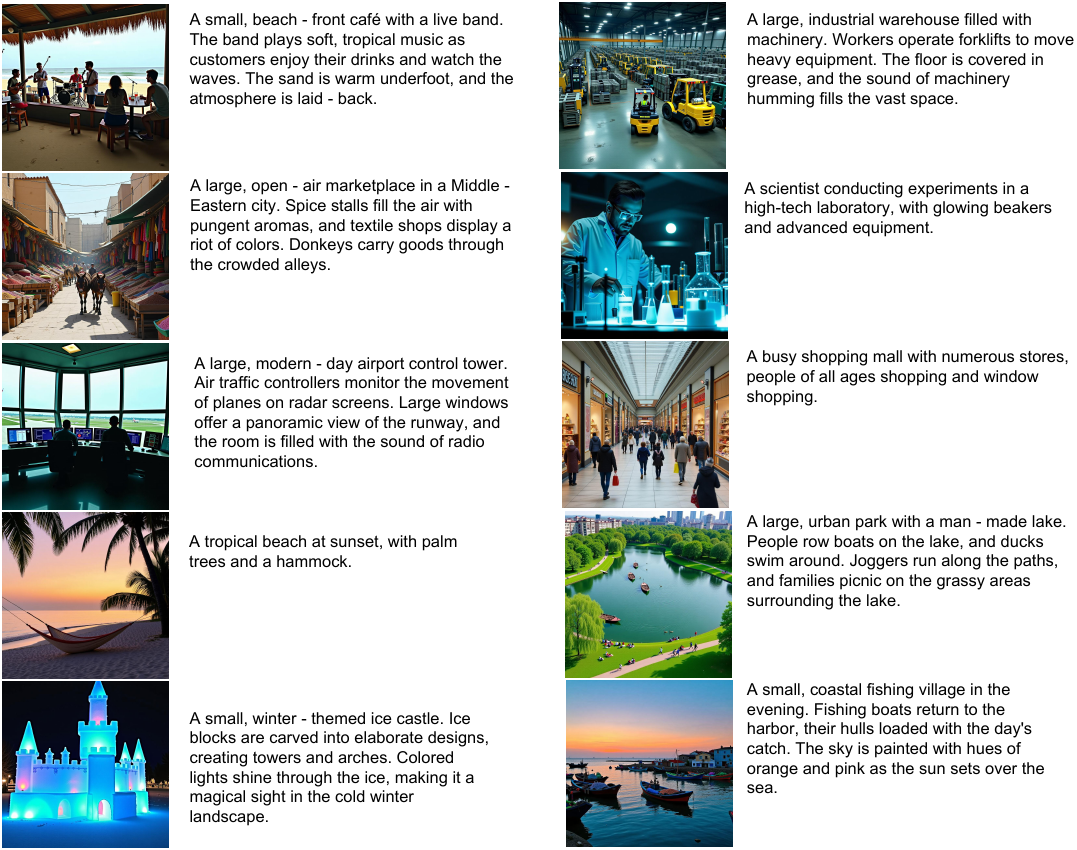}

\caption{\textbf{Qualitative visualization of images generated by DyFLUX 1/2.}}
\label{fig:flux_visuzalization1}  
\end{figure*}

\begin{figure*}[t]
    \centering
    \includegraphics[width=\textwidth]{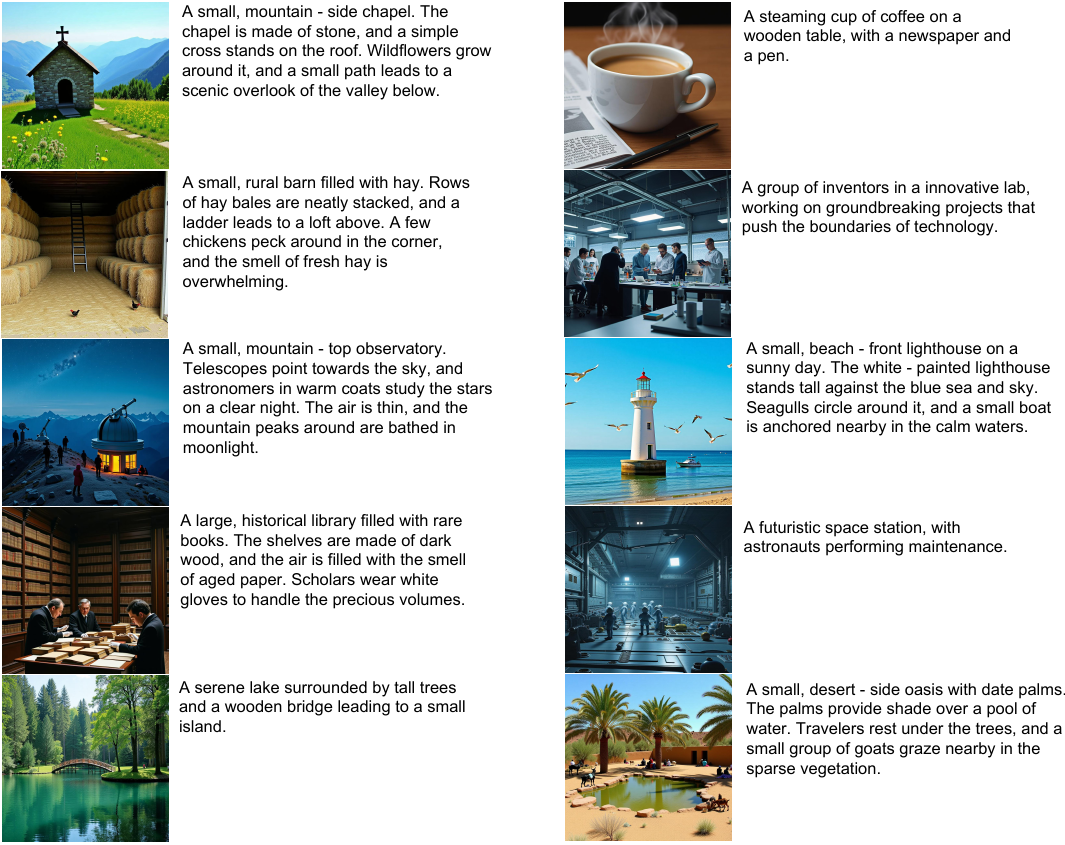}

\caption{\textbf{Qualitative visualization of images generated by DyFLUX 2/2.}}
\label{fig:flux_visuzalization2}  
\end{figure*}

In Figure~\ref{fig:flux_visuzalization1} and Figure~\ref{fig:flux_visuzalization2}, we visualize additionaly images generated by the proposed DyFLUX.

\subsection{Qualitative visualization of images generated by DyDiT-S on fine-grained datasets} \label{supp:fine_grained}
\begin{figure*}[t]
    \centering
    \includegraphics[width=\textwidth]{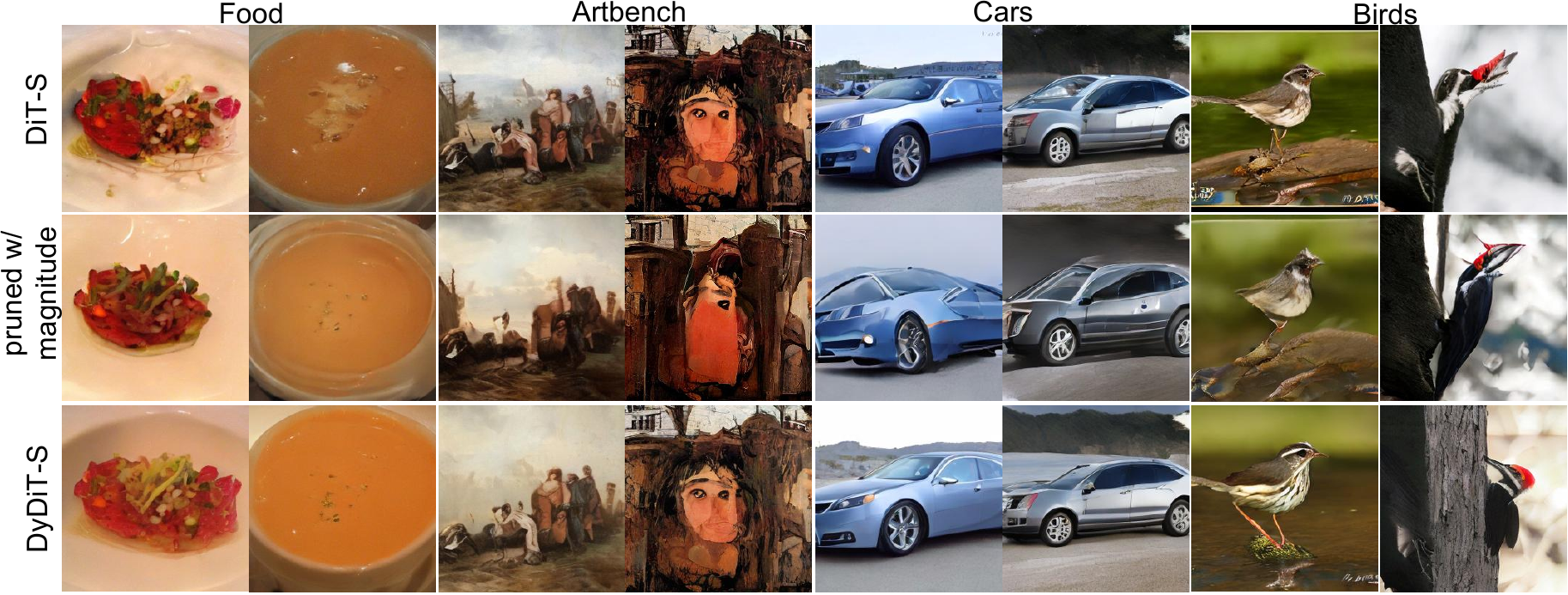}

\caption{\textbf{Qualitative comparison of images generated by the original DiT, DiT pruned with magnitude, and DyDiT.} All models are of ``S'' size. The FLOPs ratio $\lambda$ in DyDiT is set to 0.5.}
\label{fig:visuzalization}  
\end{figure*}

Figure~\ref{fig:visuzalization} presents images generated by DyDiT-S on fine-grained datasets, compared to those produced by the original or pruned DiT-S. These qualitative results demonstrate that our method maintains the FID score while producing images of quality comparable to DiT-S.

\subsection{Additional Visualization of Loss Maps} \label{app_sec:loss_map}
In Figure~\ref{app_fig:loss_map}, we visualize the loss maps (normalized to the range [0, 1]) for several timesteps, demonstrating that noise in different image patches exhibits varying levels of prediction difficulty. %

\subsection{Additional Visualization of Computational Cost on Image Patches} \label{app_sec:flops_map}
In Figure~\ref{app_fig:token_flops}, we quantify and normalize the computational cost across different image patches during generation, ranging from [0, 1]. The proposed spatial-wise dynamic token strategy learns to adjust the computational cost for each image patch.

\subsection{Visualization of samples from DyDiT-XL} \label{app:visualization}
We visualize the images generated by DyDiT-XL$_{\lambda=0.5}$ on the ImageNet~\citep{deng2009imagenet} dataset at a resolution of $256 \times 256$ from Figure~\ref{app_fig:visua1} to Figure~\ref{fig:visua_final_}. The classifier-free guidance scale is set to 4.0. All samples here are uncurated.

\subsection{Visualization of DyDiT with different $\lambda$} \label{rebuttal:visualize_lambda}

We visualize images generated from DyDiT with different  $\lambda$. Images generated from DyDiT-S and DyDiT-XL are presented in Figure~\ref{app_rebuttal:dit-s} and Figure~\ref{app_rebuttal:dit-xl}, respectively.

For DiT-S and DiT-B, increasing $\lambda$ from 0.3 to 0.7 consistently enhances visual quality. At $\lambda=0.9$, DyDiT achieves performance on par with the original DiT-S. In the case of DiT-XL, the visual quality of images generated from DyDiT with $\lambda=0.5$ is comparable to that from the original DiT-XL, attributed to substantial computational redundancy in DiT-XL.

\textcolor{red}{
\begin{figure*}[!h]
  \centering
     \includegraphics[width=1.0\textwidth]{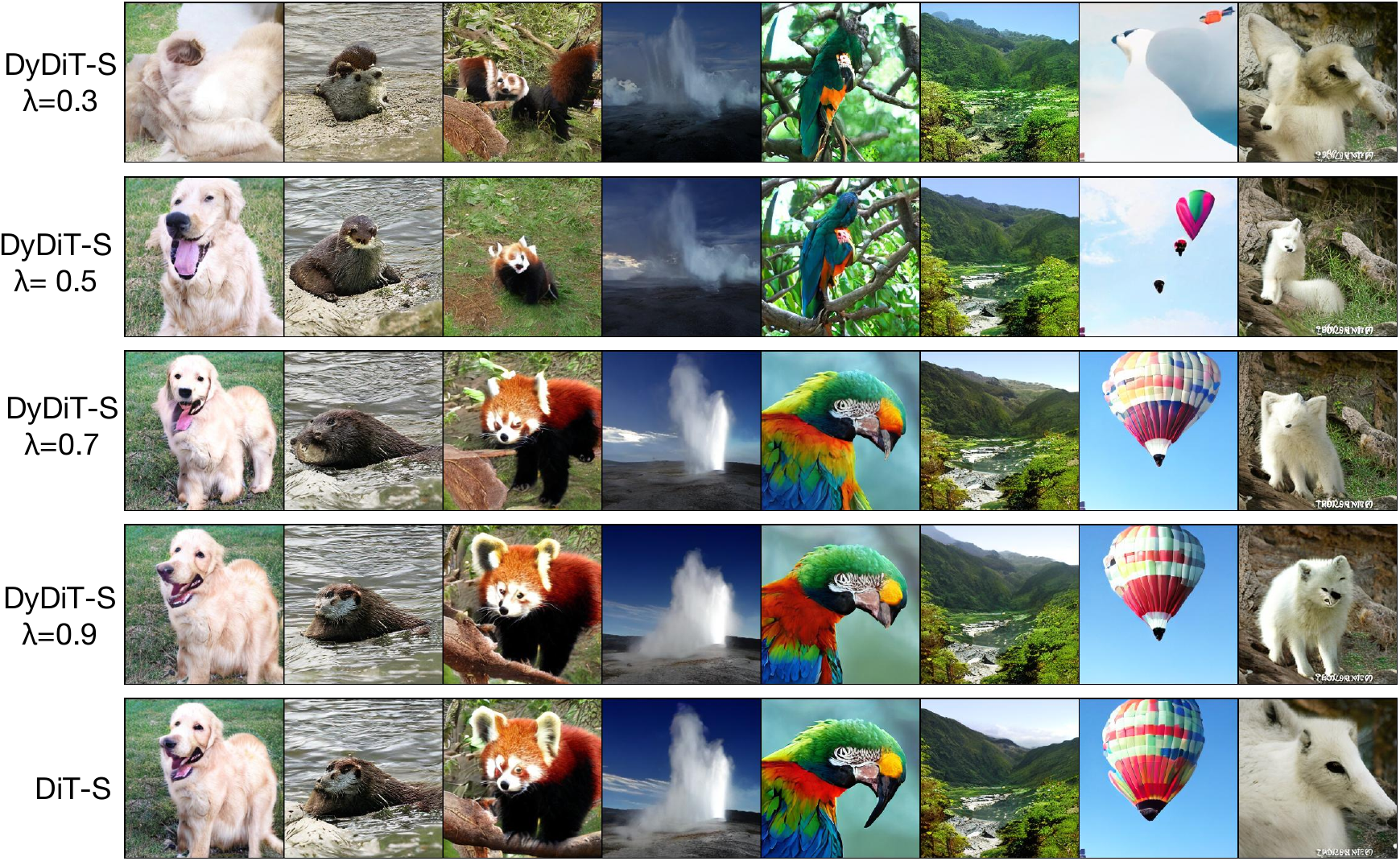}
\caption{\textbf{DyDiT-S.}} 
\label{app_rebuttal:dit-s}
\end{figure*}}

\textcolor{red}{
\begin{figure*}[!h]
  \centering
     \includegraphics[width=1.0\textwidth]{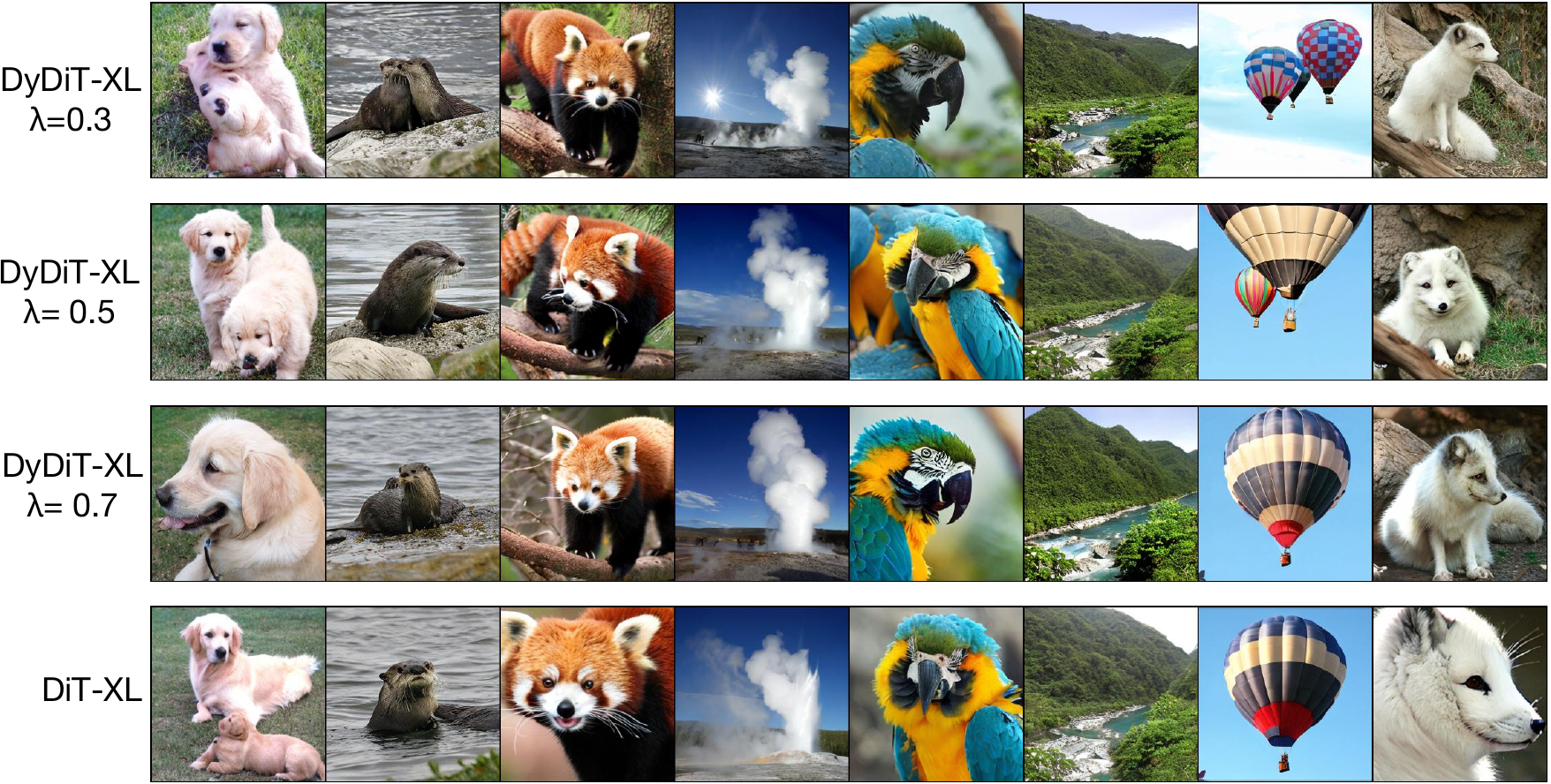}
\caption{\textbf{DyDiT-XL.}} 
\label{app_rebuttal:dit-xl}
\end{figure*}}

\subsection{Visualization of text-to-image generation on COCO} \label{rebuttal:visualize_coco}

\begin{figure*}[!h]
  \centering
     \includegraphics[width=1.0\textwidth]{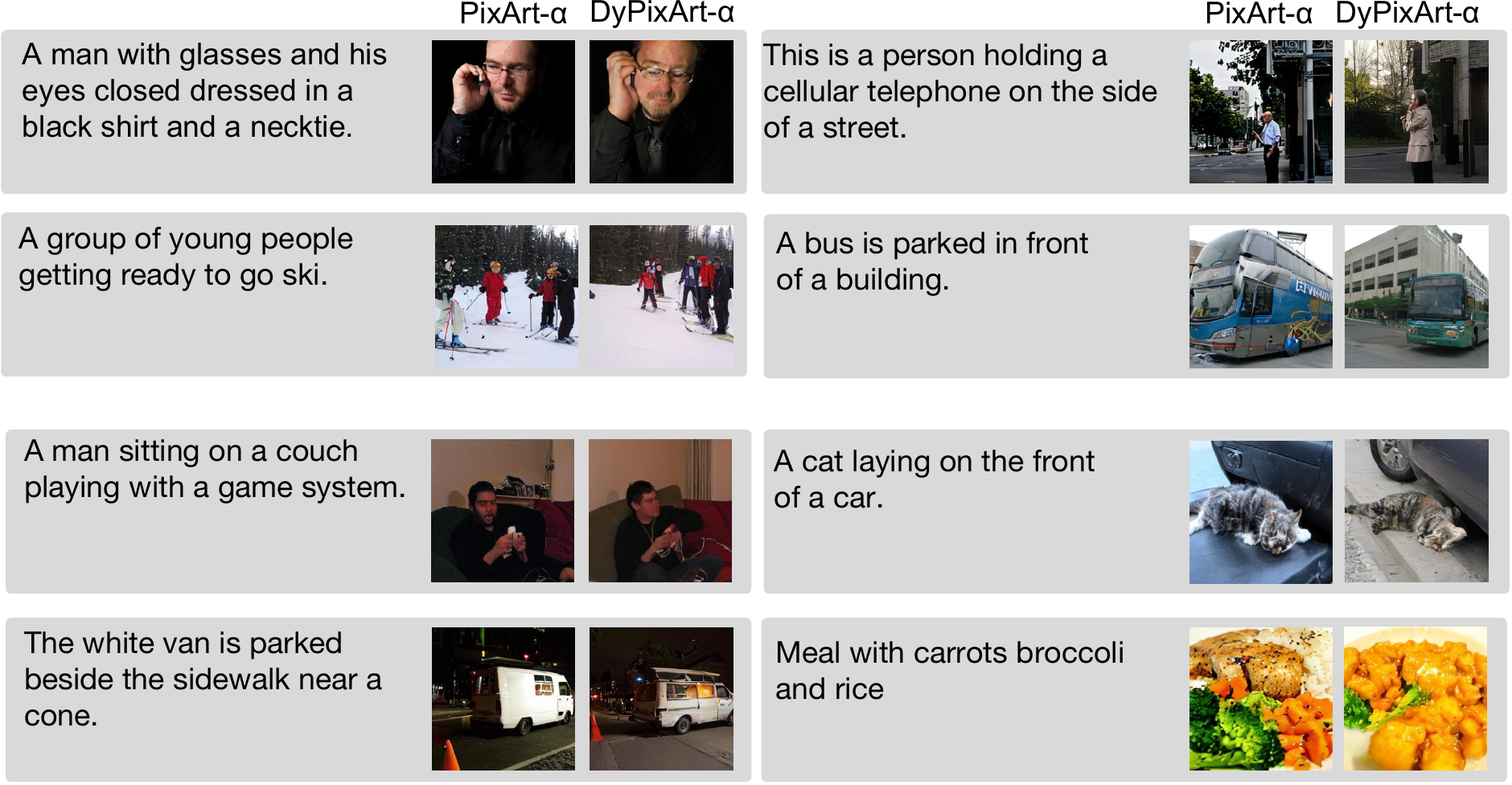}
\caption{\textbf{Visualization from the original PixArt-$\alpha$ and  DyPixArt-$\alpha$ with $\lambda=0.7$.}} 
\label{app_rebuttal:pixart}
\end{figure*}

\begin{figure*}[p]
    \centering
    \includegraphics[width=0.8\textwidth]{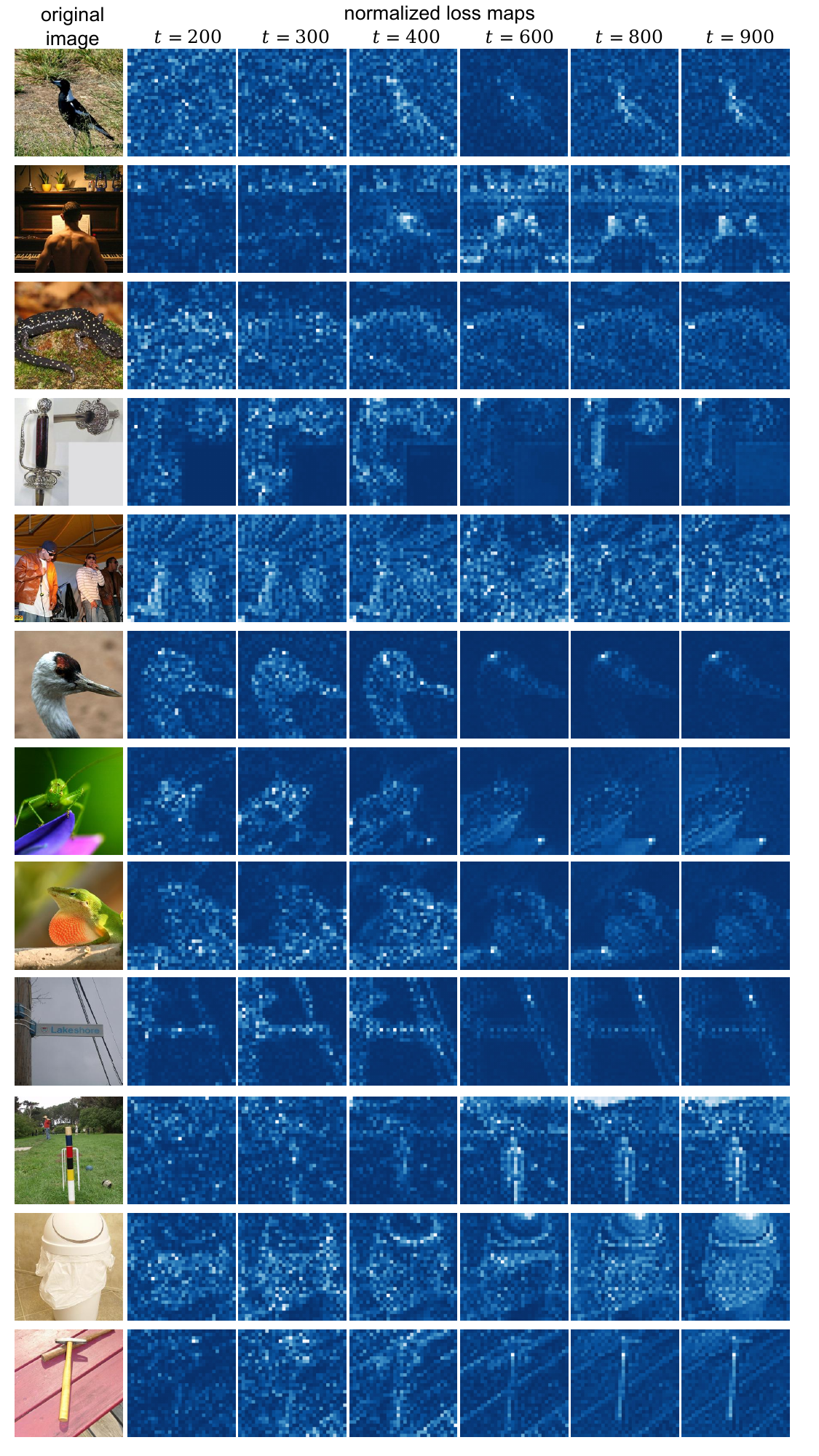}
\caption{\textbf{Additional visualization of loss maps from DiT-XL.} The loss values are normalized to the range [0, 1]. Different image patches exhibit varying levels of prediction difficulty.} 
\label{app_fig:loss_map}
\end{figure*}

\begin{figure*}[p]
    \centering
    \includegraphics[width=0.8\textwidth]{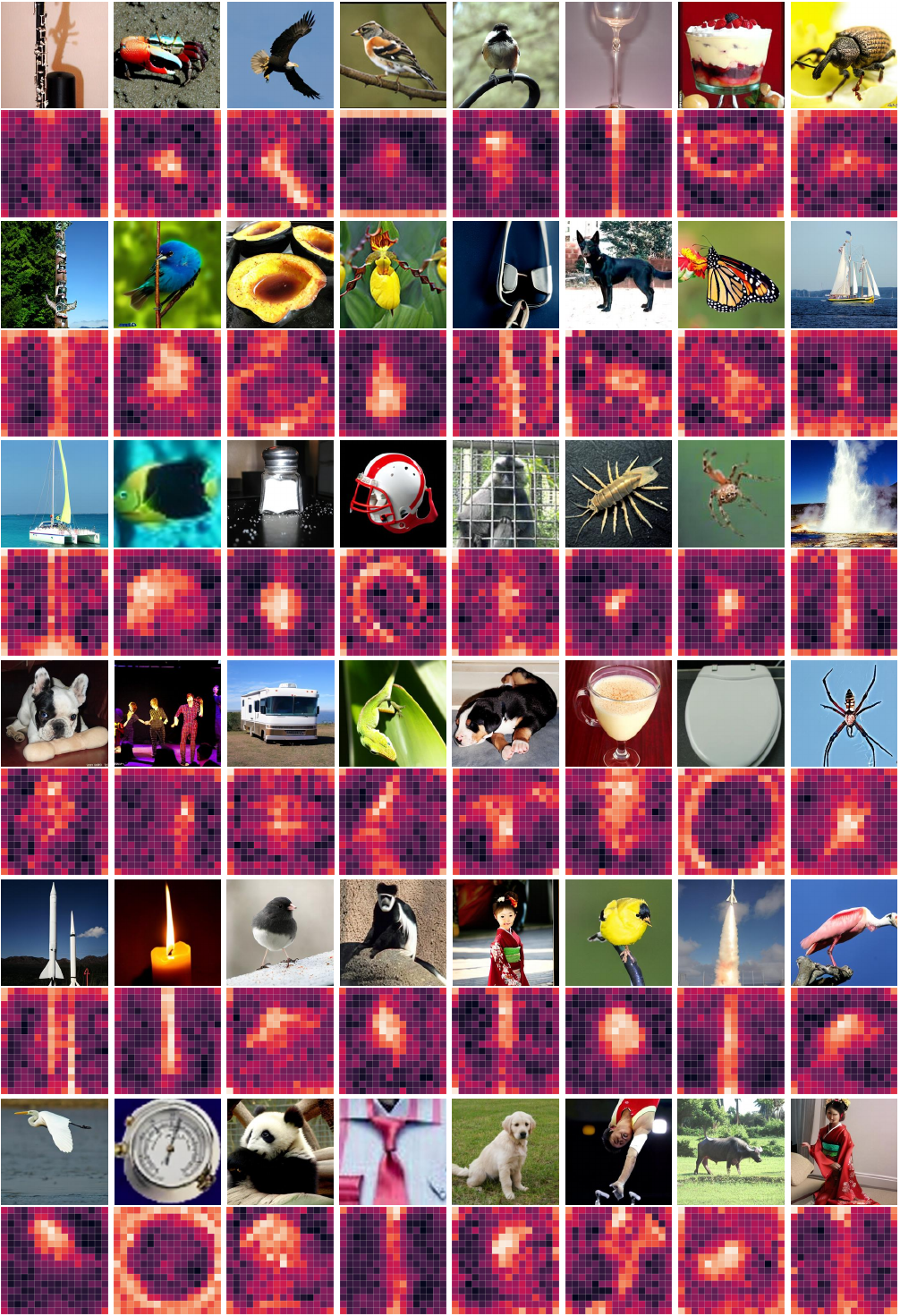}
\caption{\textbf{Additional visualizations of computational cost across different image patches.} Complementary to Figure~\ref{fig:token_flops}, we visualize more generated images and their corresponding FLOPs cost across different image patches. The map is normalized to [0, 1] for clarity.} 
\label{app_fig:token_flops}
\end{figure*}

\begin{figure}[p]
  \centering
    \includegraphics[height=0.48\textheight, width=1.0\textwidth]{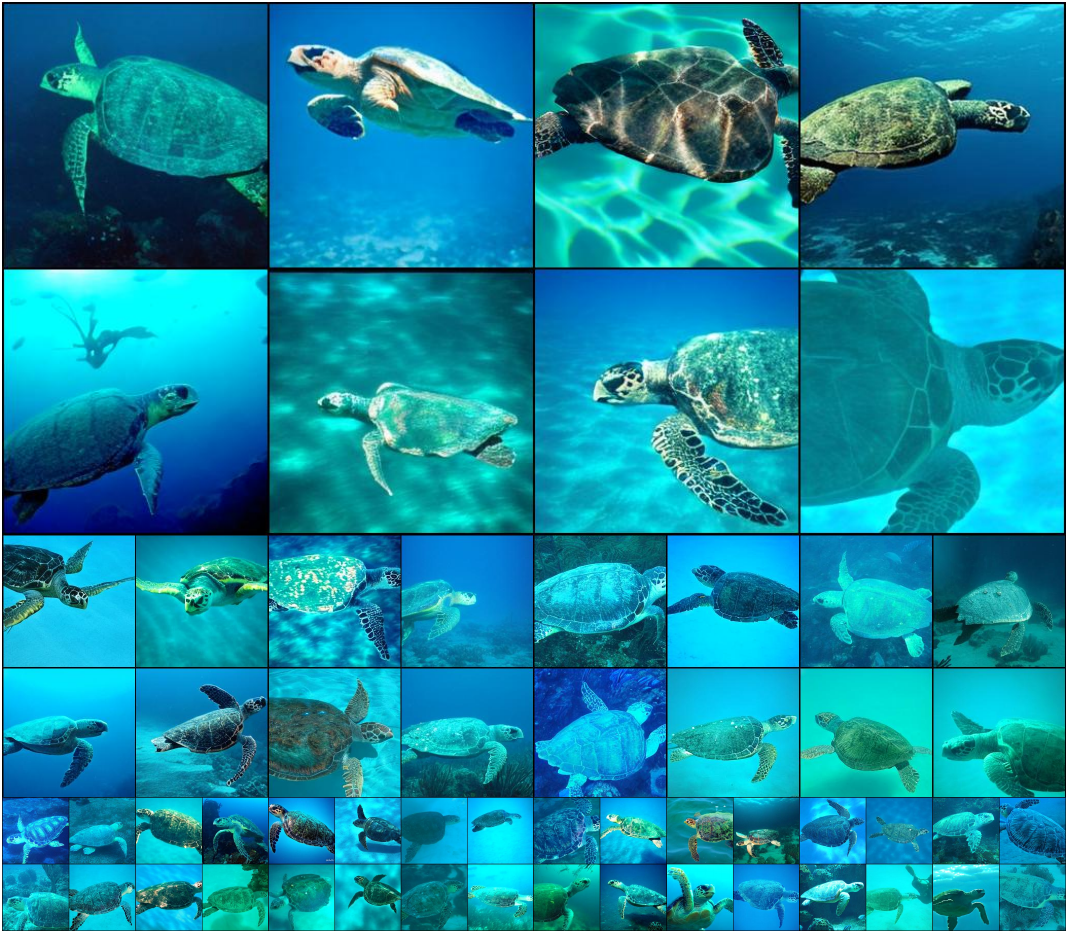}
\caption{\textbf{Uncurated 256$\times$256 DyDiT-XL$_{\lambda=0.5}$ samples. Loggerhead turtle (33).}} 
\label{app_fig:visua1}
  \vfill %
   \includegraphics[height=0.48\textheight, width=1.0\textwidth]
    {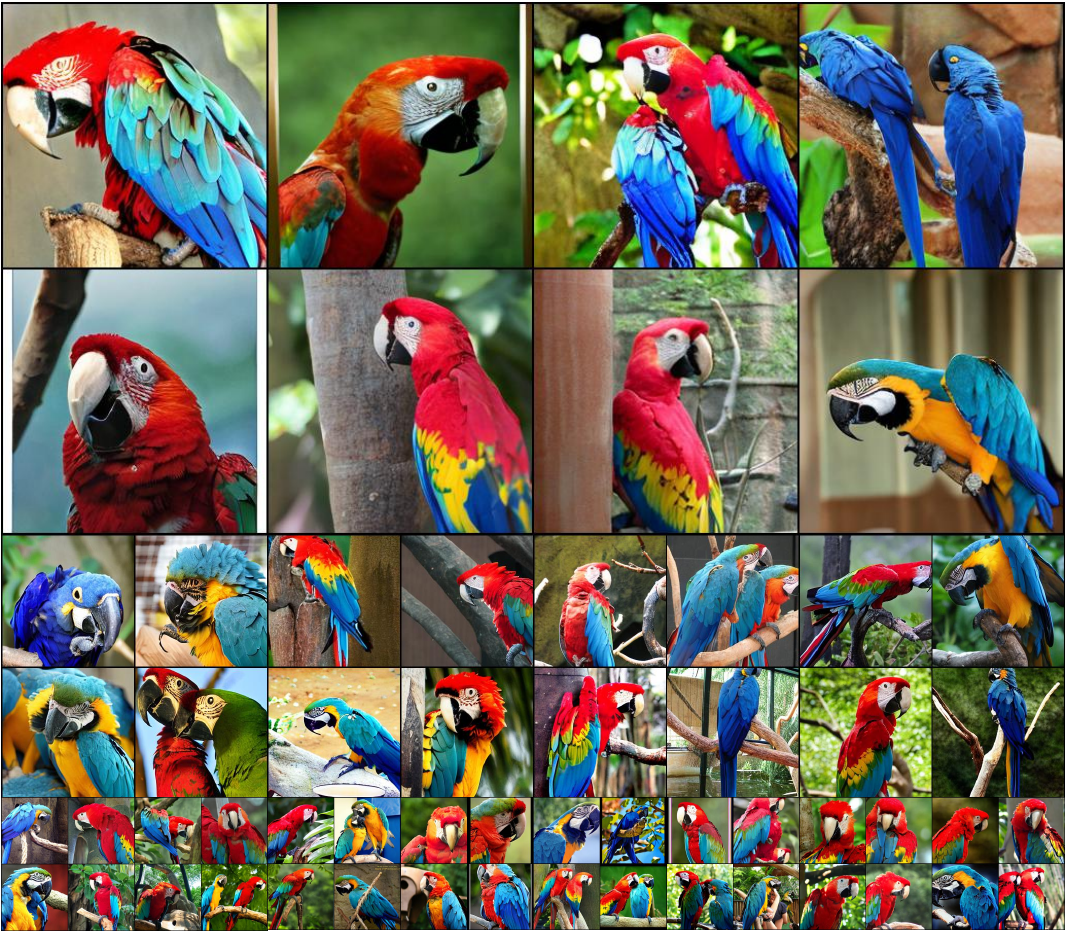}
\caption{\textbf{Uncurated 256$\times$256 DyDiT-XL$_{\lambda=0.5}$ samples. Macaw (88).}} 
\end{figure}

We visualize images generated from the original PixArt-$\alpha$~\citep{chen2023pixart} and our DyPixArt-$\alpha$ with $\lambda=0.7$ in Figure~\ref{app_rebuttal:pixart}. The visual quality of images generated from DyPixArt-$\alpha$ is comparable to that from the original PixArt-$\alpha$.

\newpage

\section{Others}

\subsection{Architecture of Latte and DyLatte} \label{supp:latte}

\paragraph{Latte.}
We present more details about the architecture of Latte~\citep{ma2024latte}. The input video tokens in Latte can be represented as $\mathbf{X} \in \mathbb{R}^{L \times N \times C}$, where $L$, $N$, and $C$ correspond to the temporal, spatial, and channel dimensions of the video in the latent space, respectively. The key to extend DiT~\citep{peebles2023scalable} to video generation lies in incorporating both spatial and temporal modeling for video frames, as opposed to the purely spatial modeling used in original DiT. 
To achieve this, Latte iteratively stacks spatial transformer layers and temporal transformer layers, which can be formulated as:
\begin{equation}
\begin{aligned}
    \mathbf{X} &\leftarrow \mathbf{X} + \alpha_{i} \text{MHSA}_{i}(\gamma_{i} \mathbf{X} + \beta_{i}), \\
\mathbf{X} &\leftarrow \mathbf{X} + \alpha^{\prime}_{i} \text{MLP}_{i}(\gamma^{\prime}_{i} \mathbf{X} + \beta^\prime_{i}),
\end{aligned}
\end{equation}
where $i \in \{\text{spatial}, \text{temporal} \}$. Specifically, $\text{MHSA}_{\text{spatial}}$ and $\text{MHSA}_{\text{temporal}}$ indicate that multi-head self-attention is applied along the spatial and temporal dimensions, respectively, to facilitate token interactions. In contrast, $\text{MLP}_{\text{spatial}}$ and $\text{MLP}_{\text{temporal}}$ operate on individual tokens without sharing parameters between them. The parameters $\{\alpha_{\text{i}}, \gamma_{\text{i}}, \beta_{\text{i}}, \alpha^{\prime}_{\text{i}}, \gamma^{\prime}_{\text{i}}, \beta^{\prime}_{\text{i}} \}$ are derived from an adaLN block~\citep{perez2018film}.

\paragraph{DyLatte.}
To reduce redundancy at the timestep level, we leverage routers to dynamically activate heads in $\text{MHSA}_{\text{spatial}}$ and $\text{MHSA}_{\text{temporal}}$ and channel groups in $\text{MLP}_{\text{spatial}}$ and $\text{MLP}_{\text{temporal}}$. This process can be expressed as:
\begin{equation}
\begin{aligned}
\mathbf{S}_{\text{head}_{i}} &= \operatorname{R}_{\text{head}_{i}}(\mathbf{E}_t) \in [0,1]^H, \\
\mathbf{S}_{\text{channel}_{i}} &= \operatorname{R}_{\text{channel}_{i}}(\mathbf{E}_t) \in [0,1]^H,
\end{aligned}
\end{equation}
where $i \in \{\text{spatial}, \text{temporal} \}$.

Meanwhile, to address spatial-temporal redundancy in token processing, we introduce two routers that dynamically select tokens to skip the computation of MLP blocks in both spatial and temporal transformer layers. This can be expressed as:
\begin{equation}
\mathbf{S}_{\text{token}_{i}} = \operatorname{R}_{\text{token}_{i}}(\mathbf{X})\in[0,1]^N,
\end{equation}
where $i \in \{\text{spatial}, \text{temporal} \}$.

\subsection{Details of internal dataset used in DyFLUX}~\label{supp:internal_dataset}

Our internal dataset contains approximately 195 million high-quality images, each accompanied by a densely annotated prompt generated using AI tools. The size of the images, in terms of height and width, primarily ranges from 1,024 to 2,048. We demonstrate is their percentage distribution in Table~\ref{supp_tab:height-distribution} and Table~\ref{supp_tab:weight-distribution}, respectively. Furthermore, we categorize the prompts into 10 distinct groups, providing their distribution in Table~\ref{supp_tab:prompt} and examples for each category in Table~\ref{supp_tab:prompt_example}.
\vspace{1cm}

\begin{table}[h!]
\centering
\tablestyle{5.0pt}{1.1}
\begin{minipage}{0.48\textwidth}
    \centering
    \begin{tabular}{cc}
    {height range (pixels)} & {percentage} \\  \midrule[1.2pt]
    1024--1535                     & 87.24\%             \\
    1536--2048                     & 12.76\%             \\
    \end{tabular}
    \caption{Height distribution of images.}
    \label{supp_tab:height-distribution}
\end{minipage}%
\hfill
\begin{minipage}{0.48\textwidth}
    \centering
    \begin{tabular}{cc}
    {weight range (pixels)} & {percentage} \\  \midrule[1.2pt]
    1024--1535                     & 50.62\%             \\
    1536--2048                     & 49.38\%             \\
    \end{tabular}
    \caption{Weight distribution of images.}
    \label{supp_tab:weight-distribution}
\end{minipage}
\end{table}

\begin{table}[h!]
\centering
\tablestyle{5.0pt}{1.1}
\begin{tabular}{cc}
{category} & {pcercentage} \\ \midrule[1.2pt]
1 Nature \& Landscape & 25.90\% \\
2 Human Activities & 20.50\% \\ 
3 Architecture \& Buildings & 15.50\% \\ 
4 Objects \& Still Life & 13.10\% \\
5 Food \& Cuisine & 10.10\% \\ 
6 Animals \& Wildlife & 8.10\% \\ 
7 Art \& Culture & 3.30\% \\ 
8 Abstract \& Patterns & 1.60\% \\ 
9 Transportation & 1.00\% \\ 
10 Technology \& Science & 0.90\% \\
\end{tabular}
\caption{Prompt category distribution.}
\label{supp_tab:prompt}
\end{table}

\begin{longtable}{|>{\RaggedRight\arraybackslash}p{3.5cm}|>{\RaggedRight\arraybackslash}p{12cm}|}
\hline
\multicolumn{1}{|c|}{category} & \multicolumn{1}{c|}{description} \\
\hline
1 Nature \& Landscape & The image is a high-resolution photograph capturing a close-up view of a cluster of blue hyacinth flowers in full bloom. The flowers are predominantly purple-blue, with a slight gradient towards lighter shades at the edges. Each flower has multiple delicate, slender petals that form a layered, bell-shaped arrangement. The petals have a smooth texture and are slightly translucent, allowing the greenish-yellow stamens to be visible at the center of each flower. The background is blurred, creating a shallow depth of field effect that focuses attention on the flowers. The background consists of a neutral, light gray wall, which contrasts beautifully with the vibrant blue of the flowers, making them stand out prominently. The lighting is natural, likely taken outdoors on a bright day, as evidenced by the even illumination and the absence of harsh shadows. The overall composition of the photograph emphasizes the delicate beauty and intricate details of the flowers, capturing the natural elegance of the hyacinth blooms. The photograph is a fine example of macro photography, highlighting the minute textures and colors of the flowers. \\
\hline
{2 Human Activities} & The image is a photograph capturing a joyful moment between a man and a woman in an outdoor setting. The man, positioned slightly behind and to the left of the woman, has short, gray hair and is wearing a white shirt. The woman, standing in the foreground, has long, dark brown hair and is smiling broadly, looking directly at the camera. She is dressed in a long-sleeved, maroon sweater. Both individuals are holding hands, their arms crossed in front of them, and are standing in a sunlit area, suggesting it is daytime. The background is filled with lush green foliage, including trees and bushes, with dappled sunlight filtering through the leaves, creating a serene and natural ambiance. The overall mood of the image is warm and cheerful, emphasizing the couple's connection and happiness. The photograph is well-lit, with natural light highlighting their expressions and the vibrant colors of their clothing and the surroundings. The texture of the leaves and the smooth fabric of their clothing contrast beautifully, adding depth to the scene. \\
\hline
{3 Architecture \& Buildings} & The image is a high-resolution photograph capturing a section of an ancient Mayan pyramid, likely located in a tropical forest. The pyramid, made of large, irregularly shaped stones, ascends from the lower left corner of the image to the upper right. The stones are predominantly light gray with some darker shades, creating a textured and rugged appearance. The structure is covered in lush greenery, with dense foliage and trees climbing up the steep sides, indicating a natural growth over the pyramid. The top of the pyramid is partially obscured by the dense vegetation, but it appears to have a flat, horizontal surface. The sky above is a clear, bright blue with a few scattered white clouds, suggesting a sunny day. In the foreground, a narrow, grassy pathway leads up to the base of the pyramid, and a few people are visible, wearing casual attire, suggesting a modern tourist site. The overall scene combines the ancient architectural wonder with the vibrant, living environment of the surrounding forest, creating a striking juxtaposition of past and present. \\
\hline
{4 Objects \& Still Life} & The image is a high-resolution photograph showcasing three metallic bolts against a stark white background. The bolts are arranged in a triangular formation, with the largest bolt positioned at the top and the other two slightly smaller ones below, forming a pyramid-like structure. All three bolts are made of a shiny silver metal, likely stainless steel, which reflects light and gives them a polished, metallic appearance. Each bolt has a hexagonal head with six sides, and the top bolt has a larger head compared to the others. The bolts' threads are clearly visible, indicating their functional purpose. The background is completely white, devoid of any additional objects or textures, ensuring that the bolts are the sole focus of the image. The lighting is bright and even, highlighting the smooth surfaces and intricate details of the bolts, making them appear sharp and well-defined. The overall composition and simplicity of the image emphasize the industrial and utilitarian nature of the bolts. \\
\hline
{5 Food \& Cuisine} & This is a high-resolution photograph featuring a vibrant arrangement of red chili peppers against a stark white background. The image showcases twelve chili peppers, each with a distinctively curved and elongated shape, varying in length from approximately 4 to 6 inches. The peppers are arranged diagonally, creating a dynamic and visually engaging composition. The chili peppers exhibit a glossy texture, with a smooth, shiny surface that reflects light, giving them a rich, almost metallic appearance. Each pepper has a slender, green stem that extends from the top, tapering into a sharp point. The peppers are uniformly bright red, with no signs of blemishes or imperfections, indicating freshness and high quality. The white background provides a stark contrast, making the vibrant red peppers stand out prominently. The photograph captures the natural texture and color of the chili peppers, emphasizing their vibrant and fresh appearance. The overall aesthetic is clean, minimalistic, and focused, drawing attention to the natural beauty and culinary significance of the chili peppers. \\
\hline
{6 Animals \& Wildlife} & This photograph captures a swarm of bats in flight against a dark, almost black background. The bats are densely clustered, creating a dynamic and chaotic visual effect. Their wings are spread wide, and their bodies are elongated, with a slight curve at the neck. The bats are predominantly white or light gray, with subtle variations in shading that give them a textured appearance. In the upper part of the image, the bats are more dispersed, suggesting they are in the process of taking off. The lower part of the image shows a denser cluster, indicating they are in mid-flight or descending. The bats' wings are thin and delicate, with a slight translucency that makes them appear almost ghostly against the dark background. In the bottom right corner, there is a rocky outcrop that provides a natural, rugged texture. The rocks are a mix of light brown and beige, with rough, jagged edges. The contrast between the bats' light colors and the dark background, along with the ruggedness of the rock, adds depth and complexity to the image. The overall mood of the photograph is dynamic and slightly eerie, capturing the essence of nocturnal activity in a natural setting. The image is rich in detail, highlighting the intricate movements and forms of the bats. \\
\hline
{7 Art \& Culture} & This photograph captures a traditional Chinese opera performance. At the center of the image, two actors are dressed in elaborate costumes, with intricate make-up and headpieces. The actor on the left is a young man with a pale complexion, wearing a black robe with blue trim and a black hat adorned with a blue ribbon. He is extending his hand towards the actor on the right, who is an older man with a weathered appearance. The older actor is dressed in a traditional opera costume with a black robe and a white undergarment, and he wears a large, ornate headpiece with intricate patterns and a tall, curved shape. The woman in the middle, also an actor, is wearing a white robe with red and black accents, and her face is heavily made-up with exaggerated features, including a wide nose and large eyes. The background features a large, green tent-like structure with a pink light illuminating it, suggesting an outdoor setting. The scene is rich in cultural and historical detail, capturing the essence of traditional Chinese opera. \\
\hline
{8 Abstract \& Patterns} & This image is a digital abstract artwork featuring a complex, layered composition. Dominating the image are several large, overlapping circles of varying sizes in shades of blue and teal. These circles are not perfectly round but have a slightly irregular, fragmented appearance. The background is a light beige color, providing a stark contrast to the dark blue and teal hues of the circles. Within the circles, there is a dense network of interwoven lines and shapes, giving the impression of a maze-like structure. These lines are in shades of blue, teal, and white, creating a sense of depth and dimension. The overall texture is smooth and glossy, suggesting a digital medium. The composition is dynamic, with the circles appearing to float or intersect in a chaotic yet balanced manner. The smaller, fragmented shapes within the circles add a sense of movement and complexity. The image has a modern, abstract style, with a focus on geometric forms and color contrast. The style is reminiscent of digital art, with a strong emphasis on clean lines and precise digital manipulation. The overall effect is visually engaging and evokes a sense of exploration and discovery within the abstract landscape. \\
\hline
{9 Transportation} & The image is a high-resolution photograph capturing a close-up view of a blue agricultural tractor and its attached plow situated in a rural field. The tractor, occupying the left side of the image, features a large, rugged design with a predominantly blue color scheme, complemented by metallic silver and black elements. The plow, attached to the tractor, extends from the right side of the frame, displaying a series of metal blades that are designed to break up and turn soil. These blades are positioned at various angles, suggesting they are ready for use. The ground in the foreground is a mix of brown soil and patches of green grass, indicating a recently cultivated or plowed field. The background shows a vast expanse of green, likely a field of crops, under a slightly overcast sky, which casts a soft, diffused light over the scene. The texture of the soil is rough and uneven, contrasting with the smooth, metallic surfaces of the tractor and plow. The overall scene conveys a sense of industriousness and agricultural activity in a rural setting. \\
\hline
{10 Technology \& Science} & The image is a high-resolution photograph depicting a close-up view of a pair of hands holding two syringes against a teal medical scrubs background. The hands are positioned centrally in the frame, with the left hand holding a syringe filled with a clear liquid and the right hand holding a smaller, empty syringe. The syringe in the left hand has a black and white striped cap and a silver needle, while the syringe in the right hand has a green and yellow striped cap and a silver needle. The background shows a portion of the teal scrubs, which have a textured, slightly rough fabric. A stethoscope is draped around the neck of the person wearing the scrubs, with the chest piece resting on the scrubs. The skin tone of the hands is light, and the hands appear to be of a healthcare professional, possibly a nurse or doctor. The overall scene conveys a medical context, with the syringes likely used for injections or medical procedures. The photograph is well-lit, highlighting the details of the syringes and the texture of the scrubs. \\
\hline
\caption{Examples of prompts.}
\label{supp_tab:prompt_example}
\end{longtable}

\subsection{Loss pattern discrepancy of DiT and SiT}~\label{supp:loss_disc}  
From Figure 2 in the main paper, we observe a distinct loss pattern difference between flow matching and the diffusion model. The divergence likely arise from the difference in prediction targets between the diffusion model and the flow matching model. Specifically, we have:

\begin{itemize}
    \item Diffusion model (DiT~\citep{peebles2023scalable}):
    In diffusion framework of DiT, the prediction target is the initial noise (commonly referred to as $\epsilon$-prediction). The loss function is formulated as:
    $\mathcal{L}_{\text {noise}}=\mathbb{E}_{t, x_0, \epsilon, }\left[\left\|\epsilon-\epsilon_\theta\left(x_t, t\right)\right\|^2\right]$, where $x_t=\sqrt{\alpha_t} x_0+\sqrt{1-\alpha_t} \epsilon$. Here, $\alpha_t$ decreases as $t$ increases, meaning that as $t$ grows larger, $x_t$ becomes dominated by the noise $\epsilon$. Consequently, \textit{for larger $t$, $x_t$ is closer to noise $\epsilon$}, which reduces the prediction difficulty for the model $\epsilon_\theta$. This results in \textit{lower loss differences} at larger timesteps.

    \item Flow matching model (SiT~\citep{ma2024sit}): 
    In the flow matching framework, the prediction target is the velocity field. The loss function is formulated as: $\mathcal{L}_\text{velocity} = \mathbb{E}_{t,x_0,x_1} \left\| v_\theta(x_t, t) - v \right\|^2$, where $v$ represents the ground-truth velocity, obtained as $v=x_T - x_0$ under the linear transport assumption used in SiT. A key observation is that the velocity target $v$ remains \textit{consistent} across different timesteps $t$, as it directly depends on the starting point $x_0$ and the end point $x_T$.  
    This consistency ensures that the \textit{variation of prediction difficulty is not as significant as that in DiT}, resulting in a relatively stable pattern of loss differences.
    
\end{itemize}

\subsection{Frequently asked questions} \label{rebuttal:freq_ques}

\paragraph{Question1: It is unclear how the ``pre-define'' in L214 benefit the sampling stage?}

Pre-define enables batched inference of our method. The activation of heads and channel groups in TWD relies solely on the timestep $t$, allowing us to pre-calculate activations prior to deployment. By storing the activated indices for each timestep, we can directly access the architecture during generation for a batch of samples. This approach eliminates the sample-dependent inference graph typical in traditional dynamic architectures, enabling efficient and realistic speedup in batched inference.

\paragraph{Question2: The proposed modules to efficient samplers or to samplers with varying sampling steps remains unclear.}

Consistent with standard practices in samplers such as DDPM, varying the sampling steps translates to differing timestep intervals. We adopt its official code to map $t$ into the range 0–1000, aligning with the 1000 total timesteps used during training. For example, in DDPM with 100 and 250 timesteps:

\emph{a)}  250-DDPM timestep: we map $t \in [249, ....5,4,3,2,1,0]$ \ into $t_{\text{250-DDPM}} \in [999, 995, .....20, 16, 12, 8, 4, 0]$.

\emph{b)}  100-DDPM timestep: we map $t \in [99, 98, ...2,1,0]$ into $t_{\text{100-DDPM}} \in [999, 989, ...20, 10, 0]$.

In TWD, we adopt $t_{\text{250-DDPM}}$ and $t_{\text{100-DDPM}}$ to predict activation masks. When $t_{\text{250-DDPM}}=t_{\text{100-DDPM}}$, the denoising process is at the same stage, resulting in identical activation masks from TWD.

\paragraph{Question3: Are there any suggestions about the selection of $\lambda$?}

\emph{a)} Depending on computational resources, users may select different $\lambda$ values during fine-tuning to balance efficiency and performance.

\emph{b)} We recommend initially setting $\lambda = 0.7$, as it generally delivers comparable performance. If the results are satisfactory, consider reducing $\lambda$ (e.g., to 0.5) for further optimization. Conversely, if performance is inadequate, increasing $\lambda$ may be beneficial.

\newpage

\begin{figure}[p]
  \centering
       \includegraphics[height=0.48\textheight, width=1.0\textwidth]{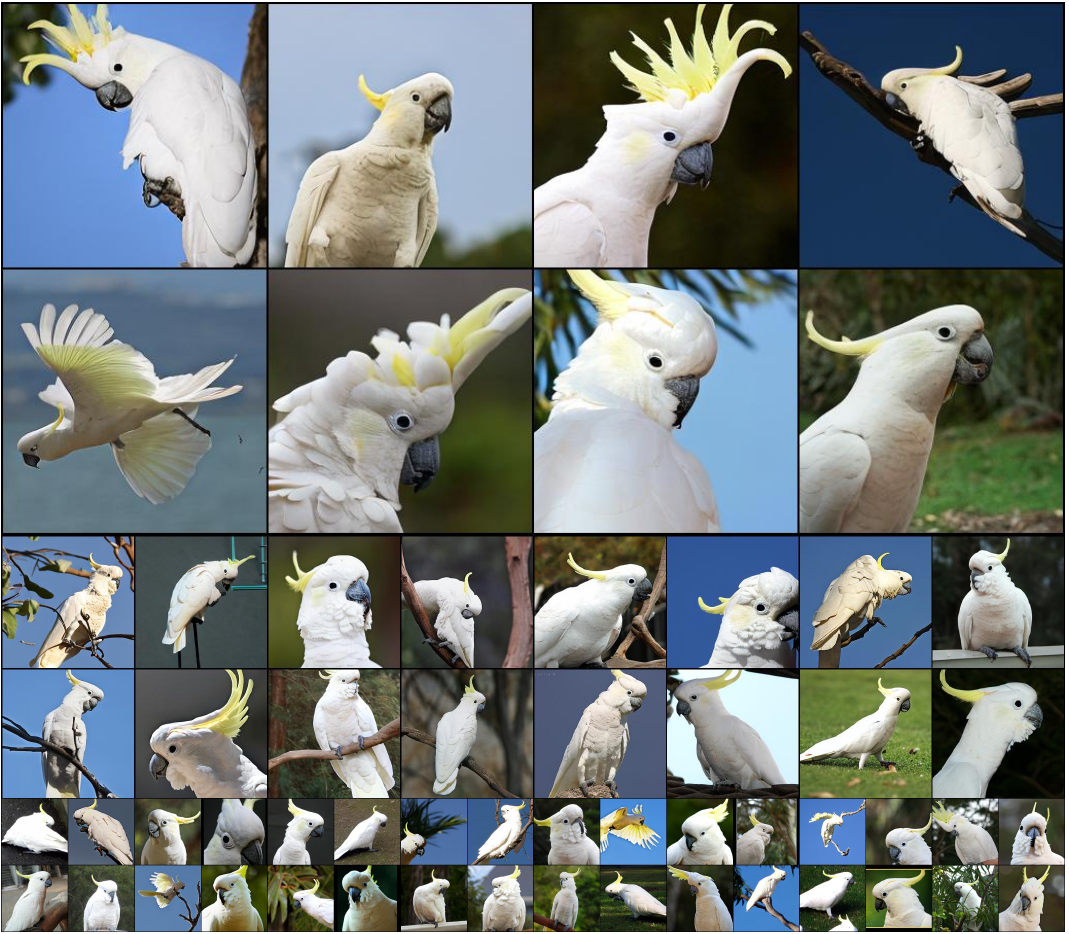}
\caption{\textbf{Uncurated 256$\times$256 DyDiT-XL$_{\lambda=0.5}$ samples. Kakatoe galerita (89).}} 
  \vfill %
   \includegraphics[height=0.48\textheight, width=1.0\textwidth]{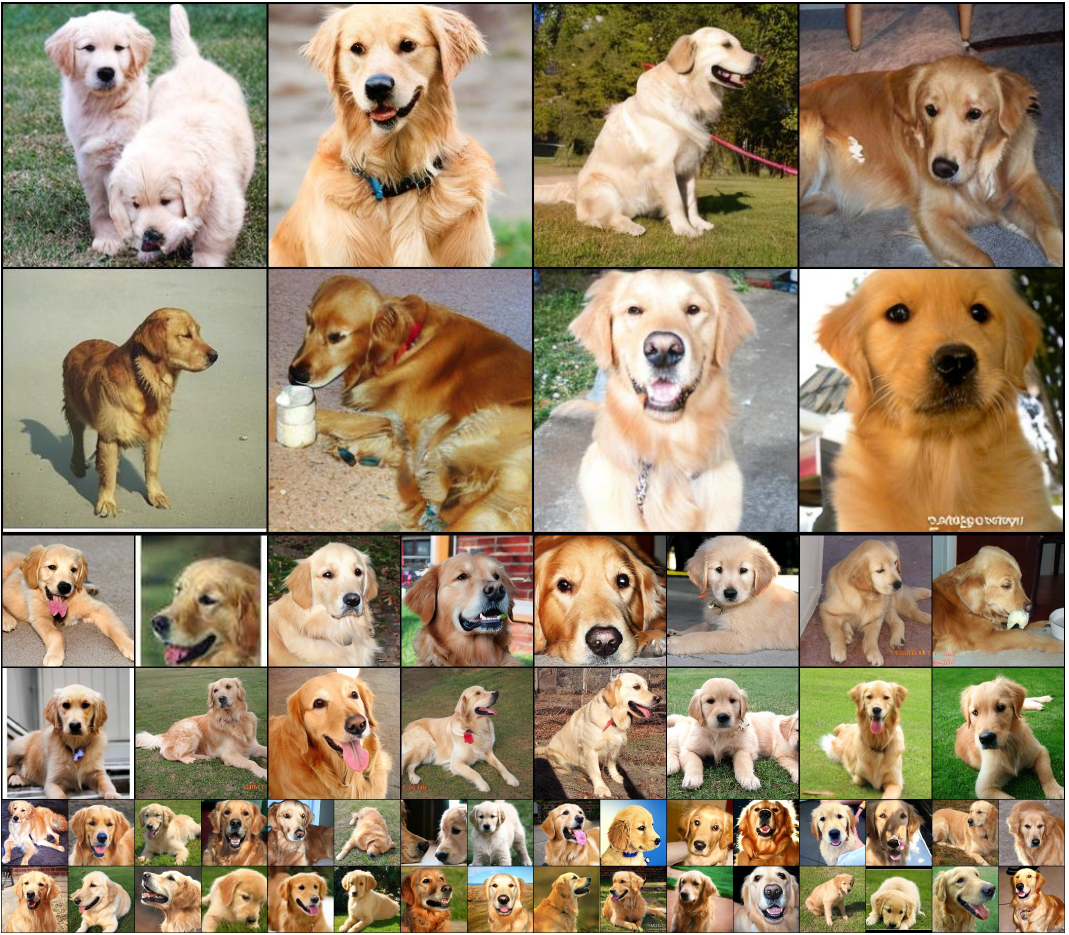}
\caption{\textbf{Uncurated 256$\times$256 DyDiT-XL$_{\lambda=0.5}$ samples. Golden retriever (207).}} 
\end{figure}

\begin{figure}[p]
  \centering
   \includegraphics[height=0.48\textheight, width=1.0\textwidth]{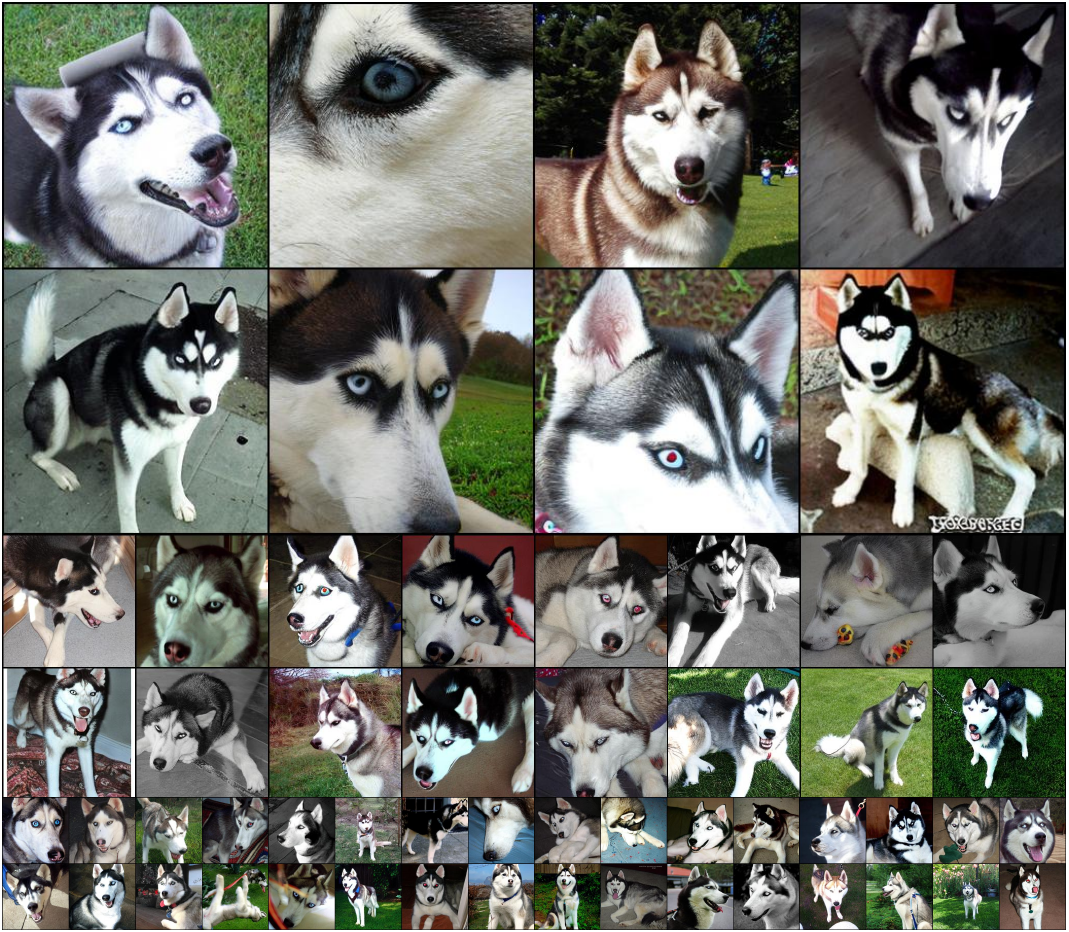}
\caption{\textbf{Uncurated 256$\times$256 DyDiT-XL$_{\lambda=0.5}$ samples. Siberian husky (250).}} 
  \vfill %
   \includegraphics[height=0.48\textheight, width=1.0\textwidth]{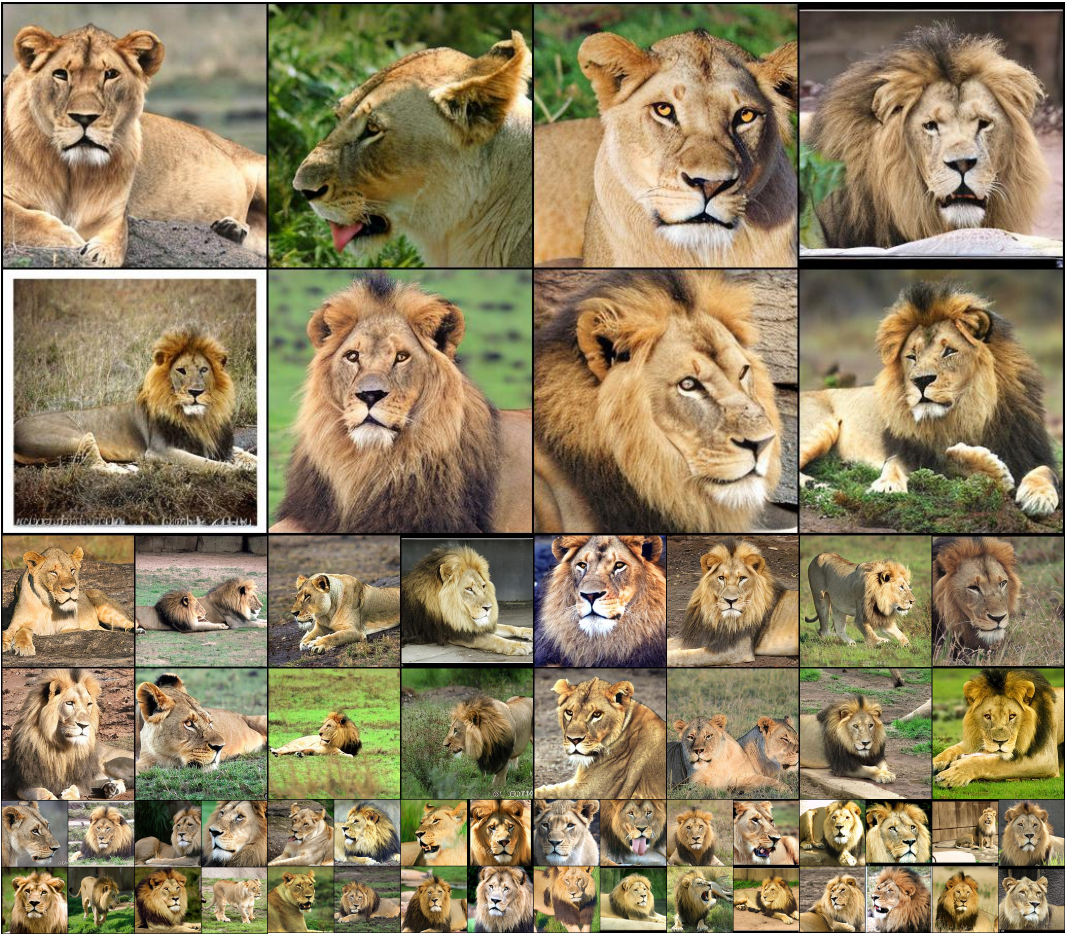}
\caption{\textbf{Uncurated 256$\times$256 DyDiT-XL$_{\lambda=0.5}$ samples. Lion (291).}} 
\end{figure}

\begin{figure}[p]
  \centering
   \includegraphics[height=0.48\textheight, width=1.0\textwidth]{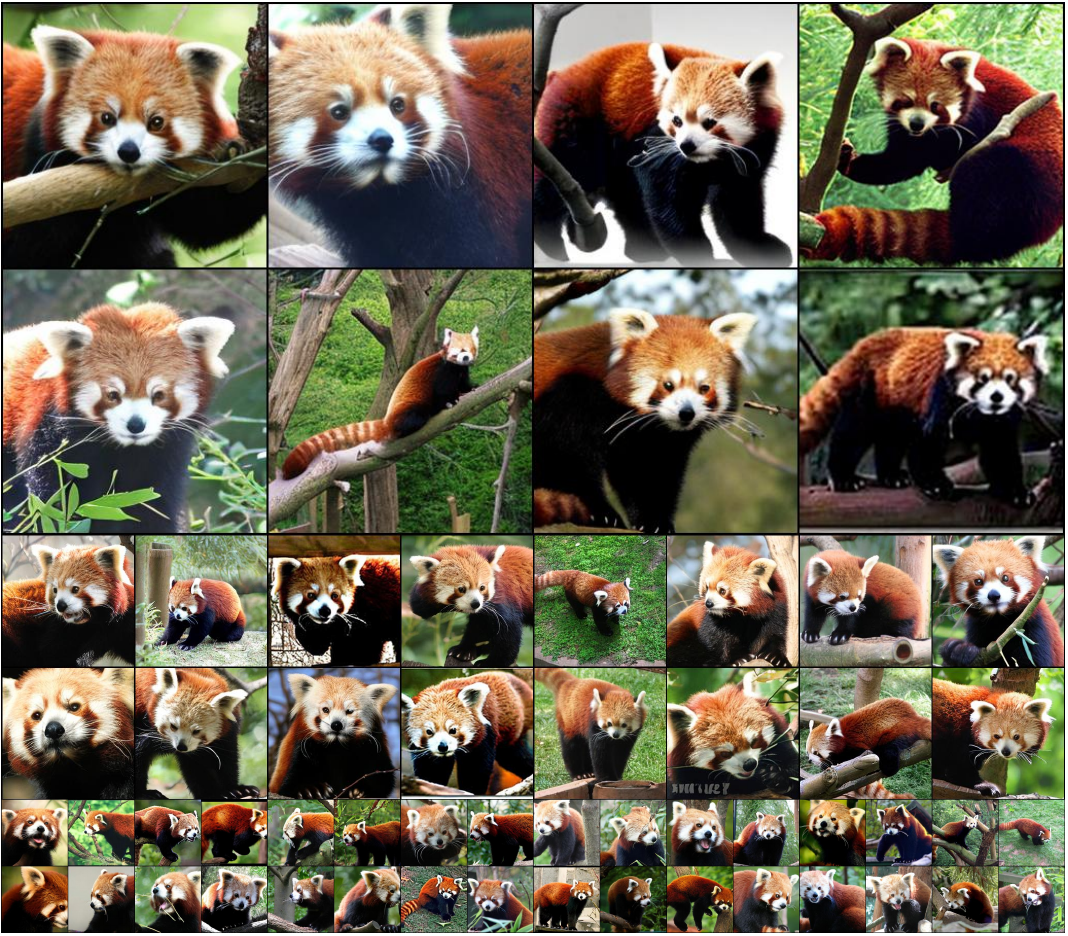}
\caption{\textbf{Uncurated 256$\times$256 DyDiT-XL$_{\lambda=0.5}$ samples. Lesser panda(387).}} 
  \vfill %
     \includegraphics[height=0.48\textheight, width=1.0\textwidth]{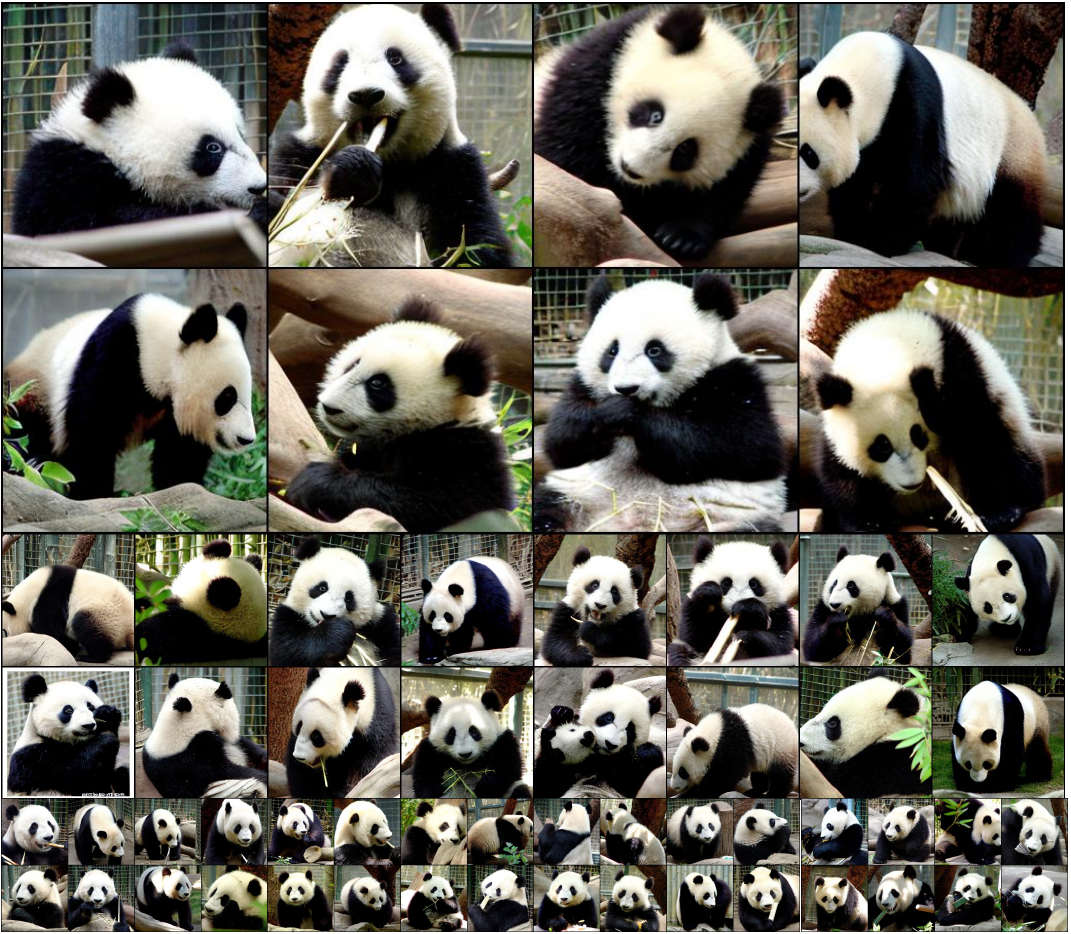}
\caption{\textbf{Uncurated 256$\times$256 DyDiT-XL$_{\lambda=0.5}$ samples. Panda (388).}} 
\end{figure}

\begin{figure}[p]
  \centering

   \includegraphics[height=0.48\textheight, width=1.0\textwidth]{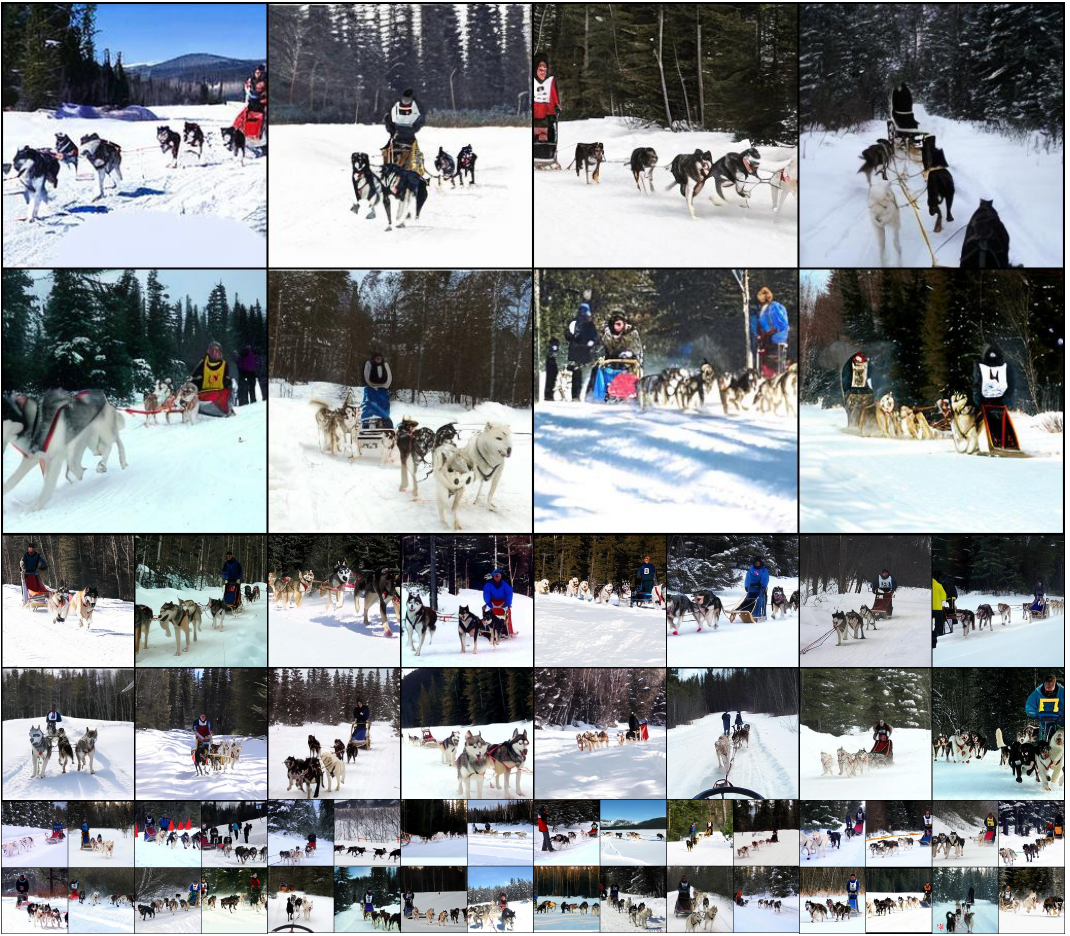}
\caption{\textbf{Uncurated 256$\times$256 DyDiT-XL$_{\lambda=0.5}$ samples. Dogsled (537).}} 
  \vfill %

   \includegraphics[height=0.48\textheight, width=1.0\textwidth]{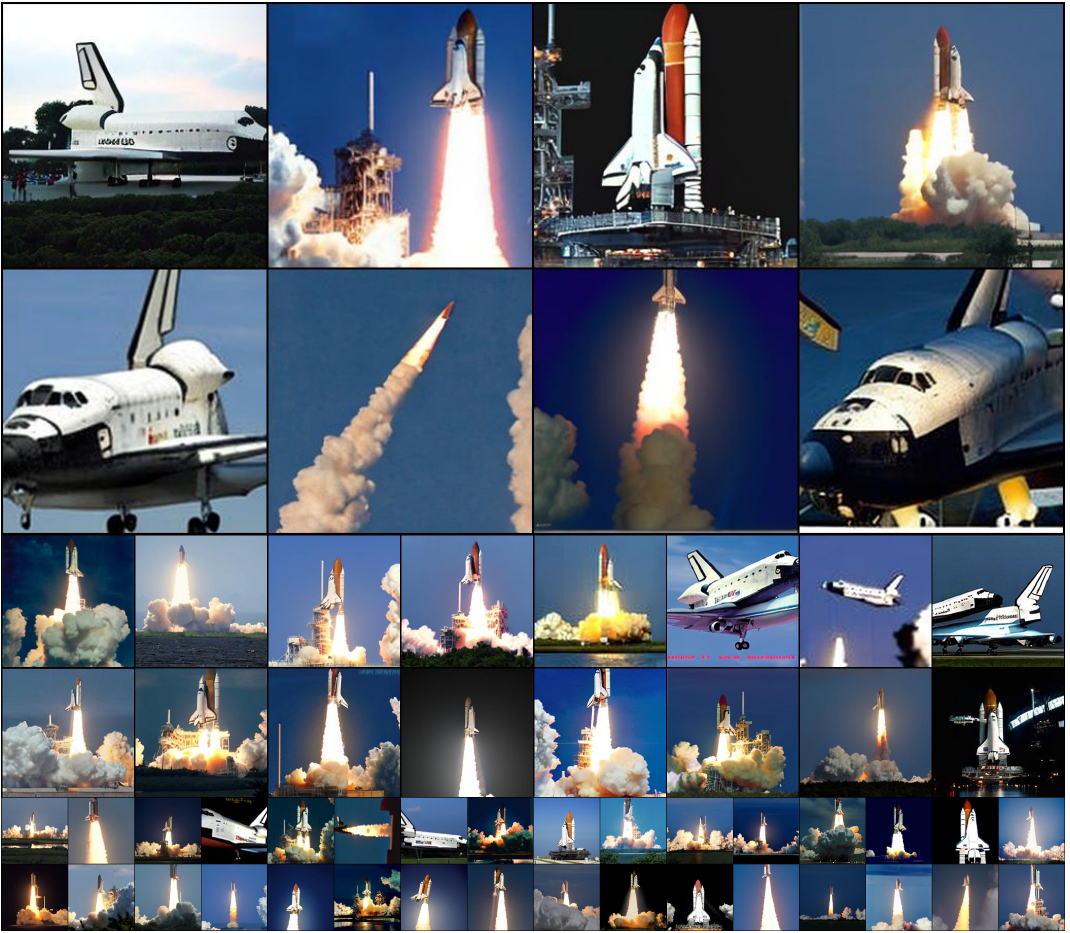}
\caption{\textbf{Uncurated 256$\times$256 DyDiT-XL$_{\lambda=0.5}$ samples. Space shuttle (812).}} 

\end{figure}

\begin{figure}[p]
  \centering

   \includegraphics[height=0.48\textheight, width=1.0\textwidth]{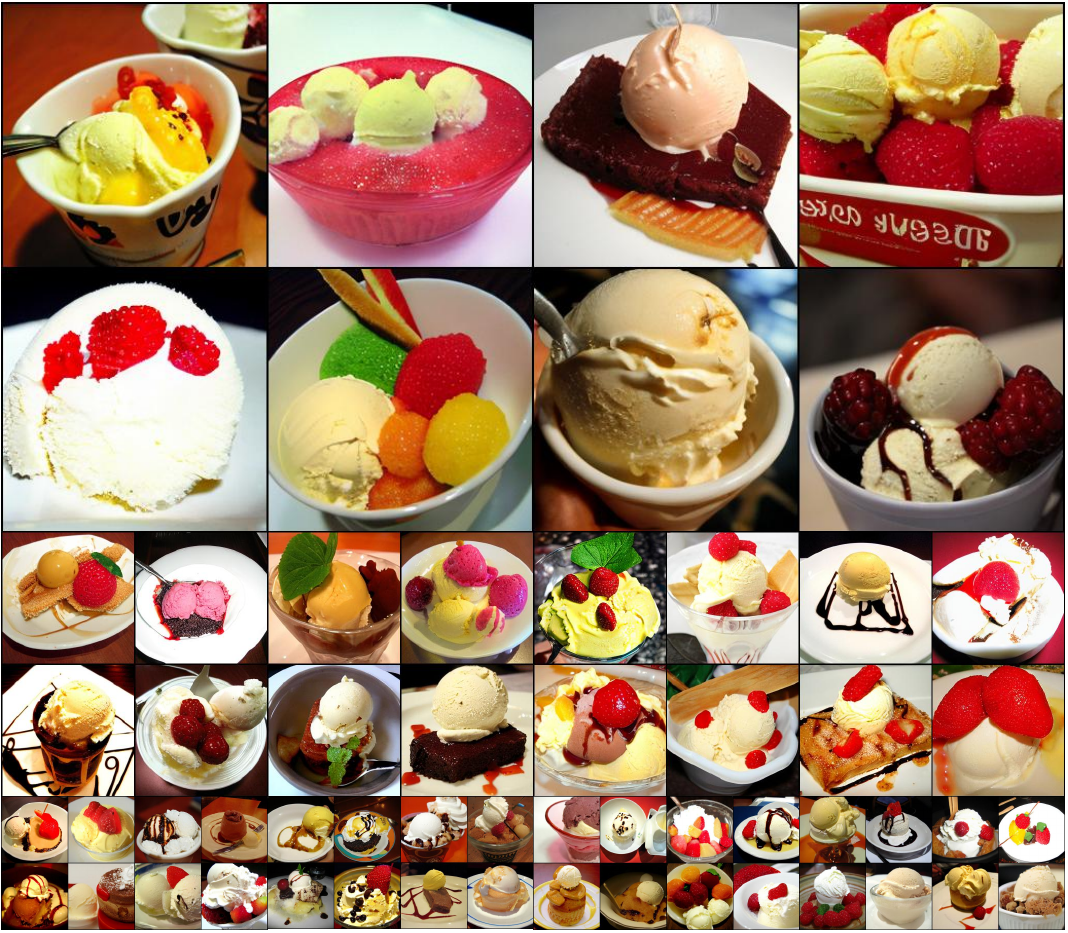}
\caption{\textbf{Uncurated 256$\times$256 DyDiT-XL$_{\lambda=0.5}$ samples. Ice cream (928).}} 
  \vfill %

     \includegraphics[height=0.48\textheight, width=1.0\textwidth]{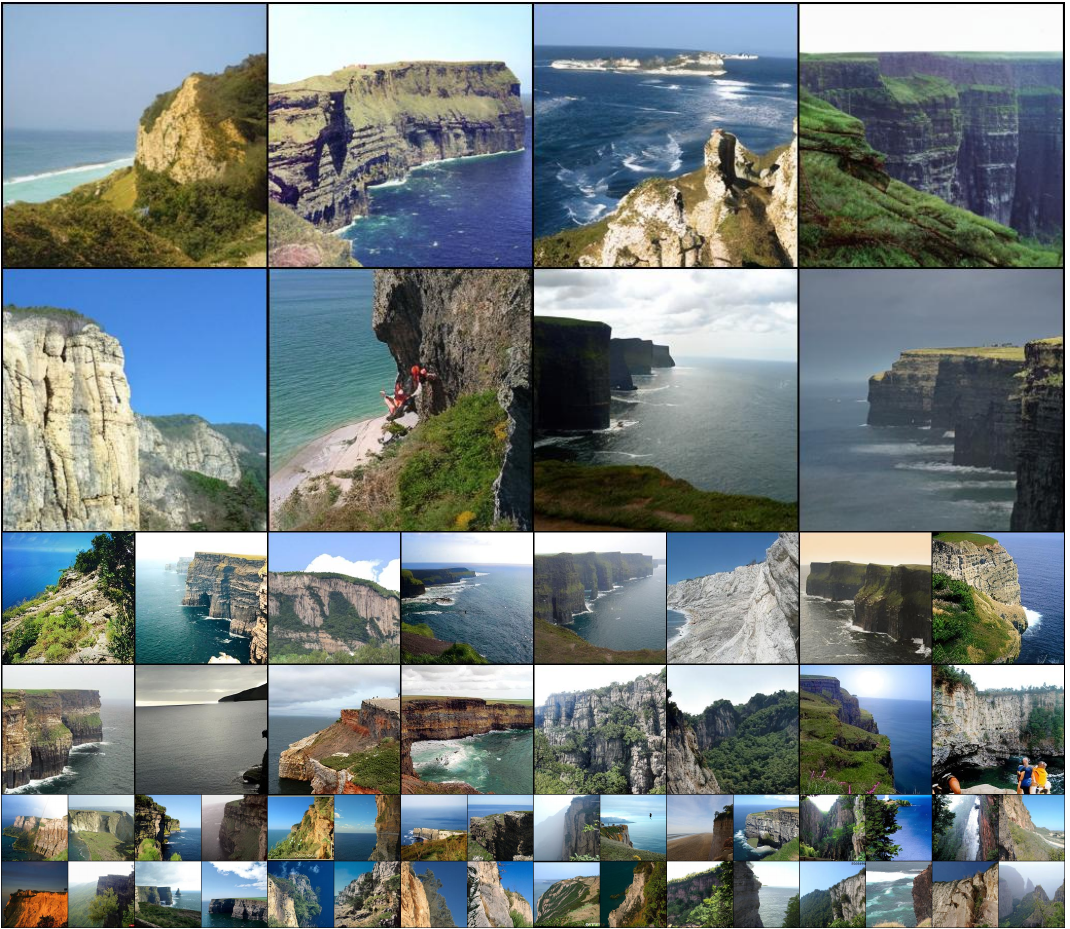}
\caption{\textbf{Uncurated 256$\times$256 DyDiT-XL$_{\lambda=0.5}$ samples. liff(972).}} 
\end{figure}

\begin{figure}[p]
  \centering
     \includegraphics[height=0.48\textheight, width=1.0\textwidth]{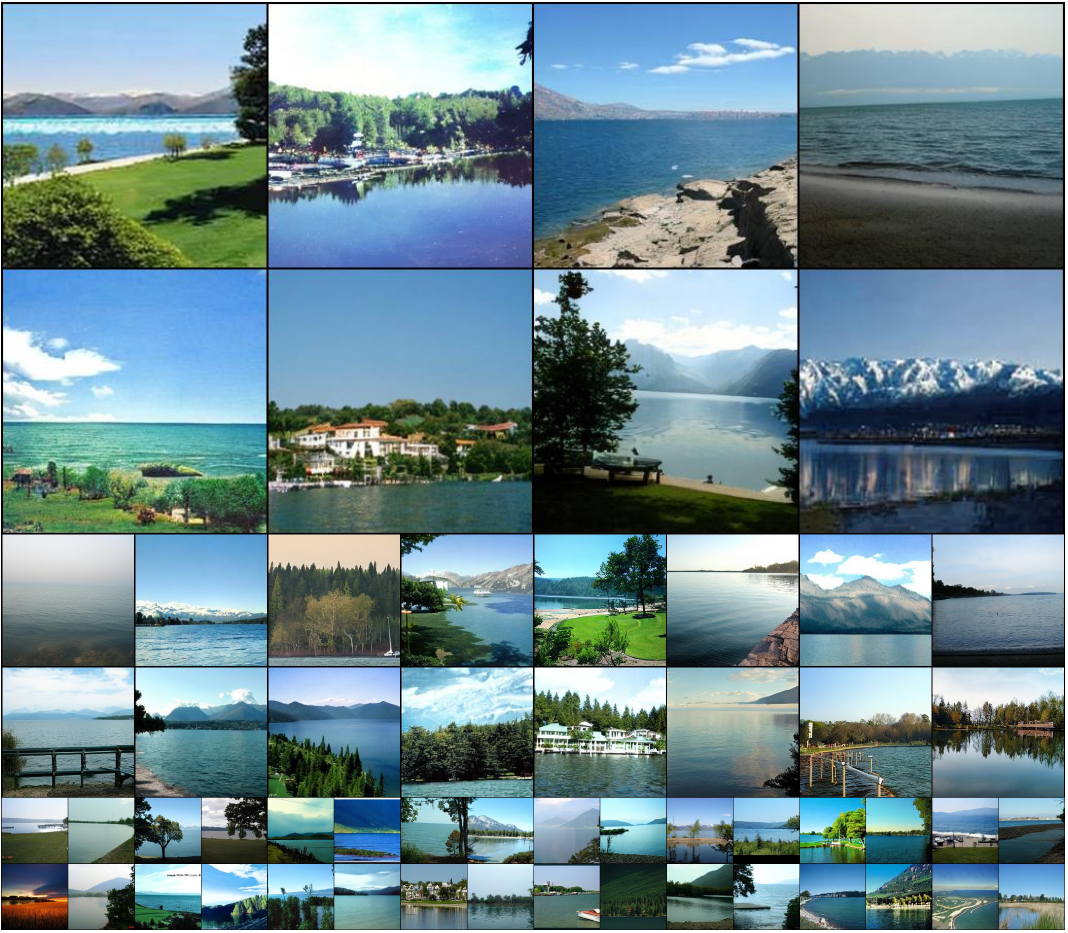}
\caption{\textbf{Uncurated 256$\times$256 DyDiT-XL$_{\lambda=0.5}$ samples. Lakeside (975).}} 
  \vfill %
   \includegraphics[height=0.48\textheight, width=1.0\textwidth]{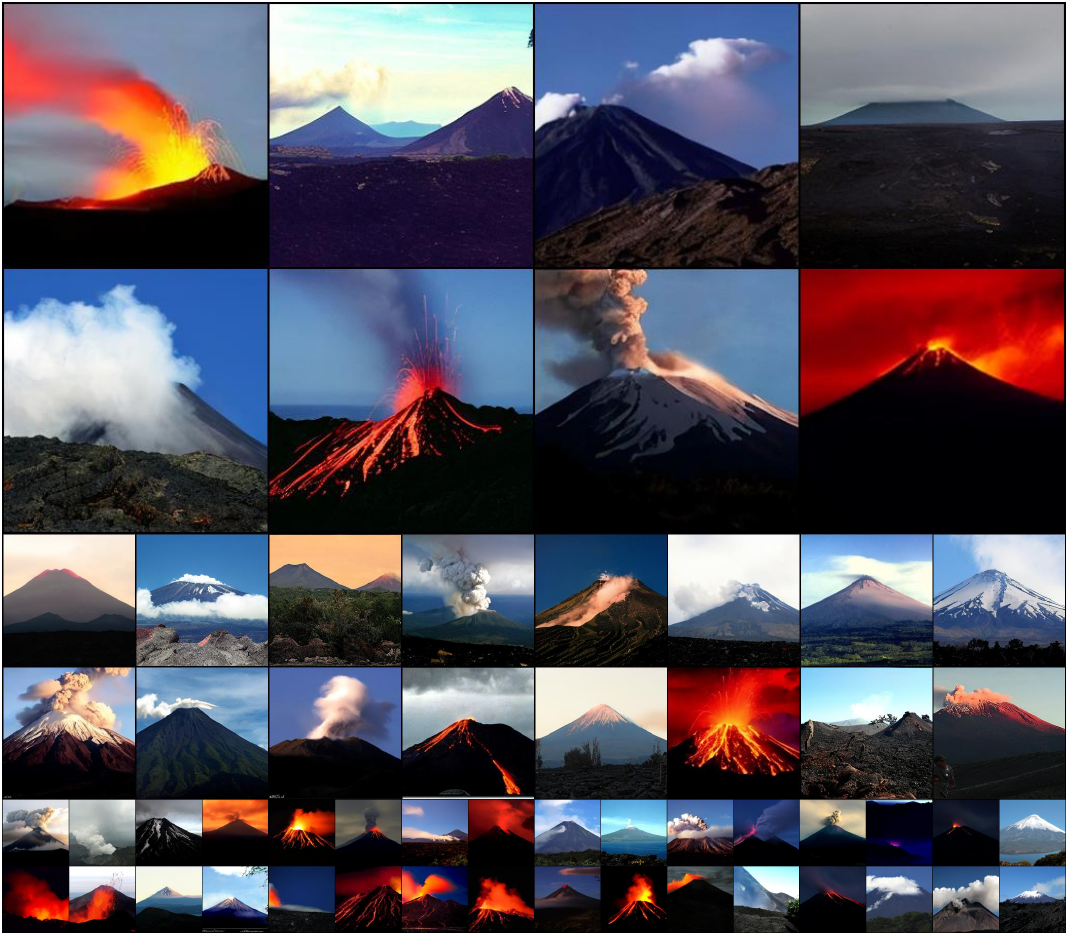}
\caption{\textbf{Uncurated 256$\times$256 DyDiT-XL$_{\lambda=0.5}$ samples. Volcano (980).}} 
\label{fig:visua_final_}
\end{figure}

\end{document}